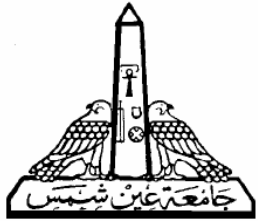

Ain Shams University
Faculty of Engineering
Computer and Systems Engineering Department

# Enhanced Secure Algorithm for Fingerprint Recognition

## A Thesis
Submitted in Partial Fulfillment of the
Requirements for the Degree of
Philosophy of Doctoral in Electrical Engineering
(Computer and Systems)

## Submitted by
**Eng. Amira Mohammad Abdel-Mawgoud Saleh**
B.Sc., Electrical Engineering
(Computer and Systems Engineering Department)
Ain Shams University, 2000
M.Sc., Electrical Engineering
(Computer and Systems Engineering Department)
Ain Shams University, 2004

## Supervised by
**Prof. Dr. Abdel-Moneim A. Wahdan**
**Prof. Dr. Ayman M. Wahba**
**Dr. Ayman Mohammad Bahaa Eldeen Sadeq**
Computer and Systems Engineering Department
Cairo, 2011

بسم الله الرحمن الرحيم

## Examiners Committee

Name:     Amira Mohammad Abdel-Mawgoud Saleh

Thesis:   Enhanced Secure Algorithm for Fingerprint Recognition

Degree:   Doctor of Philosophy in Electrical Engineering

**Name, Title, and Affiliation**                                **Signature**

**1. Prof. Dr. Ali Farag**                                              ………….
Professor of Electrical and Computer Engineering
Director, Computer Vision and Image Processing
Laboratory
Unioversity of Louisville, Kentucky, USA 40292

**2. Prof. Dr. Hani. M. Kamal Mahdi**                          …………..
Computer and Systems Engineering Department
Faculty of Engineering,
Ain Shams University, Cairo, Egypt

**3. Prof. Dr. Abdel-Moneim A. Wahdan** (Supervisor)    ……………
Computer and Systems Engineering Department
Faculty of Engineering,
Ain Shams University, Cairo, Egypt

**4. Dr. Ayman M. Bahaa El-Din** (Supervisor)            …………….
Computer and Systems Engineering Department
Faculty of Engineering,
Ain Shams University, Cairo, Egypt

Date: 12 / 9 / 2011

# Abstract

## Amira Mohammad Abdel-Mawgoud Saleh

## Enhanced Secure Algorithm for Fingerprint Recognition

Philosophy of Doctoral Dissertation

(Ain Shams University - Faculty of Engineering - 2011)

Recognition of persons by means of biometric characteristics is an emerging phenomenon in our society. It has received more and more attention during the last years due to the need for security in a large range of applications. Among the many biometric features, the fingerprint is considered one of the most practical ones. Fingerprint recognition requires a minimal effort from the user, does not capture other information than strictly necessary for the recognition process, and provides relatively good performance. Another reason for the popularity of fingerprints is the relatively low price of fingerprint sensors, which enables easy integration into PC keyboards, smart cards and wireless hardware.

A critical step in fingerprint identification system is thinning of the input fingerprint image. The performance of a minutiae extraction algorithm relies heavily on the quality of the thinning algorithm. So, a fast fingerprint *thinning* algorithm is proposed. The algorithm works directly on the gray-scale image not the binarized one as binarization of fingerprint causes many spurious minutiae and also removes many important features. The performance of the thinning algorithm is evaluated and experimental results show that the proposed thinning algorithm is both fast and accurate.

The next step after thinning of the fingerprint image is minutiae extraction. These minutiae together with the template obtained from the database are used in fingerprint matching. In this study, a new minutiae-based fingerprint *matching* technique is proposed. The main idea is that each fingerprint is represented by a minutiae table of just two columns in the database. The number of different minutiae types (terminations and bifurcations) found in each track of a certain width around the core point of the fingerprint is recorded in this table. Each row in the table represents a certain track, in the first column, the number of terminations in each track is recorded, in the second column, the number of bifurcations in each track is recorded. Since neither minutiae *position* nor *orientation* is considered in tables representing the fingerprints in database, the algorithm is rotation and translation invariant, and needs less storage size. Experimental results show that recognition accuracy is 98%, with Equal Error Rate (EER) of 2%.

Finally, the integrity of the data transmission via communication channels must be secure all the way from the scanner to the application. This is typically achieved by *cryptographic* methods. So, a watermarking algorithm is proposed, to hide the minutiae data (proposed minutiae table) in its corresponding fingerprint using two secret keys $k_1$, and $k_2$ to increase security. The watermarking algorithm is blind i.e. it does not need the original fingerprint to extract the watermark. Two security watermarking applications are studied which can be used to guarantee secure transmission of acquired fingerprint images from intelligence agencies to a central image database and to eliminate several types of biometric system attacks.

Experimental results show that watermark is both invisible and robust against Gaussian noise addition, and JPEG compression with high and moderate quality factors. After watermarking and without attacks, recognition accuracy does not change and still is 98%. After applying Gaussian noise addition, and JPEG compression with high and moderate quality factors on the watermarked fingerprint images, recognition accuracy decreases slightly to reach 96%.

# List of publications

1. A. M. Saleh, A. M. Bahaa Eldin, and A.-M. A. Wahdan, "A Modified Thinning Algorithm for Fingerprint Identification Systems," in *International Conference on Computer Engineering and Systems ICCES* Cairo, Egypt, December 14-16, 2009, pp. 371-376.

2. Amira Saleh, Ayman Bahaa and A. Wahdan (2011). Fingerprint Recognition, Advanced Biometric Technologies, Girija Chetty and Jucheng Yang (Ed.), ISBN: 978-953-307-487-0, InTech, Available from: http://www.intechopen.com/articles/show/title/fingerprint-recognition

# Acknowledgement

First, I would like to thank ALLAH for his great support to me in accomplishing this work.

I would like to express my deepest gratitude to Prof. Abdel-Moneim Wahdan for suggesting of the point of research, his valuable advices, and his great efforts revising the thesis. I can not express my thanks to him for his care and guidance that started long ago since I was an undergraduate student and during my Master of Science degree. He actually gave me the chance to express my capabilities and to put my first step in scientific research.

I would like also to thank Dr. Ayman Mohammad Bahaa Eldeen Sadeq for his great efforts along the thesis development since it was just thoughts in mind. I would like to thank him for his great support and for supplying me with all the materials and references I needed.

I would like to thank Eng. Ahmad Zaki, member of the staff of Computer and System Engineering Department, for his help, support and encouragement.

Also, I would like to thank Mrs Seham, the secretary of our department. She was always helpful, kind and answering all my questions at anytime and anywhere.

I would like also to express my deepest gratitude for my dear mother for her patience, great support and encouragement during the most critical stages of the thesis. I thank her for taking responsibility of my daughter and my son providing me with full means of comfort to concentrate well in my work. Also, I would like to thank my brothers, especially Ahmad, who supported me with useful resources, care and encouragement.

At last but not least, I would like to thank my beloved husband for his care, patience, and encouragement during the thesis development. In fact, I don't find the words that would fairly express my thankfulness and gratitude to him.

Finally, I wished my dear father, Dr. Mohammad Abdel-Mawgoud, attend this happy occasion of obtaining the doctoral degree about which he cared a lot in his last words to me. May ALLAH send his blessing upon him.

# Statement

This dissertation is submitted to Ain Shams University for the degree of Philosophy of Doctoral in Electrical Engineering (Computer and Systems Engineering).

The work included in this thesis was carried out by the author at the Computer and Systems Engineering Department, Ain Shams University.

No part of this thesis has been submitted for a degree or qualification at other university or institution.

Date:
Name: Amira Mohammad Saleh
Signature:

# Table of contents

# List of Tables

# List of Figures

# List of Symbols and Abbreviations

AFIS:      Automatic Fingerprint Identification System
ATM:      Automatic Teller Machine
BC:      Before Christ
CPU:      Central Processing Unit
DB:      Database
DCT:      Discrete Cosine Transformation
DoS:      Denial Of Service
DWT:      Discrete Wavelet Transformation
EER      Equal-Error Rate
FAR      False Acceptance Rate
FBI:      Federal Bureau of Investigation
FFT:      Fast Fourier Transform
FMR:      False Match Rate
FNMR:      False Non-Match Rate
FRR      False Rejection Rate
FTC:      Failure To Capture
FTE:      Failure To Enroll
FTM:      Failure To Match
FVC:      Fingerprint Verification Competition
GB:      Gega Byte
GHz      Gega Hertz
GMS:      Genuine Matching Scores
HVS:      Human Visual System
ICS:      Intrinsic Coordinate System
ID:      Identification
IEEE:      Institute of Electrical and Electronic Engineering
IFFT:      Inverse Fast Fourier Transform
IMS:      Imposter Matching Scores
JPEG:      Joint Photographic Experts Group
Max:      Maximum
Min:      Minimum
msec:      Milli Second
MTF:      Modulation Transfer Function
NGRA:      Number Of Genuine Recognition Attempts
NIRA:      Number Of Impostor Recognition Attempts
NIST:      National Institute Of Standards And Technology
PC:      Personal Computer

| | |
|---|---|
| PIII: | Pentium 3 |
| PIN: | Personal Identification Number |
| Pixel: | Picture Element |
| PRNG: | Pseudo Random Number Generator |
| Q: | Quality Factor of JPEG compression |
| RAM: | Random Access Memory |
| ROC: | Receiver Operating Curve |
| RSA: | Rivest-Shamir-Adleman |
| Sec: | Second |
| Sim: | Similarity |
| STFT: | Short Time Fourier Transform |
| Thr: | Threshold |
| ZFMR: | Zero False Match Rate |
| ZFNMR: | Zero False Non-Match Rate |

# Chapter 1
# Introduction

## 1.1. Biometric Recognition

As our society have become electronically connected and more mobile, surrogate representations of identity such as passwords (prevalent in electronic access control) and cards (prevalent in banking and government applications) cannot be trusted to establish a person's identity. Cards can be lost or stolen and passwords or PIN can, in most cases, be guessed. Further, passwords and cards can be easily shared and so they do not provide non-repudiation [1].

*Biometric recognition* (or simply biometrics) refers to the use of distinctive *anatomical* (e.g., fingerprints, face, iris) and *behavioral* (e.g., speech) characteristics, called *biometric identifiers, traits* or *characteristics* for automatically recognizing individuals. Biometrics are becoming essential components of effective person identification solutions because biometric identifiers cannot be shared or misplaced, and they intrinsically represent the individual's bodily identity. Recognition of a person by their body, then linking that body to an externally established "identity", forms a very powerful tool of identity management with tremendous potential consequences, both positive and negative. Consequently, biometrics is not only a fascinating pattern recognition research problem but, if carefully used, is an enabling technology with the potential to make our society safer, reduce fraud and provide user convenience (user friendly man–machine interface).

The word *biometrics* is derived from the Greek words *bios* (meaning life) and *metron* (meaning measurement); biometric identifiers are measurements from living human body. Perhaps all biometric identifiers are a combination of anatomical and behavioral characteristics and they should not be exclusively classified into either anatomical or behavioral characteristics. For example, fingerprints are anatomical in nature but the usage of the input device (e.g., how a user presents a finger to the fingerprint scanner) depends on the person's behavior. Thus, the input to the recognition engine is a combination of anatomical and behavioral characteristics. Similarly, speech is partly determined by the vocal tract that produces speech and partly by the way a person speaks. Often, a similarity can be noticed among parents, children, and siblings in their speech. The same argument applies to the face: faces of identical twins may be extremely similar at birth but during their growth and development, the faces change based on the person's behavior (e.g., lifestyle differences leading to a difference in bodyweight, etc.).

## 1.2. History of Fingerprinting

There is archaeological evidence that fingerprints as a form of identification have been used at least since 7000 to 6000 BC by the ancient Assyrians and Chinese. Clay pottery from these times sometimes contains fingerprint impressions placed to mark the potter. Chinese documents bore a clay seal marked by the thumbprint of the originator. Bricks used in houses in the ancient city of Jericho were sometimes imprinted by pairs of thumbprints of the bricklayer. However, though fingerprint individuality was recognized, there is no evidence this was used on a universal basis in any of these societies[2].

In the mid-1800's scientific studies were begun that would established two critical characteristics of fingerprints that are true still to this day: no two fingerprints from different fingers have been found to have the same ridge pattern, and fingerprint ridge patterns are unchanging throughout life. These studies led to the use of fingerprints for criminal identification, first in Argentina in 1896, then at Scotland Yard in 1901, and to other countries in the early 1900's.

Computer processing of fingerprints began in the early 1960s with the introduction of computer hardware that could reasonably process these images. Since then, automated fingerprint identification systems (AFIS) have been deployed widely among law enforcement agencies throughout the world.

In the 1980s, innovations in two technology areas, personal computers and optical scanners, enabled the tools to make fingerprint capture practical in non-criminal applications such as for ID-card programs. In the late 1990s, the introduction of inexpensive fingerprint capture devices and the development of fast, reliable matching algorithms have set the stage for the expansion of fingerprint matching to personal use.

## 1.3. Organization of The Thesis

This thesis is presented in six chapters. Chapter 1 is an introduction to fingerprinting. Chapter 2 addresses the concept of fingerprint analysis and matching. Chapter 3 describes a proposed thinning algorithm for fingerprints. A new fingerprint matching algorithm is proposed in Chapter 4. Chapter 5 discusses a new watermarking algorithm to secure Fingerprint Identification System (FIS). Chapter 6 presents conclusions and suggests future work.

# Chapter 2

# Fingerprint Analysis and Matching

Recognition of persons by means of biometric characteristics is an emerging phenomenon in modern society. It has received more and more attention during the last period due to the need for security in a wide range of applications. Among the many biometric features, the fingerprint is considered one of the most practical ones. Fingerprint recognition requires a minimal effort from the user, does not capture other information than strictly necessary for the recognition process, and provides relatively good performance. Another reason for the popularity of fingerprints is the relatively low price of fingerprint sensors, which enables easy integration into PC keyboards, smart cards and wireless hardware [3].

Figure 2.1 presents a general framework for a general fingerprint identification system (FIS) [4].

A *fingerprint-based biometric system* is essentially a pattern recognition system that recognizes a person by determining the authenticity of his fingerprint [1]. Depending on the application context, a fingerprint-based biometric system may be called either a verification system or an identification system.

Verification system:

To authenticate a person's identity by comparing the captured fingerprints with his own biometric template(s) pre-stored in the system. It conducts *one-to-one* comparison to determine whether the identity claimed by the individual is true.

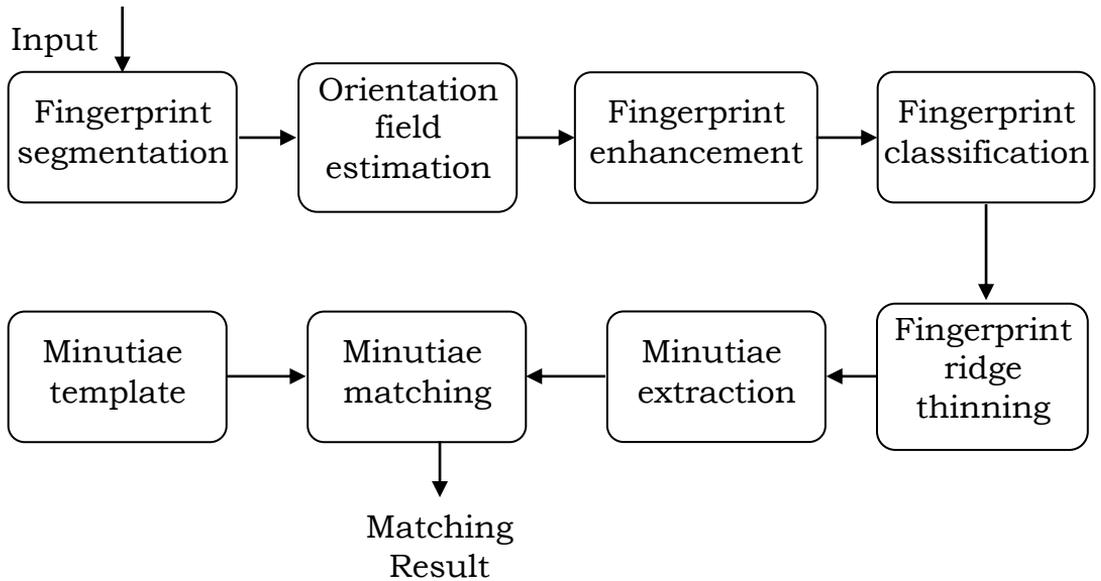

*Figure 2.1. General block diagram for a Fingerprint Identification System.*

Identification system:

To recognize an individual by searching the entire templates database for a match. It conducts *one-to-many* comparisons to establish the identity of the individual.

Block diagrams of a fingerprint-based *verification* system and an *identification* system are depicted in Figure 2.2, user *enrollment*, which is common to both tasks, is also graphically illustrated.

The *enrollment* module is responsible for registering individuals in the biometric system database (system DB). During the enrollment phase, the fingerprint of an individual is acquired by a fingerprint scanner to produce a raw digital representation. A quality check is generally performed to ensure that the acquired sample can be reliably processed by successive stages. In order to facilitate matching, the raw digital representation is usually further processed by a feature extractor to generate a compact but expressive representation, called a template.

The *verification* task is responsible for verifying individuals at the point of access. During the operation phase, the user's ID is presented to the system; the biometric reader captures the fingerprint of the individual to be recognized and converts it to a digital format, which is further processed by the feature extractor to produce a compact digital representation. The resulting representation is fed to the feature matcher, which compares it against the template of a single user (retrieved from the system DB based on the user's ID).

In the *identification* task, no ID is provided and the system compares the representation of the input biometric against the templates of all the users in the system database; the output is either the identity of an enrolled user or an alert message such as "user not identified."

It is evident from Figure 2.2 that the main building blocks of any fingerprint-based verification and identification system are:

1) Sensing,
2) Feature extraction, and
3) Matching.

In this thesis, only feature extraction and matching procedures are studied in chapter 3 and 4. Then to secure the fingerprint identification system, watermarking technique is applied in chapter 5.

## 2.1. Feature Extraction

A fingerprint is the reproduction of a fingertip epidermis, produced when a finger is pressed against a smooth surface. The most evident structural characteristic of a fingerprint is a pattern of interleaved *ridges* and *valleys*; in a fingerprint image, ridges (also called ridge lines) are dark whereas valleys are bright (see Figure 2.3). Ridges and valleys often run in parallel; sometimes they bifurcate and sometimes they terminate.

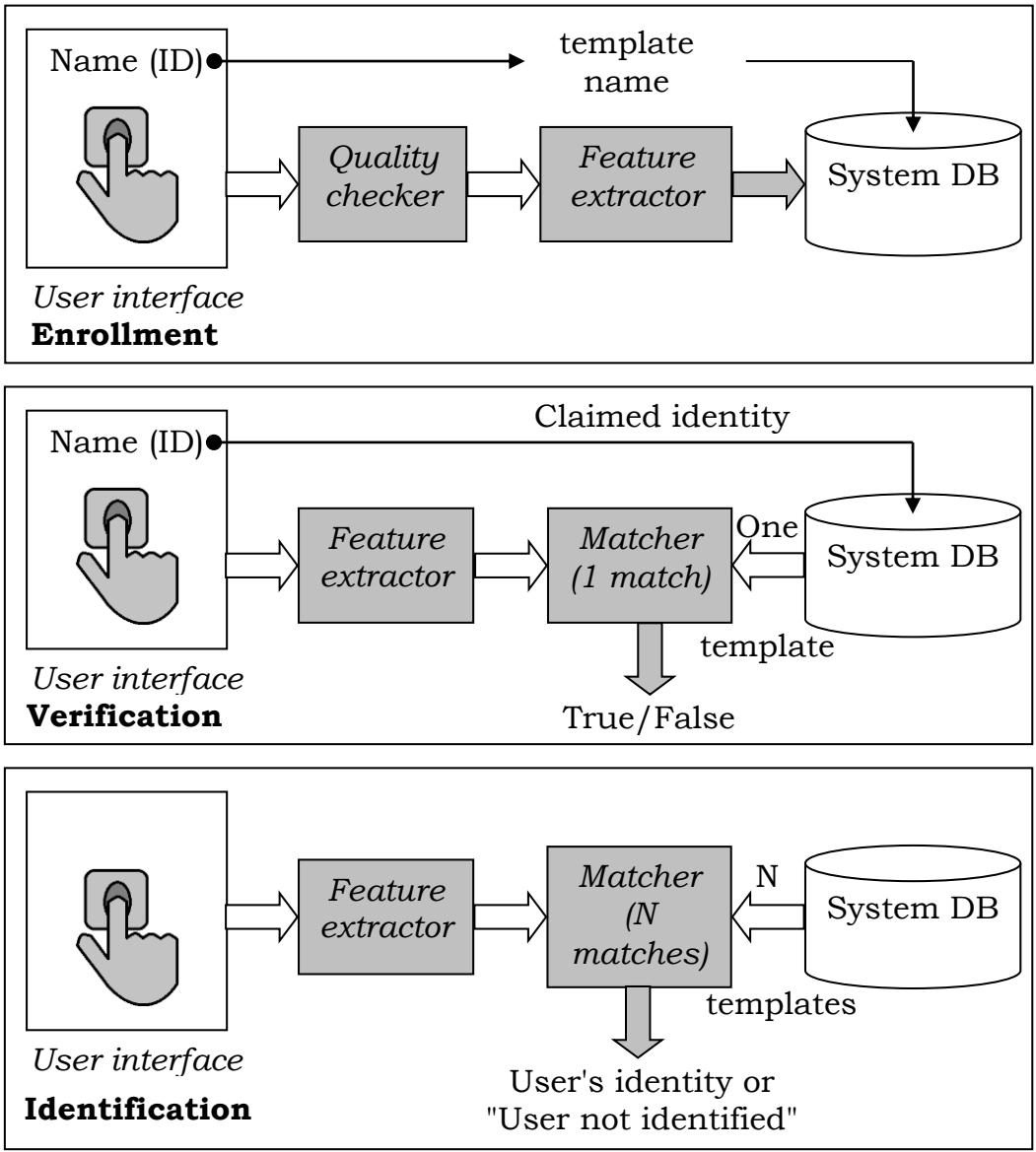

*Figure 2.2. Block diagrams of enrollment, verification and identification tasks.*

When analyzed at the global level, the fingerprint pattern exhibits one or more regions where the ridge lines assume distinctive shapes (characterized by high curvature, frequent termination, etc.). These regions (called *singularities* or *singular regions*) may be classified into three typologies: *loop*, *delta*, and *whorl* (see Figure 2.4). Singular regions belonging to loop, delta, and whorl types are typically characterized by ∩, ∆, and O shapes, respectively.

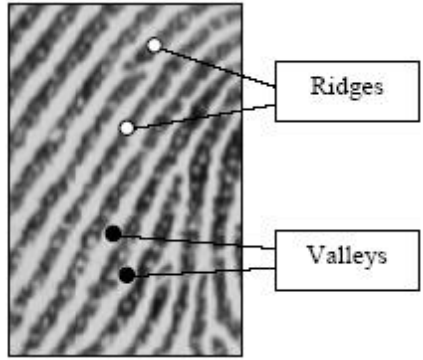

*Figure 2.3. Ridges and Valleys in a fingerprint image*

Several fingerprint matching algorithms prealign fingerprint images according to a landmark or a center point, called the *core*. The core point corresponds to the center of the top most loop type singularity. For fingerprints that do not contain loop or whorl singularities (e.g., those belonging to the Arch class in Figure 2.5), it is difficult to define the core. In these cases, the core is usually associated with the point of maximum ridge line curvature. Unfortunately, due to the high variability of fingerprint patterns, it is difficult to reliably locate a registration (core) point in all the fingerprint images.

Singular regions are commonly used for fingerprint classification (see Figure 2.5), that is, assigning a fingerprint to a class among a set of distinct classes, with the aim of simplifying search and retrieval.

At the local level, other important features, called *minutiae* can be found in the fingerprint patterns. Minutia refers to various ways that the ridges can be discontinuous.

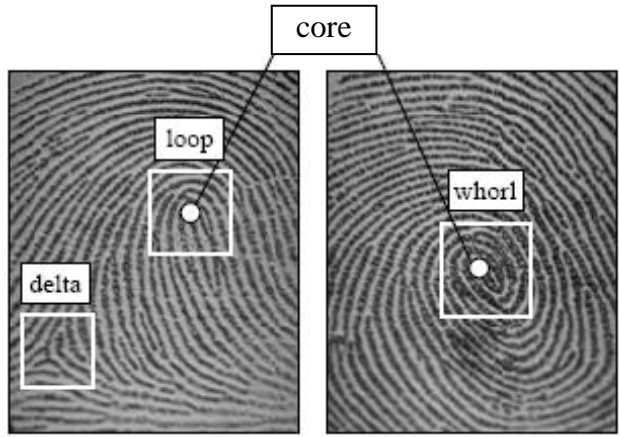

*Figure 2.4. Singular regions (white boxes) and core points (small circles) in fingerprint images.*

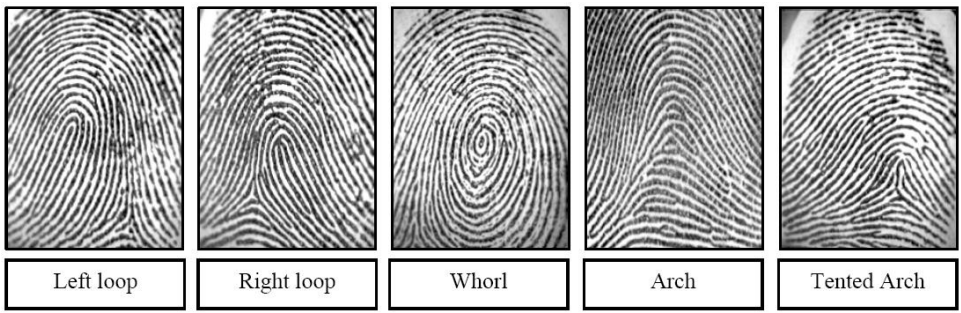

*Figure 2.5. One fingerprint from each of the five major classes defined by Henry*

For example, a ridge can suddenly come to an end (termination), or can divide into two ridges (bifurcation) (see Figure 2.6).

Some preprocessing and enhancement steps are often performed to simplify the task of minutiae extraction. In the following, these steps are briefly discussed.

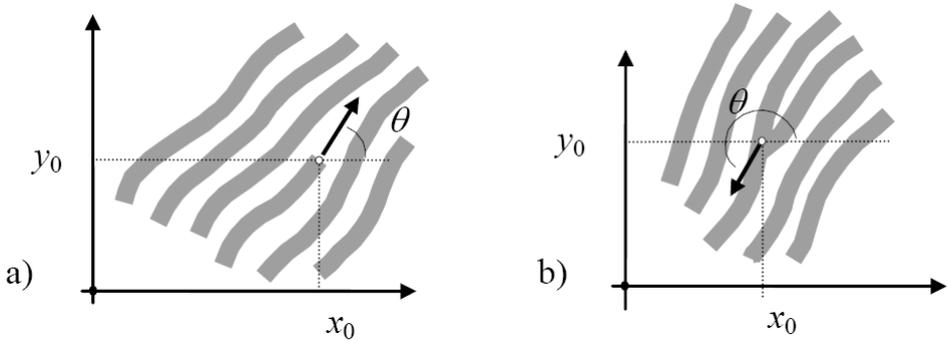

*Figure 2.6. a) A termination minutia: [$x_0$, $y_0$] are the minutia coordinates; θ is the angle that the minutia tangent forms with the horizontal axis; b) a bifurcation minutia: θ is now defined by means of the termination minutia corresponding to the original bifurcation that exists in the negative image.*

### 2.1.1.      Local Ridge Orientation

The local ridge orientation at [$x$, $y$] is the angle $θ_{xy}$ that the fingerprint ridges, crossing through an arbitrary small neighborhood centered at [$x$, $y$], form with the horizontal axis; see Figure 2.6.

The simplest and most natural approach for extracting local ridge orientation is based on computation of gradients in the fingerprint image $I$. The gradient $\nabla(x, y)$ at point [$x$, $y$] of $I$, is a two dimensional vector [$\nabla x(x, y)$, $\nabla y(x, y)$], where $\nabla x$ and $\nabla y$ components are the derivatives of $I$ at [$x$, $y$] with respect to the $x$ and $y$ directions, respectively. $\nabla x$ and $\nabla y$ components of the gradient can be determined using the classical Prewitt or Sobel convolution masks [5], then $θ_{ij}$ is computed as the arctangent of the $\nabla y/\nabla x$ ratio. It is well known that the gradient phase angle denotes the direction of the maximum intensity change. Therefore, the direction $θ_{ij}$ of a hypothetical edge that crosses the region centered at [$x_i$, $y_j$] is orthogonal to the gradient phase angle at [$x_i$, $y_j$].

This method, although simple and efficient, has some drawbacks:

- Using the classical convolution masks to determine $\nabla x$, $\nabla y$, and $\theta_{ij}$ presents problems due to the non-linearity and discontinuity around 90°.
- A single orientation estimate reflects the ridge-valley orientation at too fine a scale and is generally very sensitive to the noise in the fingerprint image.

A simple but elegant solution to the above problem, which allows local gradient estimates to be averaged, have been proposed in [6], [7], and [8].

### 2.1.2.     Local Ridge Frequency

The local ridge frequency (or density) $f_{xy}$ at point [$x$, $y$] is the inverse of the number of ridges per unit length along a hypothetical segment centered at [$x$, $y$] and orthogonal to the local ridge orientation $\theta_{xy}$. The local ridge frequency varies across different fingers, and may also noticeably vary across different regions in the same fingerprint. The local ridge frequency feature is not used along the thesis.

In [9], local ridge frequency is estimated by counting the average number of pixels between two consecutive peaks of gray-levels along the direction normal to the local ridge orientation (see Figure 2.7). The frequency $f_{ij}$ at [$x_i$, $y_j$] is computed as follows:

1. A 32 × 16 oriented window centered at [$x_i$, $y_j$] is defined in the ridge coordinate system (i.e., rotated to align the $j$-axis with the local ridge orientation).
2. The $x$-signature of the gray-levels is obtained by accumulating, for each column $x$, the gray-levels of the corresponding pixels in the oriented window. This is a sort of averaging that makes the gray-level profile smoother and prevents ridge peaks from being obscured due to small ridge breaks or pores.
3. $f_{ij}$ is determined as the inverse of the average distance between two consecutive peaks of the $x$-signature.

Another two-step procedure is proposed in [10]:

1. The average ridge distance is estimated in the Fourier domain for each 64 × 64 sub-block of the image that is of sufficient quality, and then
2. This information is propagated, according to a diffusion equation, to the remaining regions.

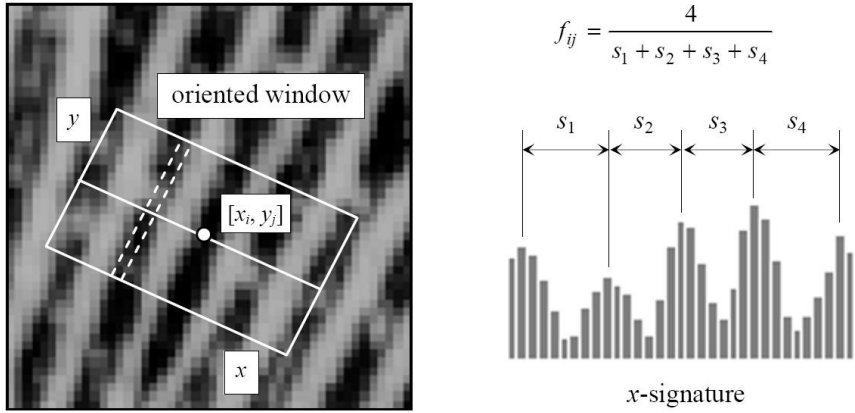

*Figure 2.7. An oriented window centered at [$x_i$, $y_j$]; the dashed lines show the pixels whose gray-levels are accumulated for a given column of the x-signature[9]. The x-signature on the right clearly exhibits five peaks; the four distances between consecutive peaks are averaged to determine the local ridge frequency.*

## 2.1.3.      Segmentation

Separating the fingerprint area from the background is useful to avoid extraction of features in noisy areas of the fingerprint and background [1].

Because fingerprint images are striated patterns, using a global or local thresholding technique [5] does not allow the fingerprint area to be effectively isolated. In fact, what really discriminates foreground and background is not the average image intensities but the presence of a striped and oriented pattern in the foreground and of an isotropic pattern (i.e., which does not have a dominant orientation) in the background. If the image background were always uniform and lighter than the fingerprint area, a simple approach based on local intensity could be effective for discriminating foreground and background; in practice, the presence of noise (such as that produced by dust and grease on the surface of live-scan fingerprint scanners) requires more robust segmentation techniques [7] and [11].

In [12], foreground and background are discriminated by using the average magnitude of the gradient in each image block; in fact, because the fingerprint area is rich in edges due to the ridge/valley alternation, the gradient response is high in the fingerprint area and small in the background.

### 2.1.4.        Singularity and Core Detection

Most of the approaches proposed in the literature for singularity detection operate on the fingerprint orientation image. The best-known method is based on Poincaré index[1].

**Poincaré index method:**
Let $G$ be a vector field and $C$ be a curve immersed in $G$; then the Poincaré index $P_{G,C}$ is defined as the total rotation of the vectors of $G$ along $C$ (see Figure 2.8).

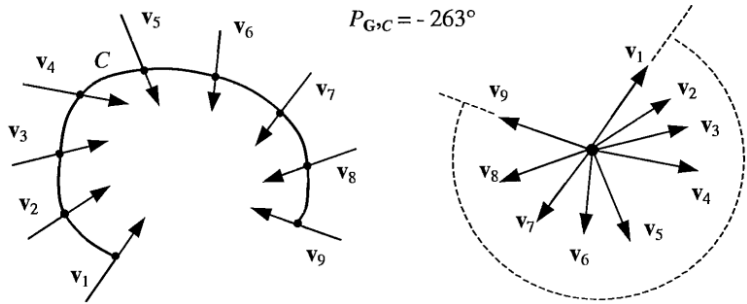

*Figure 2.8. The Poincaré index computed over a curve C immersed in a vector field G.*

Let $G$ be the field associated with a fingerprint orientation image $D$ and let $[i, j]$ be the position of the element $\theta_{ij}$ in the orientation image; then the Poincaré index $P_{G,C}(i, j)$ at $[i, j]$ is computed as follows.

- The curve *C* is a closed path defined as an ordered sequence of some elements of *D*, such that [*i*, *j*] is an internal point;

- $P_{G,C}(i, j)$ is computed by algebraically summing the orientation differences between adjacent elements of *C*. Summing orientation differences requires a direction (among the two possible) to be associated at each orientation. A solution to this problem is to randomly select the direction of the first element and assign the direction closest to that of the previous element to each successive element. It is well known and can be easily shown that, on closed curves, the Poincaré index assumes only one of the discrete values: 0°, ±180°, and ±360°. In the case of fingerprint singularities:

$$P_{G,C}(i, j) = \begin{cases} 0° \text{ if } [i, j] \notin \text{ any singular region.} \\[2ex] 360° \text{ if } [i, j] \in \text{ whorl type singular region.} \\[2ex] 180° \text{ if } [i, j] \in \text{ loop type singular region.} \\[2ex] \text{-}180° \text{ if } [i, j] \in \text{ delta type singular region.} \end{cases}$$

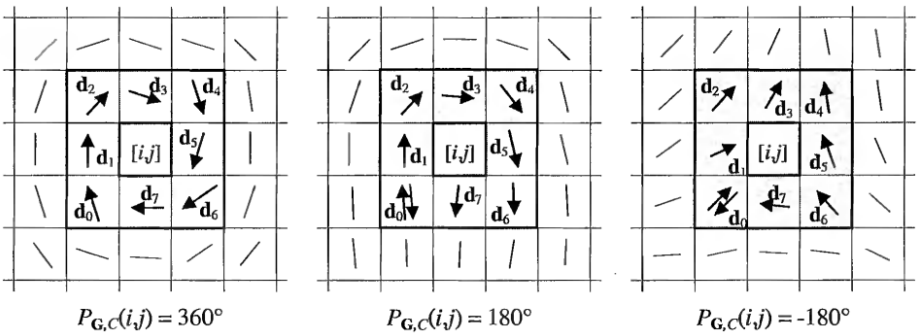

$P_{\mathbf{G},C}(i,j) = 360°$          $P_{\mathbf{G},C}(i,j) = 180°$          $P_{\mathbf{G},C}(i,j) = \text{-}180°$

*Figure 2.9. Example of computation of the Poincaré index in the 8-neighborhood of points belonging (from the left to the right) to a whorl, loop, and delta singularity, respectively. Note that for the loop and delta examples (center and right), the direction of $d_0$ is first chosen upward (to compute the angle between $d_0$ and $d_1$) and then successively downward (when computing the angle between $d_7$ and $d_0$).*

Figure 2.9 shows three portions of orientation images. The path defining C is the ordered sequence of the eight elements $d_k$ (k = 0...7) surrounding [$i, j$]. The direction of the elements $d_k$ is chosen as follows: $d_0$ is directed upward; $d_k$ (k = 1...7) is directed so that the absolute value of the angle between $d_k$ and $d_{k-1}$ is less than or equal to 90°. The Poincaré index is then computed as:

$$P_{G,C}(i, j) = \sum_{k=0...7} angle(d_k, d_{(k+1) \bmod 8})$$                    (2.1)

An interesting implementation of the Poincaré method for locating singular points was proposed in [8]: according to Green's theorem, a closed line integral over a vector field can be calculated as a surface integral over the rotation of this vector field; in practice, instead of summing angle differences along a closed path, the authors compute the "rotation" of the orientation image (through a further differentiation) and then perform a local integration (sum) in a small neighborhood of each element. Also, in [8], a method is provided for associating an orientation with each singularity; this is done by comparing the orientation image around each detected singular point with the orientation image of an ideal singularity of the same type.

Singularity detection in noisy or low-quality fingerprints is difficult and the Poincaré method may lead to the detection of false singularities. Regularizing the orientation image through a local averaging is often quite effective in preventing the detection of false singularities.

Based on the observation that only a limited number of singularities can be present in a fingerprint, in [13], authors proposed to iteratively smooth the orientation image (through averaging) until a valid number of singularities is detected by the Poincaré index. In fact, a simple analysis of the different fingerprint classes shows that:

- Arch fingerprints do not contain singularities;
- Left loop, right loop, and tented arch fingerprints contain one loop and one delta;

- Whorl fingerprints contain two loops (or one whorl) and two deltas.

Once the singularities have been extracted, the core position may be simply defined as the location of the north most loop. Some problems arise with the arch type fingerprints that do not have singularities, and sometimes present a quite flat flow field where no discriminant positions can be located. When the core point is detected with the aim of registering fingerprint images (thus obtaining invariance with respect to *x, y* displacement), its location may be quite critical and an error at this stage often leads to a failure of subsequent processing (e.g., matching). On the other hand, if the core has to be used only for fingerprint registration, it is not important to find the north most loop exactly and any stable point in the fingerprint pattern is suitable.

One of the first automatic methods for fingerprint registration was known as R92 [1], it searches for a core point independently of the other singularities. The core is searched by scanning (row by row) the orientation image to find *wellformed arches*; a well-formed arch is denoted by a sextet (set of six) of adjacent elements whose orientations comply with several rules controlled by many parameters. One sextet is chosen among the valid sextets by evaluating the orientation of the elements in adjacent rows. The exact core position is then located through interpolation (Figure 2.10). Even though R92 is quite complicated and heuristic in nature, it usually gives good results and is able to localize the core point with sub-block accuracy. This algorithm was a fundamental component of the fingerprint identification systems used by the FBI and is still extensively used by other authors.

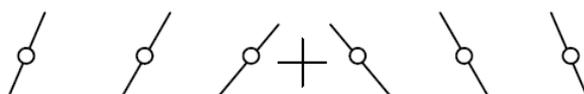

*Figure 2.10. The core point "+" located on the chosen sextet.*

Several other ideas for the location of stable registration points have been proposed. In [14], the *focal point* is defined as the point where pairs of straight lines normal to the ridges intersect. Because the ridges do not draw perfect concentric circumferences around the core, the normal lines (Figure 2.11) do not exactly cross at a single point and a sort of average point has to be defined as the center of curvature. In [14], this average point is computed as the barycenter of the crossing between pairs of normals.

Although the focal point (or the center of curvature) does not necessarily correspond to the core point, it has been experimentally demonstrated to be quite stable with respect to fingerprint variation (displacement, rotation, distortion, etc.). Therefore, it can be reliably used for fingerprint registration. The main problem of these methods is in isolating a fingerprint region characterized by a single center of curvature. In fact, if the selected fingerprint region contains more than one singularity, the result may be unpredictable.

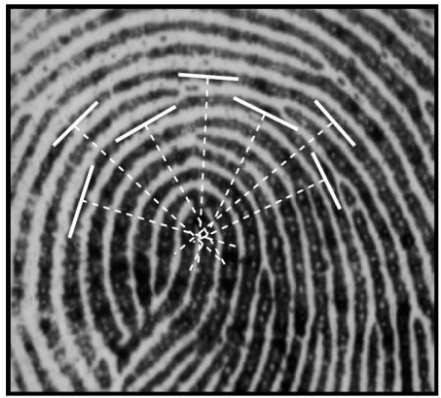

*Figure 2.11. The straight lines normal to the ridges identify a valid registration point that corresponds to the center of curvature.*

In [15], a multi-resolution approach is proposed for locating the north most loop type singularity (core) based on integration of sine components in two adjacent regions $R_I$ and $R_{II}$ (Figure 2.12). The geometry of the two regions is designed to capture the maximum curvature in concave ridges. At each scale and for each candidate position [*x, y*], the sine components of the orientation image are integrated over the two regions resulting in the values $SR_I$ and $SR_{II}$. The points [$x$, $y$] that maximize the quantity ($SR_I - SR_{II}$) are retained as candidate positions and analyzed at a finer resolution.

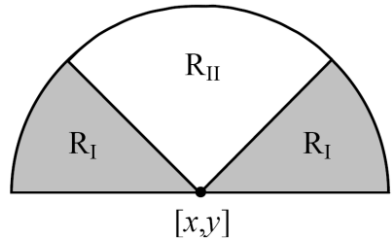

*Figure 2.12. Regions of integration of the sine components in the method proposed in [15]*

Other approaches for singularity or core detection, besides those already described, have been proposed in [16], [17], [8], [18], [19], and [20]. Appendix A shows a MATLAB source code for core point detection which is used in the proposed matching algorithm described in Chapter 4. The idea of determining the core point is taken from [21], which is described in section 4.3.1.2.

## 2.1.5.    Enhancement

The performance of minutiae extraction algorithms and other fingerprint recognition techniques relies heavily on the quality of the input fingerprint images. In an ideal fingerprint image, ridges and valleys alternate and flow in a locally constant direction. In such situations, the ridges can be easily detected and minutiae can be precisely located in the image. However, in practice, due to skin conditions (e.g., wet or dry, cuts, and bruises), sensor noise, incorrect finger pressure, and inherently low-quality fingers (e.g., elderly people, manual workers), a significant percentage of fingerprint images (approximately 10%) is of poor quality [22].

In many cases, a single fingerprint image contains regions of good, medium, and poor quality where the ridge pattern is very noisy and corrupted. In general, there are several types of degradation associated with fingerprint images:

1. the ridges are not strictly continuous.
2. parallel ridges are not well separated.
3. cuts, creases, and braises.

These three types of degradation make ridge extraction extremely difficult in the highly corrupted regions. This leads to the following problems in minutiae extraction:

i) a significant number of spurious minutiae are extracted,
ii) a large number of genuine minutiae are missed, and
iii) large errors in the location (position and orientation) of minutiae are introduced.

In order to ensure good performance of the ridge and minutiae extraction algorithms in poor quality fingerprint images, an enhancement algorithm to improve the clarity of the ridge structure is necessary.

Generally, for a given fingerprint image, the fingerprint areas resulting from the segmentation step may be divided into three categories:

• Well-defined region, where ridges are clearly differentiated from each another;
• Recoverable region, where ridges are corrupted by a small amount of gaps, creases, smudges, links, and the like, but they are still visible and the neighboring regions provide sufficient information about their true structure;
• Unrecoverable region, where ridges are corrupted by such a severe amount of noise and distortion that no ridges are visible and the neighboring regions do not allow them to be reconstructed.

Good quality regions, recoverable, and unrecoverable regions may be identified according to several criteria; in general, image contrast, orientation consistency, ridge frequency, and other local features may be combined to define a quality index. The goal of an enhancement algorithm is to improve the clarity of the ridge structures in the recoverable regions and mark the unrecoverable regions as too noisy for further processing.

Usually, the input of the enhancement algorithm is a gray-scale image. The output may either be a gray-scale or a binary image, depending on the algorithm. General-purpose image enhancement techniques do not produce satisfying and definitive results for fingerprint image enhancement. However, contrast stretching, histogram manipulation, normalization [9], and Wiener filtering [23] have been shown to be effective as initial processing steps in a more sophisticated fingerprint enhancement algorithm.

The most widely used technique for fingerprint image enhancement is based on *contextual* filters. In conventional image filtering, only a single filter is used for convolution throughout the image. In contextual filtering, the filter characteristics change according to the local context. Usually, a set of filters is pre-computed and one of them is selected for each image region. In fingerprint enhancement, the context is often defined by the local ridge orientation and local ridge frequency. In fact, the sinusoidal-shaped wave of ridges and valleys is mainly defined by a local orientation and frequency that varies slowly across the fingerprint area. An appropriate filter that is tuned to the local ridge frequency and orientation can efficiently remove the undesired noise and preserve the true ridge and valley structure.

Several types of contextual filters have been proposed in the literature for fingerprint enhancement. Although they have different definitions, the intended behavior is almost the same:

1) provide a low-pass (averaging) effect along the ridge direction with the aim of linking small gaps and filling impurities due to pores or noise;

2) perform a bandpass (differentiating) effect in the direction orthogonal to the ridges to increase the discrimination between ridges and valleys and to separate parallel linked ridges.

An effective method based on *Gabor* filters is proposed in [9]. Gabor filters have both frequency-selective and orientation-selective properties and have optimal joint resolution in both spatial and frequency domains [24]. The even symmetric two-dimensional Gabor filter has the following form.

$$g(x,y:\theta,f) = \exp\left\{-\frac{1}{2}\left[\frac{x_\theta^2}{\sigma_x^2} + \frac{y_\theta^2}{\sigma_y^2}\right]\right\}.\cos(2\pi f.x_\theta) \qquad (2.2)$$

where $\theta$ is the orientation of the filter, and $[x_\theta, y_\theta]$ are the coordinates of $[x, y]$ after a clockwise rotation of the Cartesian axes by an angle of $(90°-\theta)$.

$$\begin{bmatrix} x_\theta \\ y_\theta \end{bmatrix} = \begin{bmatrix} \cos(90°-\theta) & \sin(90°-\theta) \\ -\sin(90°-\theta) & \cos(90°-\theta) \end{bmatrix}\begin{bmatrix} x \\ y \end{bmatrix} = \begin{bmatrix} \sin\theta & \cos\theta \\ -\cos\theta & \sin\theta \end{bmatrix}\begin{bmatrix} x \\ y \end{bmatrix} \qquad (2.3)$$

In the above expressions, $f$ is the frequency of a sinusoidal plane wave, and $\sigma_x$ and $\sigma_y$ are the standard deviations of the Gaussian envelope along the x- and y-axes, respectively.

To apply Gabor filters to an image, the four parameters $(\theta, f, \sigma_x, \sigma_y)$ must be specified. Obviously, the frequency of the filter is completely determined by the local ridge frequency and the orientation is determined by the local ridge orientation. The selection of the values $\sigma_x$ and $\sigma_y$ involves a tradeoff. The larger the values, the more robust the filters are to the noise in the fingerprint image, but also the more likely to create spurious ridges and valleys. On the other hand, the smaller the values, the less likely the filters are to introduce spurious ridges and valleys but then they will be less effective in removing the noise. In fact, from the Modulation Transfer Function (MTF) of the Gabor filter, it can be shown that increasing $\sigma_x$, $\sigma_y$ decreases the bandwidth of the filter and vice versa. Based on empirical data, $\sigma_x$ and $\sigma_y$ are set to 4 in [9].

The output of a contextual fingerprint enhancement can be a gray-scale, near-binary, or binary image, usually depending on the filter parameters chosen. When selecting the appropriate set of filters and tuning their parameters, one should keep in mind that the goal is not to produce a good visual appearance of the image but facilitate robustness of the successive feature extraction steps. If the filters are tuned to strongly increase the contrast and suppress the noise, the estimation of the local context (orientation and frequency) may be erroneous in poor quality areas and the filtering is likely to produce spurious structures [25].

The need of an effective enhancement is particularly important in poor quality fingerprints where only the recoverable regions carry information necessary for the matching. On the other hand, computing local information (context) with sufficient reliability in poor quality fingerprint images is very challenging and the risk is to make things worse.

## 2.1.6.        Binarization and Minutiae Extraction

Although rather different from one another, most of the minutiae extraction methods require the fingerprint gray-scale image to be converted into a binary image. Some binarization processes greatly benefit from an a priori enhancement; on the other hand, some enhancement algorithms directly produce a binary output, and therefore the distinction between enhancement and binarization is often faded. The binary images obtained by the binarization process are usually submitted to a thinning stage which allows for the ridge line thickness to be reduced to one pixel.

Finally, a simple image scan allows the detection of pixels that correspond to minutiae through the pixel-wise computation of crossing number [26] (see Figure 2.13).

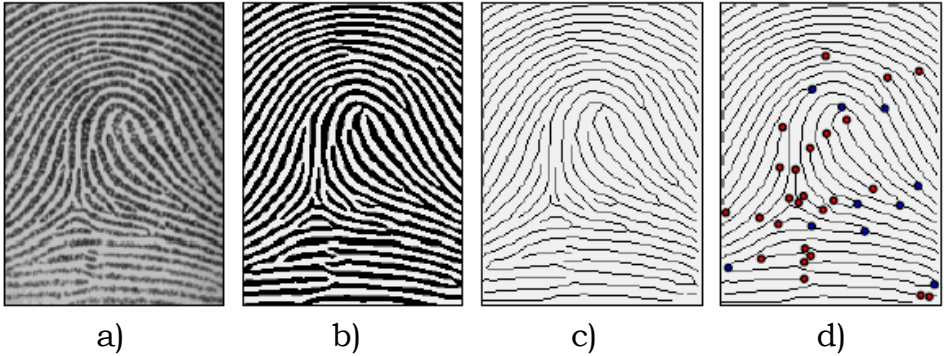

a)              b)              c)              d)

*Figure 2.13. a) A fingerprint gray-scale image; b) the image obtained after enhancement and binarization; c) the image obtained after thinning; d) termination and bifurcation minutiae detected through the pixel-wise computation of the crossing number.*

The general problem of image binarization has been widely studied in the fields of image processing and pattern recognition. The easiest approach uses a global threshold $t$ and works by setting the pixels whose gray-level is lower than $t$ to 0 and the remaining pixels to 1. In general, different portions of an image may be characterized by different contrast and intensity and, consequently, a single threshold is not sufficient for a correct binarization. For this reason, the local threshold technique changes $t$ locally, by adapting its value to the average local intensity. In the specific case of fingerprint images, which are sometimes of very poor quality, a local threshold method cannot always guarantee acceptable results and more effective fingerprint-specific solutions are necessary.

In [27], the authors introduced a binarization approach based on peak detection in the gray-level profiles along sections orthogonal to the ridge orientation (see Figure 2.14). A 16 × 16 oriented window is centered around each pixel [x, y]. The gray-level profile is obtained by projection of the pixel intensities onto the central section. The profile is smoothed through local averaging; the peaks and the two neighboring pixels on either side of each peak constitute the foreground of the resulting binary image.

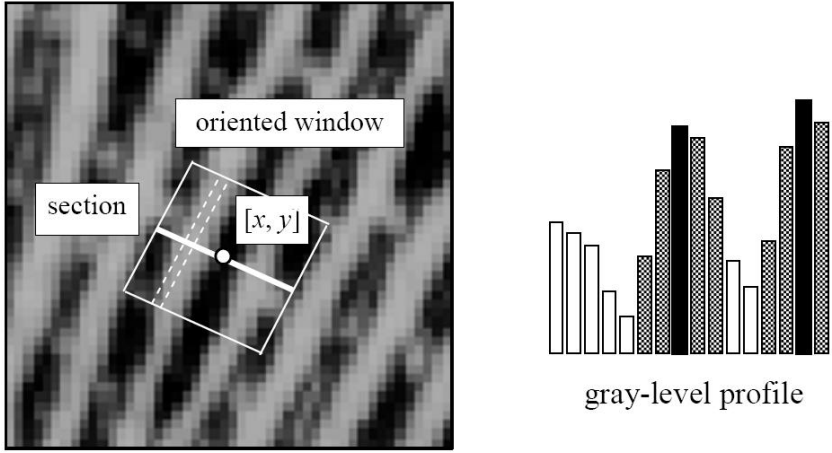

*Figure 2.14. An example of a gray-level profile obtained through projection of pixel intensities on the segment centered at [x, y] and normal to the local ridge orientation $\theta_{xy}$.*

Most of the enhancement algorithms based on contextual filtering (discussed in Section 2.1.5) may produce a clear binary image for appropriately chosen parameters. In any case, even when the output of the contextual filtering is a gray-scale image, a simple local thresholding technique often results in satisfactory binarization.

Minutiae detection from binary images is usually performed after an intermediate *thinning* step that reduces the width of the ridges to one pixel. Unfortunately, thinning algorithms are rather critical and the aberrations and irregularity of the binary-ridge boundaries have an adverse effect on the skeletons (i.e., the one-pixel-width ridge structure), resulting in "hairy" growths (spikes) that lead to the detection of spurious minutiae.

With the aim of improving the quality of the binary images, some researchers have introduced regularization techniques which usually work by filling holes, removing small breaks, eliminating bridges between ridges, and other artifacts. To remove the spikes that often characterize the thinned binary images, the authors of [27] implement a morphological "open" operator [5] whose structuring element is a small box oriented according to the local ridge orientation.

As far as thinning techniques are concerned [28], a large number of approaches are available in the literature due to the central role of this processing step in many pattern recognition applications: character recognition, document analysis, map and drawing vectorization, and so on [26].

Once a binary skeleton has been obtained, a simple image scan allows the pixels corresponding to minutiae to be detected. In fact the pixels corresponding to minutiae are characterized by a *crossing number* different from the value 2. The crossing number $cn(p)$ of a pixel p in a binary image is defined [26] as half the sum of the differences between pairs of adjacent pixels in the 8-neighborhood of p:

$$cn(p) = \frac{1}{2} \sum_{i=1\ldots8} \left| val(p_{i\,\mathrm{mod}\,8}) - val(p_{i-1}) \right|, \qquad (2.4)$$

where $p_0$, $p_1$, ...$p_7$ are the pixels belonging to an ordered sequence of pixels defining the eight neighborhood of p and $val(p) \in \{0, 1\}$ is the pixel value. It is simple to note (Figure 2.15) that a pixel p with $val(p) = 1$:

- Is an intermediate ridge point if $cn(p) = 2$.
- Corresponds to a ridge ending minutia if $cn(p) = 1$.
- Corresponds to a bifurcation minutia if $cn(p) = 3$.
- Defines a more complex minutia (e.g., crossover) if $cn(p) \geq 3$.

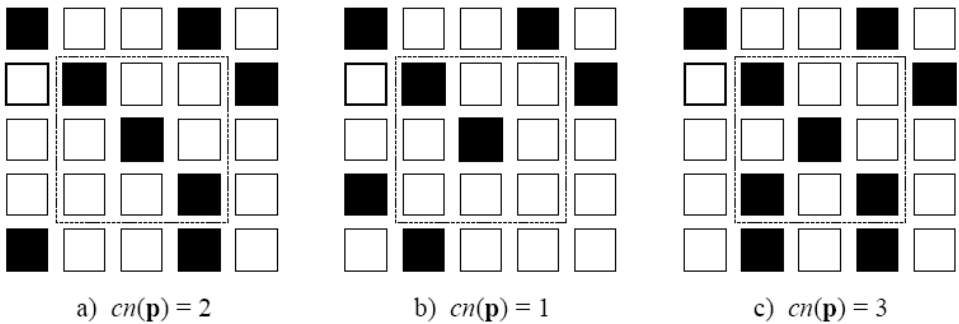

a) $cn(\mathbf{p}) = 2$       b) $cn(\mathbf{p}) = 1$       c) $cn(\mathbf{p}) = 3$

*Figure 2.15. a) intra-ridge pixel; b) ridge ending minutia;*
*c) bifurcation minutia.*

Appendix B shows a MATLAB source code for the Minutiae extraction process described above, which is used in the proposed matching algorithm described in Chapter 4.

Some authors have proposed minutiae extraction approaches that work directly on the gray-scale images without binarization and thinning [12]. The basic idea is to track the ridge lines in the gray-scale image, by "sailing" according to the local orientation of the ridge pattern. The ridge line extraction algorithm attempts to locate, at each step, a local maximum relative to a section orthogonal to the ridge direction. By connecting the consecutive maxima, a polygonal approximation of the ridge line can be obtained.

### 2.1.7.        Minutiae Filtering

A post-processing stage (called minutiae filtering) is often useful in removing the spurious minutiae detected in highly corrupted regions or introduced by previous processing steps (e.g., thinning). In [26], two main post-processing types have been proposed: structural post-processing, and minutiae filtering in the gray-scale domain.

- Structural post-processing

Simple structural rules may be used to detect many of the false minutiae that usually affect thinned binary fingerprint images. In [29], the authors identified the most common false minutiae structures and introduced an ad hoc approach to remove them (Figure 2.16). The underlying algorithm is rule-based and requires some numerical characteristics associated with the minutiae as input:

- The length of the associated ridge(s),
- The minutia angle, and
- The number of facing minutiae in a neighborhood.

As shown in Figure 2.16, the algorithm connects facing endpoints (a, b), removes bifurcations facing with endpoints (c) or with other bifurcations (d), and removes spurs (e), bridges (f), triangles (g), and ladder structures (h). This minutiae filtering method is used in the MATLAB source code of Appendix B.

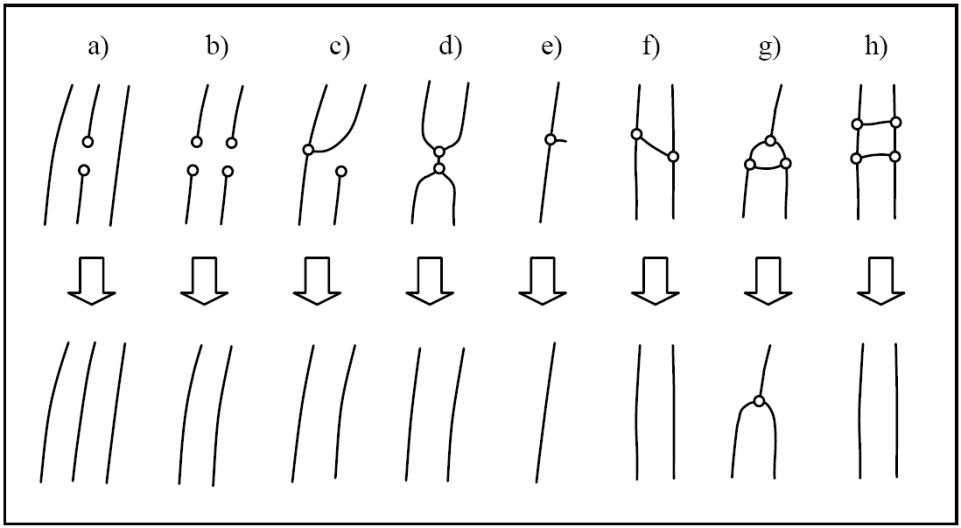

*Figure 2.16. The most common false-minutiae structures (on the top row) and the structural changes resulting from their removal (bottom row).*

• Minutiae filtering in the gray-scale domain

A direct gray-scale minutiae filtering technique reexamines the gray-scale image in the spatial neighborhood of a detected minutia with the aim of verifying the presence of a real minutia.

Maio and Maltoni used a shared-weights neural network to verify the minutiae detected by their gray-scale algorithm [12]. The minutiae neighborhoods in the original gray-scale image are normalized, with respect to their angle and the local ridge frequency, before passing them to a neural network classifier, which classifies them as termination, bifurcation, and non-minutia.

## 2.2.  Fingerprint Verification Competition FVC

The database (Fingerprint Verification Competition) FVC2000 [30] is used during experiments. FVC consists of four different databases (DB1, DB2, DB3 and DB4) which were collected by using the following sensors/technologies:

- DB1: low-cost optical sensor "Secure Desktop Scanner" by KeyTronic
- DB2: low-cost capacitive sensor "TouchChip" by ST Microelectronics
- DB3: optical sensor "DF-90" by Identicator Technology
- DB4: synthetic fingerprint generation.

Each database is 110 fingers wide (w) and 8 impressions per finger deep (d) (see Table 2.1) (880 fingerprints in all); fingers from 101 to 110 (set B) have been made available to the participants to allow parameter tuning before the submission of the algorithms; the benchmark is then constituted by fingers numbered from 1 to 100 (set A).

|      | Image Size | Set A (w×d) | Set B (w×d) | Resolution |
|------|------------|-------------|-------------|------------|
| **DB1** | 300×300 | 100×8 | 10×8 | 500 dpi |
| **DB2** | 256×364 | 100×8 | 10×8 | 500 dpi |
| **DB3** | 448×478 | 100×8 | 10×8 | 500 dpi |
| **DB4** | 240×320 | 100×8 | 10×8 | about 500 dpi |

*Table 2.1. Specifications of each database [30].*

Figure 2.17 shows a sample image from each database:

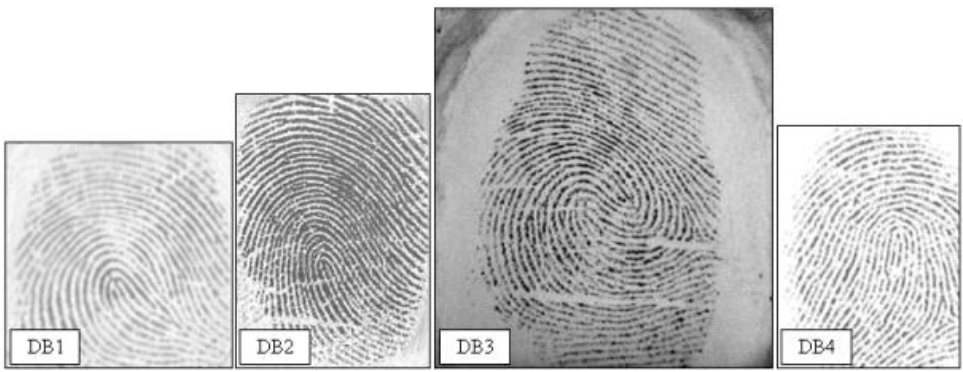

*Figure 2.17. Sample images taken from DB1, DB2, DB3, and DB4. In order to show the different image sizes of each database, the four images are displayed at the same scale factor [30].*

## 2.3. Fingerprint Matching

A fingerprint matching algorithm compares two given fingerprints and returns either a degree of similarity (without loss of generality, a score between 0 and 1) or a binary decision (mated/non-mated). Only a few matching algorithms operate directly on grayscale fingerprint images; most of them require that an intermediate fingerprint representation be derived through a feature extraction stage[26].

The fingerprint feature extraction and matching algorithms are usually quite similar for both fingerprint verification and identification problems. This is because the fingerprint identification problem (i.e., searching for an input fingerprint in a database of $N$ fingerprints) can be implemented as a sequential execution of $N$ one-to-one comparisons (verifications) between pairs of fingerprints. The fingerprint classification and indexing techniques are usually exploited to speed up the search in fingerprint identification problems [1].

Matching fingerprint images is an extremely difficult problem, mainly due to the large variability in different impressions of the same finger (i.e., large intra-class variations). The main factors responsible for intra-class variations are:

- Displacement,
- Rotation,
- Partial overlap,
- Non-linear distortion,
- Pressure and skin condition,
- Noise, and
- Feature extraction errors.

A large number of automatic fingerprint matching algorithms have been proposed in the pattern recognition literature. Most of these algorithms have no difficulty in matching good quality fingerprint images. However, fingerprint matching remains a challenging pattern recognition problem to date due to the difficulty in matching low-quality and partial fingerprints.

The large number of approaches to fingerprint matching can be coarsely classified into three families.

- Correlation-based matching: Two fingerprint images are superimposed and the correlation between corresponding pixels is computed for different alignments (e.g., various displacements and rotations).
- Minutiae-based matching: This is the most popular and widely used technique, being the basis of the fingerprint comparison made by fingerprint examiners. Minutiae are extracted from the two fingerprints and stored as sets of points in the two-dimensional plane. Minutiae-based matching essentially consists of finding the alignment between the template and the input minutiae sets that results in the maximum number of minutiae pairings.
- Ridge feature-based matching: Minutiae extraction is difficult in very low-quality fingerprint images. However, whereas other features of the fingerprint ridge pattern (e.g., local orientation and frequency, ridge shape, texture information) may be extracted more reliably than minutiae, their distinctiveness is generally lower. The approaches belonging to this family compare fingerprints in term of features extracted from the ridge pattern.

### 2.3.1.      Correlation-Based Techniques

Let T and I be the two fingerprint images corresponding to the template and the input fingerprint, respectively. Then an intuitive measure of their diversity is the sum of squared differences (SSD) between the intensities of the corresponding pixels:

$$SSD(T,I) = \|T-I\|^2 = (T-I)^T(T-I) = \|T\|^2 + \|I\|^2 - 2T^TI \qquad (2.5)$$

where the superscript "T" denotes the transpose of a vector. If the terms $\|T\|^2$ and $\|I\|^2$ are constant, the diversity between the two images is minimized when the cross-correlation (CC) between T and I is maximized:

$$CC(T,I) = T^T I \qquad (2.6)$$

Note that the quantity (-2.CC(T,I)) appears as the third term in Equation (2.5). The cross-correlation (or simply correlation) is then a measure of image similarity. Due to the displacement and rotation that unavoidably characterize two impressions of a given finger, their similarity cannot be simply computed by superimposing T and I and applying Equation (2.6).

Let $I^{(\Delta x, \Delta y, \theta)}$ represent a rotation of the input image I by an angle $\theta$ around the origin (usually the image center) and shifted by $\Delta x$, $\Delta y$ pixels in directions x and y, respectively; then the similarity between the two fingerprint images T and I can be measured as

$$S(T,I) = \max_{\Delta x, \Delta y, \theta} CC(T, I^{(\Delta x, \Delta y, \theta)}) \qquad (2.7)$$

The direct application of Equation (2.7) rarely leads to acceptable results mainly due to the following problems:

1. Non-linear distortion makes impressions of the same finger significantly different in terms of global structure,
2. Skin condition and finger pressure cause image brightness, contrast, and ridge thickness to vary significantly across different impressions, and
3. A direct application of Equation (2.7) is computationally very expensive.

The fingerprint distortion problem (point 1 in the above list) is usually addressed by computing the correlation locally instead of globally.

As to the computational complexity of the correlation technique, smart approaches may be exploited to achieve efficient implementations.

## 2.3.2.      Minutiae-based Methods

Minutiae matching is certainly the most well-known and widely used method for fingerprint matching, thanks to its strict analogy with the way forensic experts compare fingerprints and its acceptance as a proof of identity in the courts of law in almost all countries.

### 2.3.2.1.      Problem formulation

Let T and I be the representation of the template and input fingerprint, respectively. Unlike in correlation-based techniques, where the fingerprint representation coincides with the fingerprint image, here the representation is a feature vector (of variable length) whose elements are the fingerprint minutiae. Each minutia may be described by a number of attributes, including its location in the fingerprint image, orientation, type (e.g., ridge termination or ridge bifurcation), a weight based on the quality of the fingerprint image in the neighborhood of the minutia, and so on. Most common minutiae matching algorithms consider each minutia as a triplet $m = \{x, y, \theta\}$ that indicates the x,y minutia location coordinates and the minutia angle $\theta$:

$$T = \{m_1, m_2, \ldots, m_m\}, \ m_i = \{x_i, y_i, \ldots, \theta_i\} \ i = 1..m$$
$$I = \{m'_1, m'_2, \ldots, m'_m\}, \ m'_i = \{x'_i, y'_i, \ldots, \theta'_i\} \ j = 1..n$$

where m and n denote the number of minutiae in T and I, respectively.

A minutia $m'_j$ in I and a minutia $m'_i$ in T are considered "matching," if the spatial distance (sd) between them is smaller than a given tolerance $r_0$ and the direction difference (dd) between them is smaller than an angular tolerance $\theta_0$:

$$sd(m'_j, m_i) = \sqrt{(x'_j - x_i)^2 + (y'_j - y_i)^2} \leq r_0, \quad and \qquad (2.8)$$

$$dd(m'_j, m_i) = \min\left(\left|\theta'_j - \theta_i\right|, 360° - \left|\theta'_j - \theta_i\right|\right) \leq \theta_0 \qquad (2.9)$$

Equation (2.9) takes the minimum of $|\theta'_j - \theta_i|$ and $360°$ $-|\theta'_j - \theta_i|$ because of the circularity of angles (the difference between angles of 2° and 358° is only 4°). The tolerance boxes (or hyper-spheres) defined by $r_0$ and $\theta_0$ are necessary to compensate for the unavoidable errors made by feature extraction algorithms and to account for the small plastic distortions that cause the minutiae positions to change.

Aligning the two fingerprints is a mandatory step in order to maximize the number of matching minutiae. Correctly aligning two fingerprints certainly requires displacement (in x and y) and rotation (θ) to be recovered, and likely involves other geometrical transformations:

- Scale has to be considered when the resolution of the two fingerprints may vary (e.g., the two fingerprint images have been taken by scanners operating at different resolutions);
- Other distortion-tolerant geometrical transformations could be useful to match minutiae in case one or both of the fingerprints is affected by severe distortions.

Let map(.) be the function that maps a minutia $m'_j$ (from I) into $m''_j$ according to a given geometrical transformation; for example, by considering a displacement of [Δx, Δy] and a counterclockwise rotation around the origin:

$$map_{\Delta x, \Delta y, \Delta \theta}(m'_j = \{x'_j, y'_j, ..., \theta'_j\}) = m''_j = \{x''_j, y''_j, ..., \theta'_j + \theta\}, \quad \text{(2.10)}$$

where

$$\begin{bmatrix} x''_j \\ y''_j \end{bmatrix} = \begin{bmatrix} \cos\theta & -\sin\theta \\ \sin\theta & \cos\theta \end{bmatrix} \begin{bmatrix} x'_j \\ y'_j \end{bmatrix} + \begin{bmatrix} \Delta x \\ \Delta y \end{bmatrix}. \quad \text{(2.11)}$$

Let mm(.) be an indicator function that returns 1 in the case where the minutiae $m''_j$ and $m_i$ match according to Equations (2.8) and (2.9):

$$mm(m''_j, m_i) = \begin{cases} 1 & sd(m''_j, m_i) \le r_0 \quad and \quad dd(m''_j, m_i) \le \theta_0 \\ 0 & otherwise \end{cases} \quad \text{(2.12)}$$

Then, the matching problem can be formulated as

$$\underset{\Delta x, \Delta y, \theta, P}{\text{maximise}} \sum_{i=1}^{m} mm(map_{\Delta x, \Delta y, \theta}(m'_{P(i)}), m_i) \qquad\qquad 2.13)$$

where P(i) is an unknown function that determines the pairing between I and T minutiae; in particular, each minutia has either exactly one mate in the other fingerprint or has no mate at all:

1. P(i) =j indicates that the mate of the $m_i$ in T is the minutia $m'_j$ in I;
2. P(i) = null indicates that minutia $m_i$ in T has no mate in I;
3. a minutia $m'_j$ in I, such that $\forall i = 1..m$, P(i) ≠ j has no mate in T;
4. $\forall i = 1..m$, k=1..m, i ≠ k⟹ P(i) ≠ P(k) or P(i) = P(k) = null (this requires that each minutia in I is associated with a maximum of one minutia in T).

Note that, in general, P(i) =j does not necessarily mean that minutiae $m'_j$ and $m_i$, match in the sense of Equations (2.8) and (2.9) but only that they are the most likely pair under the current transformation.

Solving the minutiae matching problem (Equation (2.13)) is trivial when the correct alignment (Δx, Δy, θ) is known; in fact, the pairing (i.e., the function P) can be determined by setting for each i = 1..m:

• P(i)=j if $m''_j = map_{\Delta x, \Delta y, \theta}(m'_j)$ is closest to $m_i$ among the minutiae:
    {$m''_k = map_{\Delta x, \Delta y, \theta}(m'_k)$ | k=1..n, mm($m''_k$, $m_i$)=1};
• P(i) = null if $\forall$ k=1..n, mm($map_{\Delta x, \Delta y, \theta}(m'_k), m_i$)=0

To comply with constraint 4 above, each minutia $m''_j$ already mated has to be marked, to avoid mating it twice or more.

To achieve the optimum pairing (according to Equation (2.13)), a slightly more complicated scheme should be adopted: in fact, in the case when a minutia of I falls within the tolerance hyper-sphere of more than one minutia of T, the optimum assignment is that which maximizes the number of mates.

The maximization in Equation (2.13) can be easily solved if the function P (minutiae correspondence) is known; in this case, the unknown alignment ($\Delta x$, $\Delta y$, $\theta$) can be determined in the least square sense [31] and [32].

In the pattern recognition literature, the minutiae matching problem has been generally addressed as a point pattern matching problem. Even though a small difference exists due to the presence of a direction associated with each minutia point, the two problems may be approached analogously. Because of its central role in many pattern recognition and computer vision tasks (e.g., object matching, remote sensing, camera calibration, motion estimation), point pattern matching has been extensively studied yielding families of approaches known as relaxation methods, algebraic and operational research solutions, tree-pruning approaches, energy-minimization methods, Hough transform, and so on.

### 2.3.2.2.    Minutiae matching with pre-alignment

Embedding fingerprint alignment into the minutiae matching stage, certainly leads to the design of robust algorithms, which are often able to operate with noisy and incomplete data. On the other hand, the computational complexity of such methods does not provide a high matching throughput (e.g., 10,000 matches per second), as required by AFIS or civil systems.

Storing pre-aligned templates in the database and pre-aligning the input fingerprint before the minutiae matching can be a valid solution to speed up the 1 :N identification. In theory, if a perfect pre-alignment could be achieved, the minutiae matching could be reduced to a simple pairing. Two main approaches for pre-alignment have been investigated.

Absolute pre-alignment: Each fingerprint template is pre-aligned, independently of the others, before storing it in the database. Matching an input fingerprint I with a set of templates requires I to be independently registered just once, and the resulting aligned representation to be matched with all the templates.

Relative pre-alignment: The input fingerprint I has to be pre-aligned with respect to each template T in the database; l:N identification requires N independent pre-alignments. Relative pre-alignment may determine a significant speed up with respect to the algorithms that do not perform any pre-alignment, but cannot compete in terms of efficiency with absolute pre-alignment. However, relative pre-alignment is in general more effective (in terms of accuracy) than absolute pre-alignment, because the features of the template T may be used to drive the registration process.

Relative pre-alignment may be performed in several ways; for example:
- by superimposing the singularities,
- by correlating the orientation images (template matching), or
- by correlating ridge features (e.g., length and orientation of the ridges).

### 2.3.2.3.    Avoiding alignment

Fingerprint alignment is certainly a critical and time-consuming step. To overcome problems involved in alignment, and to better cope with local distortions, some authors perform minutiae matching locally. A few other attempts have been proposed that try to globally match minutiae without requiring explicit recovery of the parameters of the transformation. In [33], an intrinsic coordinate system (ICS) is introduced whose axes run along hypothetical lines defined by the local orientation of the fingerprint pattern. First, the fingerprint is partitioned in regular regions (i.e., regions that do not contain singular points). In each regular region, the ICS is defined by the orientation field.

When using intrinsic coordinates instead of pixel coordinates, minutiae are defined with respect to their position in the orientation field. Translation, displacement, and distortion move minutiae with the orientation field they are immersed in and therefore do not change their intrinsic coordinates. On the other hand, some practical problems such as reliably partitioning the fingerprint in regular regions and unambiguously defining intrinsic coordinate axes in low-quality fingerprints still remain to be solved.

## 2.3.2.4.    Global versus Local Minutiae Matching

Local minutiae matching consists of comparing two fingerprints according to local minutiae structures; local structures are characterized by attributes that are invariant with respect to global transformation (e.g., translation, rotation, etc.) and therefore are suitable for matching without any a priori global alignment. Matching fingerprints based only on local minutiae arrangements relaxes global spatial relationships which are highly distinctive and therefore reduce the amount of information available for discriminating fingerprints. Global versus local matching is a tradeoff among simplicity, low computational complexity, and high distortion-tolerance (local matching), and high distinctiveness on the other hand (global matching). Interesting approaches have been proposed where the advantages of both methods are exploited ([34] and [35]), a fast local matching is initially carried out for recovering alignment or an early rejection of very different fingerprints; a consolidation step is then performed to check whether the postulated coincidence hypotheses hold at global level.

## 2.3.3.    Ridge Feature-based Matching Techniques

Three main reasons induce designers of fingerprint recognition techniques to search for other fingerprint distinguishing features, beyond minutiae:

- reliably extracting minutiae from poor quality fingerprints is very difficult. Although minutiae may carry most of the fingerprint discriminatory information, they do not always constitute the best tradeoff between accuracy and robustness;
- minutiae extraction is time consuming. This was a serious problem in the past, when the computational power of desktop computers was limited. Although computers are much faster now, matching speed remains a concern because of the increasing interest in embedding fingerprint recognition algorithms in low-cost, standalone (embedded) systems. Furthermore, registering minutiae representation is very challenging; minutiae-based matching algorithms are rather slow, especially for large-scale identification tasks;

Additional features may be used in conjunction with minutiae (and not as an alternative) to increase system accuracy and robustness.

The more commonly used alternative features are:
1. size of the fingerprint and shape of the external fingerprint silhouette;
2. number, type, and position of singularities;
3. spatial relationship and geometrical attributes of the ridge lines;
4. shape features;
5. global and local texture information;
6. sweat pores;
7. fractal features.

Features listed in 1) and 2) above are, in general, very unstable, and they vary depending on which part of the finger touches the sensor. Sweat pores are undoubtedly highly discriminant but their detection requires high-resolution expensive scanners.

Local texture analysis has proved to be more effective than global feature analysis; although most of the local texture information is carried by the orientation and frequency images, most of the proposed approaches extract texture by using a specialized bank of filters.

### 2.3.4.    Minutiae vs. pattern based fingerprint templates

In the following two subsections, the difference between pattern-based and minutiae-based fingerprint templates is summarized [36].

### 2.3.4.1.    Pattern-Based Templates

A capture device is used to take a graphical image of a fingerprint, typically captured as a TIFF (Tagged Image File Format) image. The graphical image obtained from the capture device is commonly referred to as a live scan to distinguish it from a template or print stored in a database. Processing software examines the fingerprint image and locates the image center, which may be off-center from the fingerprint core. The image is then cropped a fixed distance around this graphical center. The rectangle in Figure 2.18 details this cropped region. The cropped region is then compressed and stored for subsequent matches.

Fingerprint matching with pattern-based templates involves making a graphical comparison of the two templates and determining a measure of the difference. The greater the difference the less likely the prints match.

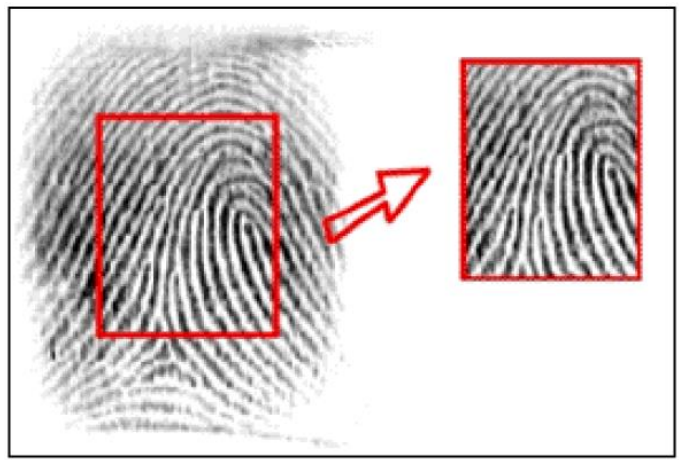

*Figure 2.18. Pattern-based template*

## 2.3.4.2.    Minutia-based Templates

As in a pattern-based system, a capture device is used to take a graphical image of a fingerprint (live scan). Special software then analyzes the fingerprint image and determines if the image actually contains a fingerprint, determines the location of the core, the pattern type (e.g. right loop, left arch, etc.), estimates the quality of the ridge lines, and finally extracts minutia. Minutiae, from a simple perspective, indicate where a significant change in the fingerprint occurs. These changes are shown in Figure 2.19.

Understanding that dark lines in the image represent ridges and light lines represent valleys, Arrow A shows a region where one ridge splits into two ridges (called a bifurcation) and Arrow B shows where a ridge ends.

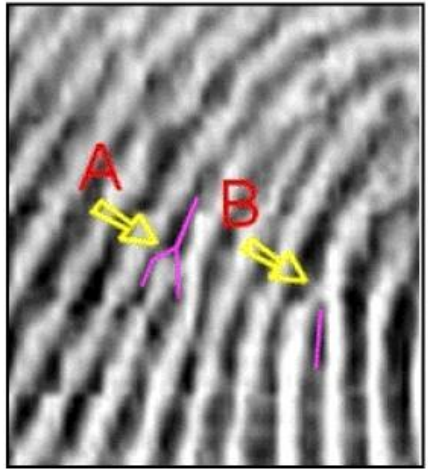

*Figure 2.19. Fingerprint Changes*

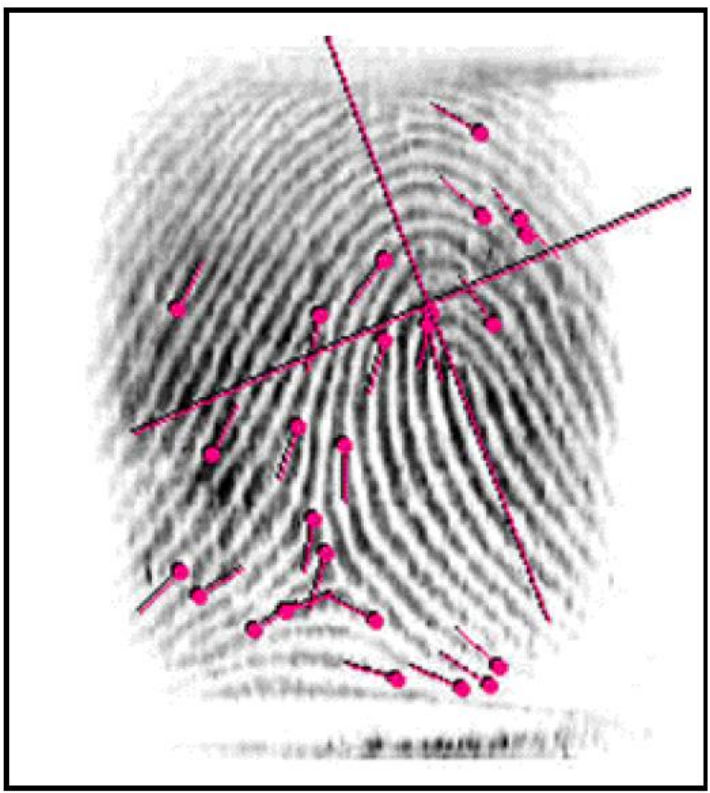

*Figure 2.20. Extracted minutiae and axes*

After locating these features in the fingerprint, the minutia extraction software determines a significant direction of the change (using Arrow B as an example, the significant direction starts at the end of the ridge and moves downward. The resultant minutia, in their simplest form are then the collection of all reasonable bifurcations and ridge endings, their location and their significant direction. Minutiae are also assigned a measure of their strength. A set of minutiae is shown in Figure 2.20.

In Table 2.2, a technical comparison is shown between minutia and pattern based fingerprint templates.

|  | Minutia | Pattern |
|---|---|---|
| Definition | Analyzes the points at which the ridges on the | Graphical comparison of fingerprint image. |

|  | fingerprints split, intersect or end. |  |
|---|---|---|
| How it works | Capture device analyzes the fingerprint image to determine the location of the fingerprint core, the pattern type (i.e., right loop, left arch, etc.), estimates the quality of the ridgelines and extracts the points in which the ridges split, intersect or end. These points are called minutiae. | Graphical center of fingerprint image (not necessarily defined by the fingerprint core) is cropped a fixed distance and compressed for subsequent match. The greater the difference between the stored template and the live comparison, the less likely the match. |
| Template size | • Small template size (can be as small as 120 bytes; average size is 350 bytes) • Template size can be controlled by specifying the number of minutia to be analyzed. | • Relatively large template size (500-700 bytes when compressed) • Cannot easily control template size without compromising accuracy |
| Search speed | Directly related to template size; the smaller the template, the faster the search speed. | |
| Sensitivity to Physical Changes | • Less sensitive due to the fact that only 30% of the available minutia are required for matching. Cuts | • If the scar or blemish affects the region of the fingerprint image that was scarred, a new template may be required. |

| | and scars usually will not affect all the minutiae on the fingerprint. | |
|---|---|---|
| Template Efficacy | Can extract minutia from partial prints (as often found in crime scenes), making it more feasible to criminal related applications | • Requires the same central region to be patterned for the match to occur. • Not suitable for criminal applications where partial prints are often used as the basis for investigation. |
| Sensitivity to Time | • Less sensitive to changes over time | • More sensitive to physical changes and differences in the fingerprint placement on the sensor – both which become greater over time. |
| Standards | X.509 AAMVA B10.8 | None |
| Leading vendors | Identix | BioScrypt, Precise Biometrics, Digital Persona, Sony |

*Table 2.2. Minutiae vs. Pattern based fingerprint templates.*

## 2.3.5.    Comparing the Performance of Matching Algorithms

An explicit answer has not been provided to the question: what is the best algorithm for matching fingerprints? The two main reasons why it is difficult to assess the relative performance of the various matching algorithms are as follows [26]:

- The performance of a fingerprint recognition method involves a tradeoff among different performance measures: accuracy (e.g., False Match Rate (FMR) and False Non-Match Rate (FNMR), refer to Section 2.3.6), efficiency (enrollment time, verification time), scalability to 1:N identification, template size, and so on. Different applications have different performance requirements. For example, an application may prefer a fingerprint matching algorithm that is lower in accuracy but has a small template size over an algorithm that is more accurate but uses a large template size;
- Most of the scientific work published in the literature includes experimental results carried out on proprietary databases using different protocols, which are usually not shared with the research community. This makes it difficult to compare the results of different methods; the performance measured is related to the "difficulty" of the benchmark.

Before the fingerprint verification competitions FVC2000[30], the only large public domain fingerprint datasets were the National Institute of Standards and Technology (NIST) databases [37]; although these databases constitute an excellent benchmark for AFIS development [38] and fingerprint classification studies, they are not well suited for the evaluation of algorithms operating with live-scan (dab) images.

FVC2000 were organized with the aim of providing fingerprint databases to any interested researcher and to track performance of the state-of-the-art fingerprint matching algorithms. Fortunately, several authors now report the results of their experiments on these databases according to the proposed protocol, thus producing results that can be compared across the whole scientific community. It is hoped that this will become a common practice for scientists and practitioners in the field.

## 2.3.6.        **Recognition Rate**

## 2.3.6.1.     **Terminology and Measurement**

The ultimate measure of utility of a fingerprint system for a particular application is recognition rate. This can be described by two values[2]:

- The *false acceptance rate* (*FAR*) is the ratio of the number of instances of pairs of different fingerprints found to (erroneously) match to the total number of match attempts.
- The *false rejection rate* (*FRR*) is the ratio of the number of instances of pairs of the same fingerprint found not to match to the total number of match attempts.

FAR and FRR trade off against one another. That is, a system can usually be adjusted to vary these two results for the particular application, however decreasing one increases the other and vice versa. FAR is also called, *false match rate* or *Type II error*, and FRR is also called *false non-match rate* or *Type I error*. These are expressed as values in [0, 1] interval or as percentage values.

Some other "compact" indices are also used to summarize the accuracy of a verification system[1]:

- Equal-Error Rate (EER) denotes the error rate at the threshold $t$ for which false match rate and false non-match rate are identical: FMR($t$) = FNMR($t$) (see Figure 2.21). In practice, because the matching score distributions are not continuous (due to the finite number of matched pairs and the quantization of the output scores), an exact EER point might not exist: In this case, instead of a single value, an interval should be reported [30]. Although EER is an important indicator, in practice, a fingerprint-based biometric system is rarely used at the operating point corresponding to EER, and often a more stringent threshold is set to reduce FMR in spite of a rise in FNMR.

- ZeroFNMR is the lowest FMR at which no false non-matches occur (see Figure 2.21).

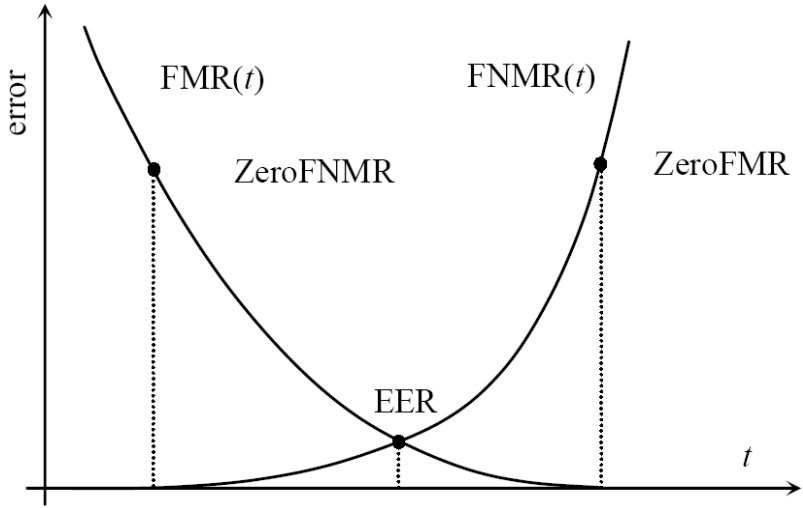

*Figure 2.21.An example of FMR(t) and FNMR(t) curves, where the points corresponding to EER, ZeroFNMR, and ZeroFMR are highlighted.*

- ZeroFMR is the lowest FNMR at which no false matches occur (see Figure 2.21).
- Failure To Capture (FTC) rate is associated with the automatic capture function of a biometric device and denotes the percentage of times the device fails to automatically capture the biometric when it is presented to a sensor. A high failure to capture rate makes the biometric device difficult to use.
- Failure To Enroll (FTE) rate denotes the percentage of times users are not able to enroll in the recognition system. There is a tradeoff between the FTE rate and the accuracy (FMR and FNMR) of a system. FTE errors typically occur when the recognition system performs a quality check to ensure that only good quality templates are stored in the database and rejects poor quality templates. As a result, the database contains only good quality templates and the system accuracy (FMR and FNMR) improves.
- Failure To Match (FTM) rate is the percentage of times the input cannot be processed or matched against a valid template because of insufficient quality. This is different from a false non-match error; in fact, in a failure to match error, the system is not able to make a decision, whereas in false non-match error, the system wrongly decides that the two inputs do not come from the same finger.

The ROC-curve plots FAR versus FRR for a system [2]. (ROC stands for Receiver Operating Curve for historical reasons. Yes, "ROC-curve" is redundant, but this is the common usage.) ROC-curves are shown in Figure 2.22. The FAR is usually plotted on the horizontal axis as the independent variable. The FRR is plotted on the vertical axis as the dependent variable. Because of the range of FAR values, this axis is often on a logarithmic scale. Figure 2.22 contains two solid curves. The solid curves do not represent any particular data; they are included for illustrative purposes to show better and worse curve placements. The typical ROC-curve has a shape whose "elbow" points toward (0, 0) and whose asymptotes are the positive $x$-and $y$-axes. The sharper the elbow and (equivalently) the closer is the ROC-curve to the $x$- and $y$-axes, the lower is the recognition error and the more desirable is the result.

The procedure for using the ROC-curve is as follows. Choose an acceptable level of FAR. In Figure 2.22, a dashed line is shown at 0.01% FAR. The FRR corresponding to this choice is the attainable FRR, in this example about 4%.

Alternatively, the FRR can be specified and the FAR found on the curve. There is no single set of FAR and FRR specifications useful for all different applications. If the fingerprint system is specified for very high security situations such as for military installations, then the FAR will be chosen to be very low (e.g.,<0.001%). However, this results in higher FRR, sometimes in the range from 5% to 20%. Typical customer applications such as for automatic teller machines cannot afford to alienate users with such a high FRR. Therefore, the choice in these applications is low FRR (e.g., <0.5%), at the sacrifice of higher FAR. (An FRR specification that is sometimes quoted for automatic teller machines is less than 1 per 100,000 false rejections.)

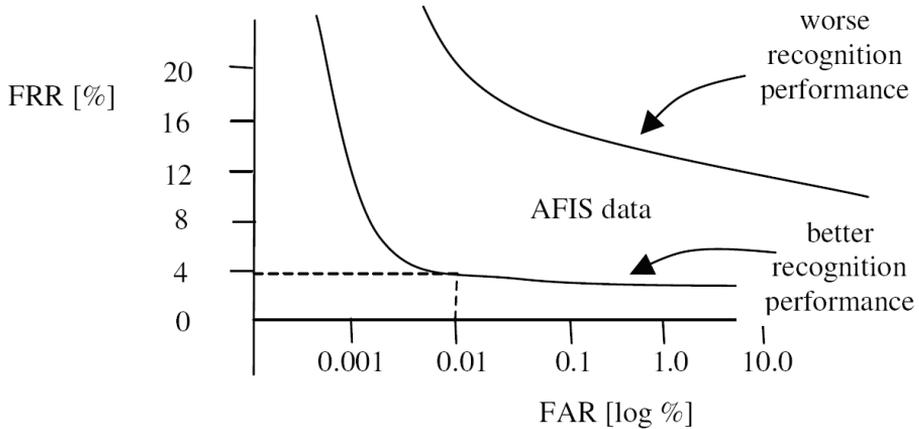

*Figure 2.22. ROC-curves. The 2 solid curves are of hypothetical data illustrating desirable and less desirable recognition performance.*

### 2.3.6.2.    Specifying and Evaluating Recognition Rate Statistics

For statistical results to be properly evaluated, they must be accompanied by the following information:

• *Sample size*: it should contain the following information: the number of subjects (people), the number of fingers, the number of images per finger, and the total number of fingerprint images. In addition, the image number should be broken out into number of match and non-match images. For example, a test might consist of 100 subjects, 2 fingers per subject and 4 images per finger. The total number of images is 100×2×4=800. If each finger (200) is compared against each of the other images from the same finger (4 choose 2 = 6 pairs per unique finger), there are 200×6=1200 matching pairs. If one image from each finger (200) is compared against all images from different subjects (99×4×2=792), there are 200×792=158,400 non-match pairs.

• *Description of population*: it states the type of subjects included in the sample. Of particular importance in judging fingerprint statistics is the type of work engaged in by the subjects. A study involving masons will have different statistical results than those involving white-collar workers whose hands are subject to less abuse. The age statistics should be described, at least stating a relative breakdown on the number of children, adults, and elderly people included in the sample. The proportion of males and females should also be stated.

- *Test design*: should be described. Of particular interest is who performed the tests. The strong preference is that a reputable third-party conducts and reports the test. Was the capture procedure supervised or not? Were the subjects given training or visual feedback to place the finger correctly on the fingerprint capture device? Was the sample manually filtered in any way to remove "goats" (people whose fingerprints are very difficult to capture and match with reliable quality)? Was the procedure adjusted using a practice sample of fingerprints, then tested separately on different images to yield the published results? What was the range of rotational and translational variance allowed, or were the fingerprints manually centered in the image? What were the make and specifications of the capture device? Where and when were the tests conducted (e.g., Florida humid summer or Minnesota dry winter)? What components of the system were involved in the test: just matching algorithms, just sensor, full system? Most test results do not list all these conditions, but the most possible information enables more valid evaluation.

It is important to emphasize that results cannot be compared if determined under different test conditions. It is a misrepresentation of test data to state that a matcher achieved certain results for test design A, so it can be compared against the results from test design B. Valid comparisons between results can be done only for the same database under the same conditions.

## 2.4.   Industrial Fingerprint Matching System

SAN JOSE, Calif. and EMERYVILLE, Calif. – October 11, 2006 – Cypress Semiconductor Corp. (NYSE: CY) and UPEK®, Inc. today introduced a reference design to protect data on external hard disk drives (HDDs) by adding biometric security via fingerprint recognition technology. The reference design delivers the industry's strongest security by authenticating fingerprint matches on-board the external HDD system instead of on the PC, ensuring no data is passed between the HDD and PC until a match has occurred. The new CY4661 reference design uses UPEK's TouchStrip™ Fingerprint High Security Authentication Solution along with Cypress's EZ-USB FX2LP™ USB controller to address the growing need for secure personal information and enterprise data on portable drives.

The reference design enables manufacturers of HDDs to add biometric security that only allows those with a registered fingerprint to access data on the drives. It features match-on-chip, hardware-based authentication for the strongest data and privacy protection available. All fingerprint matching functions (template extraction, storage and matching) are conducted in the secure UPEK fingerprint hardware-based architecture on the external drive, providing true portability and security from PC security risks. UPEK's hardware-based security architecture, capable of encrypting communications, can also enable future online authentication services. The CY4661 is bundled with UPEK's industry leading Protector Suite™ Token application software, which simplifies access to data, applications and websites. The software includes a password bank to securely register and access password-protected websites and applications. The end-to-end solution ensures plug-and-play ease of use and eliminates the need to download software on a host PC.

"The Cypress/UPEK reference design delivers maximum protection for the sensitive data that consumers increasingly store on their portable drives, and it does so with the ease and convenience of the swipe of a finger," said Rajiv Nema, product marketing manager for Cypress's High Speed Peripherals Business Unit. "The flexibility and performance of the EZ-USB FX2LP controller is a perfect fit for the needs of the rapidly growing biometric security market."

"UPEK's complete hardware-based fingerprint solution offers the highest level of biometric security and convenience in the market today," said Vito Fabbrizio, Director of Portable Storage Solutions for UPEK, Inc. "This unparalleled protection is enhanced by the performance of Cypress's industry-leading USB technology and is an essential product for individuals on the go."

The UPEK/Cypress CY4661 biometric reference design enables fast time-to-market and reduced integration costs with its hardware and software solution for secure portable hard disk drives. The reference design includes a field-proven integrated UPEK enrollment tutorial that offers consumers an "easy to start" video training and practice session with performance feedback to ensure an excellent user experience. This results in a satisfying end user experience and greater security due to high-quality matches, leading to fewer support calls, product returns, and improved customer satisfaction.

### 2.4.1.    About UPEK's TouchStrip™ Fingerprint Authentication Solution

The TouchStrip Fingerprint Authentication Solution is an end-to-end solution combining a high image quality silicon sensor, a secure companion processor for biometric processes, and easy-to-use software for protection and convenient access to digital assets. Integrated in the world's largest brand PCs and portable storage devices, the TouchStrip Fingerprint Authentication Solution leverages hardware-based security to protect sensitive data, access to websites and applications, and future online services.

### 2.4.2.    About EZ-USB FX2LP

The EZ-USB FX2LP is the world's lowest power fully integrated programmable peripheral controller, including an 8051 microprocessor, a serial interface engine, a high-speed USB 2.0 transceiver, on-chip RAM and FIFO and a General Programmable Interface (GPIF). It handles all basic USB functions, enabling the host system's microprocessor to focus on application-specific functions and ensuring sustained high-speed data transfer rates. Featuring a data rate of 480 Mbps, 16 Kbytes of on-chip memory, and up to 40 programmable I/Os, the EZ-USB FX2LP solution provides complete design flexibility. There are several package options, including a 5mm x 5mm VFBGA package, ideal for use in small form-factor mobile applications. The CY3684 EZ-USB FX2LP Development Kit is available to facilitate the design process and speed time-to-market.

### 2.4.3.    About UPEK

UPEK, Inc., a global leader in biometric fingerprint security solutions, offers integrated end-to-end solutions including comprehensive design & integration services to the world's leading consumer and industrial products companies. UPEK solutions enable the strongest fingerprint authentication security available, packaged for high user convenience and rapid integration into existing products and network architectures. The company has been pioneering biometric fingerprint technology since 1996 and shipping product in volume since 1999. UPEK is headquartered in Emeryville, California, with offices in Prague, Singapore, Taipei and Tokyo. UPEK and the UPEK logo are registered trademarks or trademarks of UPEK, Inc. in the United States and other countries. All other trademarks or registered trademarks are the property of their respective owners.

### 2.4.4.    About Cypress

Cypress solutions perform: consumer, computation, data communications, automotive, industrial, and solar. Leveraging proprietary silicon processes, Cypress's product portfolio includes a broad selection of wired and wireless USB devices, CMOS image sensors, timing solutions, specialty memories, high-bandwidth synchronous and micropower memory products, optical solutions and reconfigurable mixed-signal arrays. Cypress trades on the NYSE under the ticker symbol CY. Cypress and the Cypress logo are registered trademarks and EZ-USB FX2LP is a trademark of Cypress Semiconductor Corporation. UPEK and the UPEK logo are registered trademarks and TouchStrip and Protector Suite are trademarks of UPEK, Inc. All other trademarks are the property of their respective owners.

As always the case, all these fingerprint systems are not available for analysis and are trade secret. In the following section limitations and motivation are enumerated.

## 2.5. Limitations in conventional fingerprint matching system and motivation

### 2.5.1. Limitations

#### a. Storage
The data size to be stored in conventional fingerprint matching system database for all fingerprints is relatively large where the common approach for matching is comparing the tuples of x, y coordinates and orientation θ of each minutiae in the fingerprint.

#### b. Performance
The time needed until the system responds is relatively large because of many matching stages that consumes more computational power or needs special hardware.

#### c. Sensitivity to noise and rotation
Any translation or rotation to a fingerprint having a template stored in the system database may increase the false match and false non-match rates.

#### d. Vulnerability of the whole systems to attacks
The communication channels between different parts of a fingerprint identification system and the parts themselves are vulnerable to different types of attacks. Also, the system database can be tampered by deleting, inserting, or editing in the fingerprint template records.

## 2.5.2.    Motivation

From the above limitations, there is a need for an improved fingerprint identification system with the following requirements:

### a. Storage
The data size to be stored in the database needs to be less to be able to be inserted on-chip and to make search faster.

### b. Performance
The time needed until the system responds have to be low by using less computational power.

### c. Sensitivity to noise and rotation
A fingerprint matching system that is invariant to translation and rotation is needed so that it can lessen the false match and false non-match rates. A new matching algorithm is proposed that achieve requirements a, b and c in chapter 4.

### d. Vulnerability of the whole systems to attacks
Many scenarios can be applied to secure the whole database or different parts of fingerprint identification systems and communication channels between those different parts. In chapter 5, watermarking is applied on the fingerprint identification system to secure communication channels.

# Chapter 3
# Proposed Fingerprint Thinning Algorithm

As mentioned in Chapter 2, Thinning is a critical pre-processing step to obtain skeletons for pattern analysis in fingerprint recognition[39]. Thinning is the process of reducing the thickness of each line of patterns to a single pixel[40].

Most of the current thinning algorithms used in fingerprint preprocessing operations need the step of binarization before thinning as in [3], [4], [41], [42], [43],[44], [45], and [46].

Binarization step causes the following problems:

1. A significant amount of information may be lost during the binarization process;
2. Binarization is time consuming and may introduce a large number of spurious minutiae;
3. In the absence of an a priori enhancement step, most of the binarization techniques do not provide satisfactory results when applied to low-quality images.

The effect of these mistaken minutiae is to decrease recognition accuracy and hence increase the false match and false non match rates.

Motivated by this analysis, a new thinning algorithm is proposed in this chapter. This novel thinning algorithm works directly on the gray-scale images without needing the binarization step. The proposed thinning algorithm is described in section 3.1, Experimental results are drawn in section 3.2, and finally performance evaluation is shown in section 3.3.

## 3.1. The proposed thinning algorithm

The main idea is that the gray values of pixels of ridges in the fingerprint image gradually decrease going from an edge towards the center of the ridge line and increase again going towards the other edge. This is the definition of a *local minimum*. This local minimum is the line we want to reach to convert the ridge of (e.g. 5) pixels wide into one pixel wide.

There is a difference in the procedure of searching for a local minimum between a horizontal (or near horizontal) ridge (referred to as type1) and a vertical (or near vertical) ridge (referred to as type2),see Figure 3.1, where the fingerprint image (Figure 3.1.a) is taken from the standard fingerprint image database FVC2000[30].

*Figure 3.1. a) The fingerp___ ___(horizontal) zoomed block of size 8x8, c) a t___ ___l block of size 8x8*

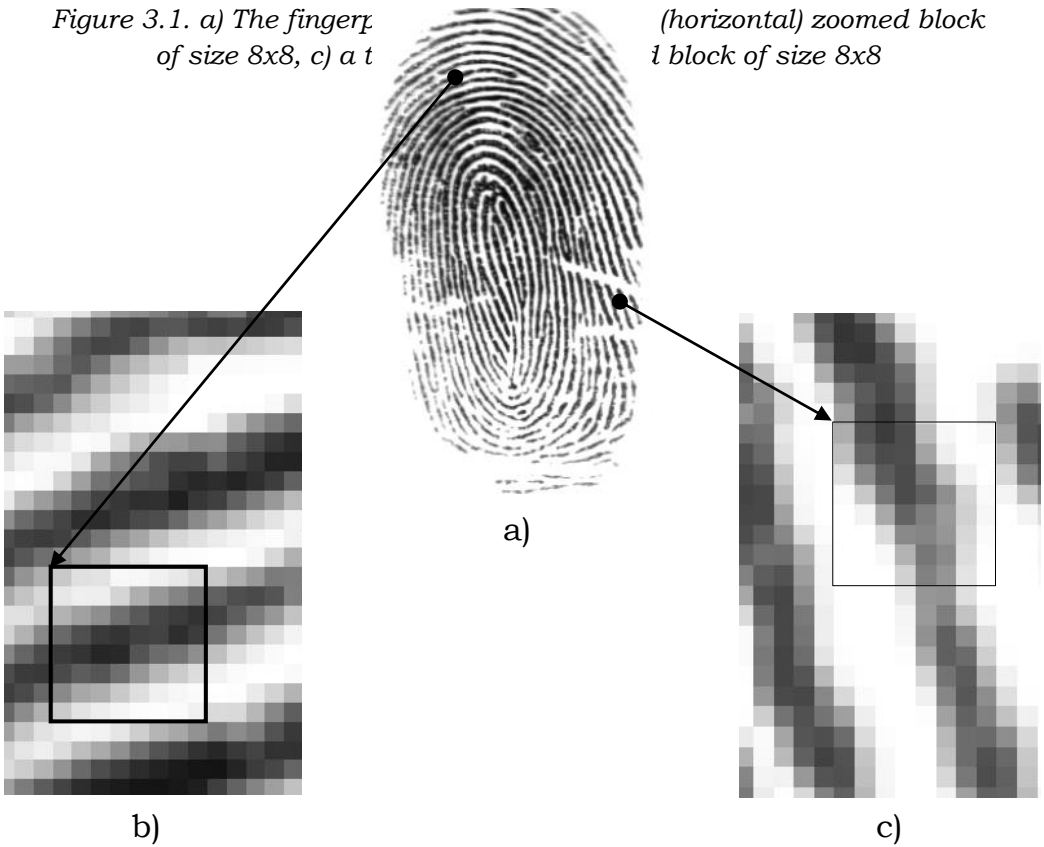

a)

b)                                                                c)

For type1 ridges, the block is scanned column by column, during this scan, the local minimum is searched for in each column, and in a new block initialized by zeros, all local minima in those columns will be replaced with ones.

For type2 ridges, the block is scanned row by row, during this scan, the local minimum is searched for in each row, and in a new block initialized by zeros, all local minima in those rows will be replaced with ones.

### 3.1.1. Algorithm steps

3.1.1.1.    Divide the fingerprint image into type1 and type2 blocks

1. Divide the fingerprint image into blocks of size 8×8,
2. Read the gray values of the block, and then
3. Discriminate between type1 and type2 blocks.

Figure 3.2 shows the discrimination process steps. To discriminate between type1 and type2 blocks:

For each block of size 8×8
    {If the block is in the fingerprint image area (not the background)
        {
        Sum_of_columns = sum of the gray values of each column in the block;
        Sum_of_rows = sum of the gray values of each row in the block;
        Min1 = minimum value of the vector "sum_of_columns";
        Min2 = minimum value of the vector "sum_of_rows";
        }
    If min1>min2
        This block is identified as type1;
    Else
        This block is identified as type2;
    }

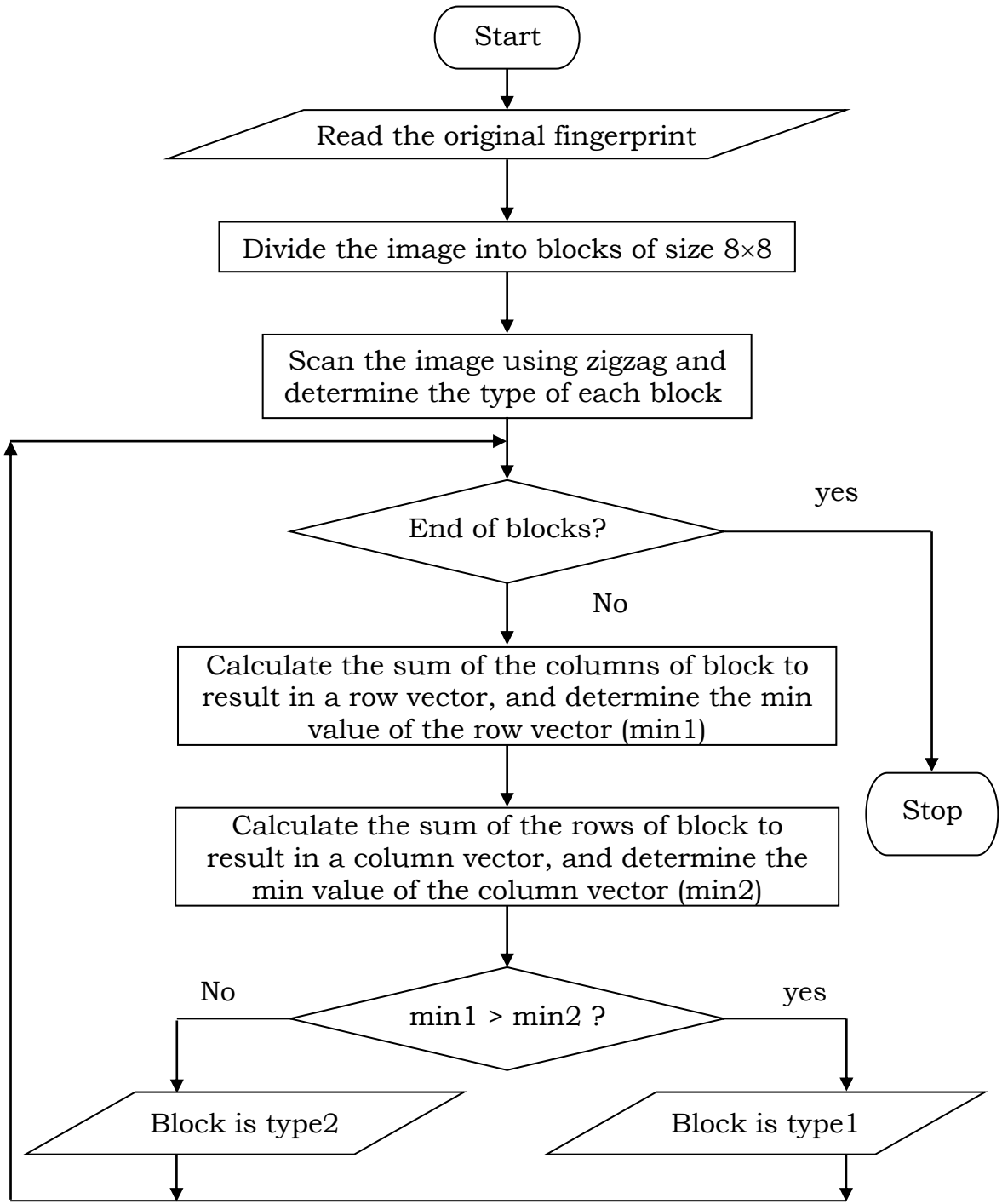

*Figure 3.2. The steps of discrimination between type1 and type2 blocks.*

Figure 3.3 presents the flowchart of the proposed algorithm, which is divided into the following steps:

1. Divide the fingerprint image into blocks (of size 8×8) of two types (near horizontal ridge (type1), or near vertical ridge (type2)). After this step, each block is reduced from a gray-scale block (the values are from 0 (black) to 255 (white)) of size 8×8 into a single value block (1 for type1, 2 for type2, and 0 for background) of size 1×1.

2. Segment the image to deal only with blocks of type1 and type2 containing the fingerprint area without background.

3. Determine the indices of blocks (coordinates of blocks in the image) of horizontal type alone and that of vertical type alone.

4. Initialize a new array of the same size of the original fingerprint with zeros to represent the thin fingerprint.

5. Scan each group of consecutive blocks of type1 vertically column by column to find the local minimum values and set these pixels to one in the new image.

6. Scan each group of consecutive blocks of type2 horizontally row by row to find the local minimum values and set these pixels to one in the new image.

The proposed algorithm is implemented and tested at [47]. Appendix C shows the MATLAB source code implementation for the proposed thinning algorithm.

Consider the fingerprint image shown in Figure 3.1.a, the size of the image is 480×640 and after segmentation, size is 376×208. As an example, consider the block (of size 8×8) located at the intersection of rows (57-64) and columns (65-72), the bold rectangle shown in Figure 3.1.b, the gray values obtained from the fingerprint image in this bold rectangle are shown in Table 3.1.

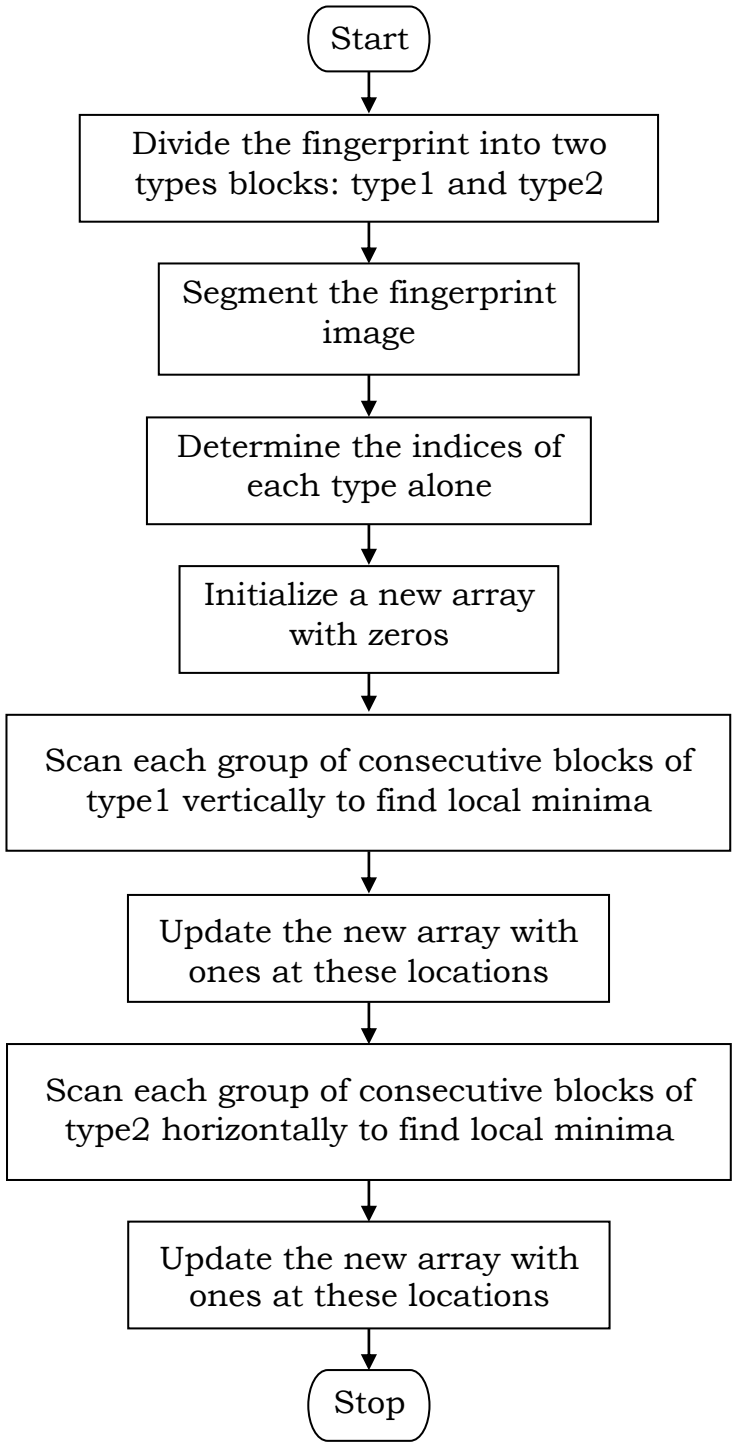

*Figure 3.3. A flowchart of the proposed algorithm.*

| row \ col | 65 | 66 | 67 | 68 | 69 | 70 | 71 | 72 | Sum_of_rows |
|---|---|---|---|---|---|---|---|---|---|
| 57 | 201 | 224 | 227 | 248 | 247 | 241 | 233 | 218 | 1839 |
| 58 | 231 | 239 | 238 | 213 | 175 | 152 | 132 | 112 | 1492 |
| 59 | 179 | 161 | 143 | 112 | 89 | 69 | 60 | 50 | 863 |
| 60 | 105 | 79 | 58 | 52 | 64 | 67 | 46 | 60 | 531(min2) |
| 61 | 59 | 52 | 38 | 41 | 84 | 108 | 116 | 143 | 641 |
| 62 | 74 | 93 | 85 | 111 | 151 | 183 | 204 | 221 | 1122 |
| 63 | 138 | 168 | 175 | 202 | 223 | 240 | 247 | 251 | 1644 |
| 64 | 203 | 227 | 238 | 245 | 252 | 253 | 251 | 241 | 1910 |
| Sum_of_columns | 1190 min1 | 1243 | 1202 | 1224 | 1285 | 1313 | 1289 | 1296 | |

*Table 3.1. The gray values of a type1 block.*

As seen from Table 3.1, the center line of the ridge is shaded and this is what the proposed thinning algorithm should reach. When sum_of_columns and sum_of_rows vectors (vectors are of size 1×8) are calculated and the minimum value of each vector (min1 and min2 respectively) is determined, it is found that min1>min2 so that we can identify the block as a horizontal one (i.e. type1) and this is clear from Figure 3.1.b.

As another example, taking another block (of size 8×8) located at the intersection of rows (225-232) and columns (193-200), the rectangle shown in Figure 3.1.c, the values obtained are shown in Table 3.2.

As seen from Table, the center line of the ridge is shaded and this is what we want the thinning algorithm to reach. Calculating the sum_of_columns and sum_of_rows vectors and getting the minimum value of each vector (min1 and min2 respectively), it is found that min1<min2 so the block can be identified as a vertical one (i.e. type2) and this is clear from Figure 3.1.c.

The result of this step is shown in Figure 3.4, where type1 blocks are drawn with gray and type2 blocks are drawn with black.

| col / row | 193 | 194 | 195 | 196 | 197 | 198 | 199 | 200 | Sum_of _rows |
|---|---|---|---|---|---|---|---|---|---|
| 57 | 174 | 84 | 60 | 77 | 189 | 247 | 255 | 249 | 1335(min2) |
| 58 | 192 | 101 | 75 | 78 | 164 | 238 | 255 | 254 | 1357 |
| 59 | 226 | 139 | 91 | 75 | 120 | 193 | 243 | 255 | 1342 |
| 60 | 251 | 200 | 120 | 75 | 97 | 142 | 213 | 250 | 1348 |
| 61 | 255 | 244 | 169 | 98 | 127 | 154 | 208 | 247 | 1502 |
| 62 | 255 | 255 | 220 | 137 | 133 | 150 | 215 | 251 | 1616 |
| 63 | 255 | 255 | 248 | 179 | 119 | 138 | 221 | 254 | 1669 |
| 64 | 255 | 255 | 255 | 209 | 129 | 149 | 228 | 255 | 1735 |
| Sum_of _columns | 1863 | 1533 | 1238 | 928 | 1078 | 1411 | 1838 | 2015 | |
| | | | | min1 | | | | | |

*Table 3.2. The gray values of a type2 block*

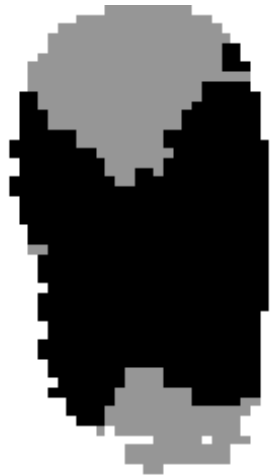

*Figure 3.4. The resulted discrimination between type1 blocks (drawn with gray) and type2 blocks (drawn with black).*

### 3.1.1.2.    Fingerprint Segmentation

Step 2 in the algorithm is for segmentation of fingerprint image, the steps are as follows:

After step1, we have for each block of size 8×8 a corresponding block of size 1×1 with a value of 0, 1 or 2. A closing operation is performed (dilation followed by erosion) followed by an opening operation (erosion followed by dilation) [5]. After that, the perimeter of the fingerprint area is detected where we will have a binary image of ones in the perimeter and zeros elsewhere.

So, now we can find the left most and right most blocks indices with value 1, also we can find the top most and bottom most blocks indices with value 1. These four blocks indices will guide us to determine the rectangle location containing the fingerprint area without background, and the segmentation result was an image of size 376×208 after the original size 480×640.

### 3.1.1.3.    Find local minima in each block

Step3 in the algorithm is implemented to deal separately with each type alone. During recording the indices, consecutive blocks of a certain type is tagged with a zero to separate it from the next consecutive blocks of the same type.

After implementing remaining steps, the thin fingerprint is obtained as shown in Figure 3.5. As seen, the ridges are one pixel wide.

Considering the same 2 blocks shown before in Table 3.1 and Table 3.2, Table 3.3 and Table 3.4 respectively show the results of the thinning algorithm.

### 3.1.1.4.    Correcting errors

After implementing the thinning algorithm, it is found that most bifurcations are separated with one or two pixels, i.e. not connected. So, a complete scan on the resulted thinned image must be performed to search for these cases and update their values as follows:

For each pixel in the thinned image
    {

If pixel_value = 1 (pixel is in the skeleton of the thinned image)

    {Get the number of ones in the 3×3 neighbors of this pixel (n1);

    If n1=1

        {Get the 3×3 neighbors of this pixel in the gray-scale version of the fingerprint;

        Get the location c of the minimum value among the 8 neighbors of this pixel;

        If this pixel at location c in the thinned image = 0

          {

        Get the 3×3 neighbors of this pixel in the thinned image;

        If there are three or more of the neighbor pixels = 1

            Update the value of this pixel to be 1 and keep only three "1" neighbors unchanged to be a valid bifurcation

          }

        }

    }

}

| col / row | 65 | 66 | 67 | 68 | 69 | 70 | 71 | 72 |
|---|---|---|---|---|---|---|---|---|
| 57 | 0 | 0 | 0 | 0 | 0 | 0 | 0 | 0 |
| 58 | 0 | 0 | 0 | 0 | 0 | 0 | 0 | 0 |
| 59 | 0 | 0 | 0 | 0 | 0 | 0 | 0 | 1 |
| 60 | 0 | 0 | 0 | 0 | 1 | 1 | 1 | 0 |
| 61 | 1 | 1 | 1 | 1 | 0 | 0 | 0 | 0 |
| 62 | 0 | 0 | 0 | 0 | 0 | 0 | 0 | 0 |
| 63 | 0 | 0 | 0 | 0 | 0 | 0 | 0 | 0 |
| 64 | 0 | 0 | 0 | 0 | 0 | 0 | 0 | 0 |

*Table 3.3. The result of thinning of block of Table 3.1.*

After applying these steps on Figure 3.5, the result is shown in Figure 3.6, where all bifurcations are connected correctly.

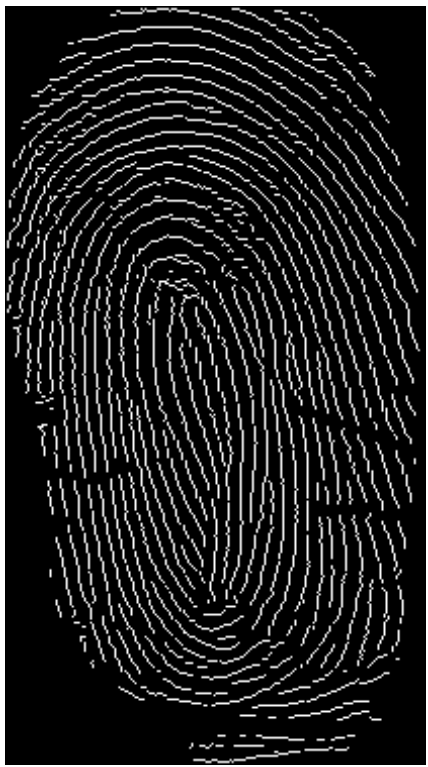

*Figure 3.5. The thinned fingerprint image.*

| col\row | 193 | 194 | 195 | 196 | 197 | 198 | 199 | 200 |
|---|---|---|---|---|---|---|---|---|
| 225 | 0 | 0 | 1 | 0 | 0 | 0 | 0 | 0 |
| 226 | 0 | 0 | 1 | 0 | 0 | 0 | 0 | 0 |
| 227 | 0 | 0 | 0 | 1 | 0 | 0 | 0 | 0 |
| 228 | 0 | 0 | 0 | 1 | 0 | 0 | 0 | 0 |
| 229 | 0 | 0 | 0 | 1 | 0 | 0 | 0 | 0 |
| 230 | 0 | 0 | 0 | 0 | 1 | 0 | 0 | 0 |
| 231 | 0 | 0 | 0 | 0 | 1 | 0 | 0 | 0 |
| 232 | 0 | 0 | 0 | 0 | 1 | 0 | 0 | 0 |

*Table 3.4. The result of thinning of block of Table 3.2.*

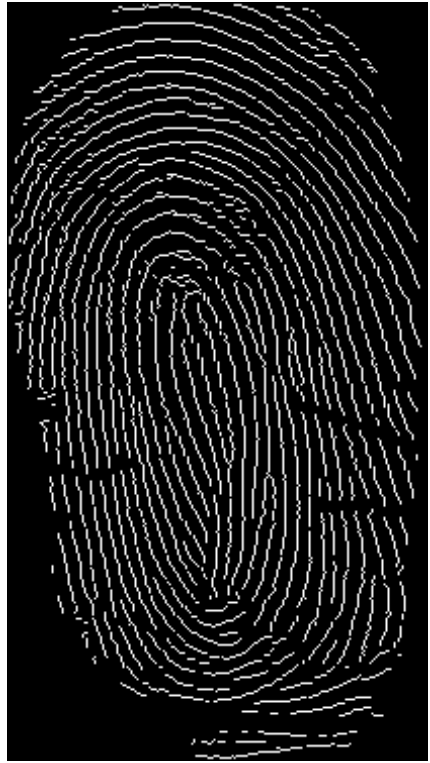

*Figure 3.6. The thinned fingerprint after fixing the bifurcation problems.*

## 3.2. Wuzhili Thinning Method

Before thinning, wuzhili program [48] apply the following pre-processing steps to the fingerprint:

1. Subtract all pixel values from 255.
2. Histogram equalization.
3. Fast Fourier Transformation FFT.
4. Adaptive thresholding to binarize the image.
5. Estimate the orientation image of the fingerprint.
6. Determine the fingerprint area and remove all background.

Then, the Algorithm used to thin the fingerprint uses the method described in [28]:

1. Divide the image into two distinct subfields in a checkerboard pattern.
2. In the first subiteration, delete pixel $p$ from the first subfield if and only if the conditions $G1$, $G2$, and $G3$ are all satisfied.
3. In the second subiteration, delete pixel $p$ from the second subfield if and only if the conditions $G1$, $G2$, and $G3'$ are all satisfied.

Where:

Condition G1 is:
$X_H(p)=1$
where

$$X_H(p) = \sum_1^4 b_i$$

$$b_i = \begin{cases} 1 & \text{if } x_{2i-1} = 0 \text{ and } (x_{2i} = 1 \text{ or } x_{2i+1} = 1) \\ 0 & \text{otherwise} \end{cases}$$

$x_1$, $x_2$, ..., $x_8$ are the values of the eight neighbors of $p$, starting with the east neighbor and numbered in counter-clockwise order.

Condition G2 is:
$2 \leq \min \{n_1(p), n_2(p)\} \leq 3$
where

$$n_1(p) = \sum_{k=1}^4 x_{2k-1} \vee x_{2k}$$

$$n_2(p) = \sum_{k=1}^4 x_{2k} \vee x_{2k+1}$$

Condition G3 is:
$(x_2 \vee x_3 \vee \bar{x}_8) \wedge x_1 = 0$
Condition G3' is:
$(x_6 \vee x_7 \vee \bar{x}_4) \wedge x_5 = 0$

The two subiterations together make up one iteration of the thinning algorithm. The iterations can be repeated until the image stops changing.

## 3.3.  Experimental Results

As shown, the proposed thinning algorithm does not apply any pre-processing operations (including binarization) on the fingerprint image before thinning.

The execution time of the proposed thinning algorithm is found to be 0.52 s. Comparing the proposed algorithm with Wuzhili method implemented using MATLAB in [48] with the same fingerprint, the result of thinning of this method is shown in Figure 3.7 where the execution time was 1.1 s. Then, the result of the proposed thinning algorithm is better than that in Figure 3.7 and consumes less time. The above results was implemented using MATLAB version 7.0.1.24704 (R14) Service Pack1 September 13, 2004 as the programming platform. The program was tested on a 2.00GHz personal computer with 1.99 GB of RAM.

## 3.4.  Performance Evaluation

Most of the algorithms performing binarization before thinning of the image, so if the image is of size N×N, the step of binarization alone needs $O(N^2)$ comparison operations because the main step in binarization is comparing the gray value of each pixel under consideration with a predefined threshold. The proposed algorithm does not need the binarization process during thinning so it does not have these $O(N^2)$ comparison operations. This and the segmentation process are the main cause of the fastness of the proposed algorithm.

Putting Figure 3.6 beside Figure 3.7 in Figure 3.8 to be easy to be compared, it is seen that the fingerprint in the left side (which is the result of the proposed thinning method) is clearer than the fingerprint in the right side (which is the result of Whuzhili method) due to lower number of H breaks and spikes. So, it is expected that both false positive rate and false negative rate will be lowered because the performance of a minutiae extraction algorithm relies heavily on the quality of the thinning algorithm.

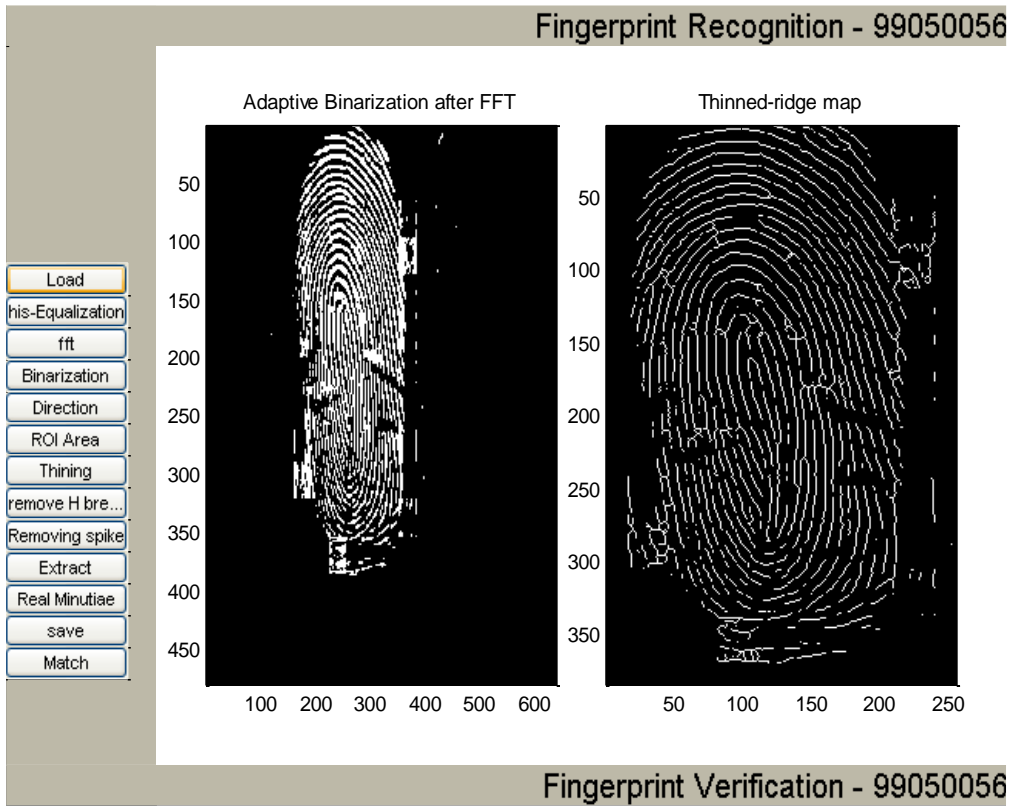

*Figure 3.7. The thinning step of the program implemented in [48]*

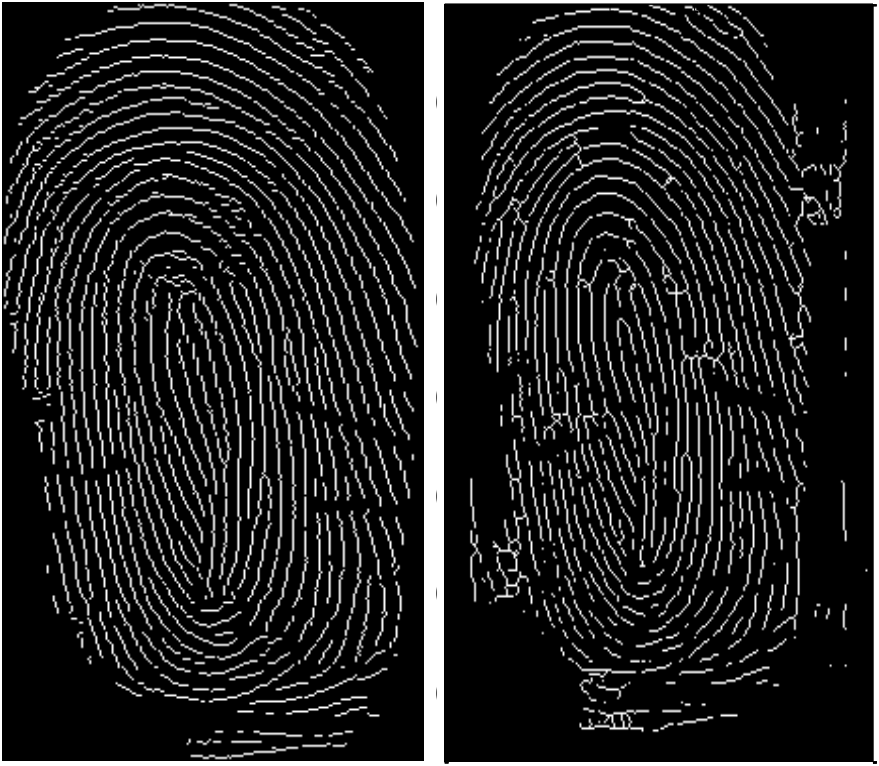

*Figure 3.8. comparing both thinning results for the same fingerprint*

# Chapter 4
# Proposed Fingerprint Matching Algorithm

As mentioned in Chapter 2, fingerprint matching is the last step in Automatic Fingerprint Identification System (AFIS). Fingerprint matching techniques can be classified into three types:

- Correlation-based matching,
- Minutiae-based matching, and
- Non-Minutiae feature-based matching.

Minutiae-based matching is the most popular and widely used technique, being the basis of the fingerprint comparison.

In [3], a novel minutiae matching method is presented that describes elastic distortions in fingerprints by means of a thin-plate spline model, which is estimated using a local and a global matching stage. After registration of the fingerprints according to the estimated model, the number of matching minutiae can be counted using very tight matching thresholds. For deformed fingerprints, the algorithm gives considerably higher matching scores compared to rigid matching algorithms, while only taking 100 ms on a 1 GHz P-III machine. Furthermore, it is shown that the observed deformations are different from those described by theoretical models proposed in the literature.

In [49], minutia polygons are used to match distorted fingerprints. A minutia polygon describes not only the minutia type and orientation but also the minutia shape. This allows the minutia polygon to be bigger than the conventional tolerance box without losing matching accuracy. In other words, a minutia polygon has a higher ability to tolerate distortion. Furthermore, the proposed matching method employs an improved distortion model using a Multi-quadric basis function with parameters. Adjustable parameters make this model more suitable for fingerprint distortion. Experimental results show that the proposed method is two times faster and more accurate (especially, on fingerprints with heavy distortion) than the method in [3].

In[34], a new fingerprint minutia matching technique is proposed, which matches the fingerprint minutiae by using both the local and global structures of minutiae. The local structure of a minutia describes a rotation and translation invariant feature of the minutia in its neighborhood. It is used to find the correspondence of two minutiae sets and increase the reliability of the global matching. The global structure of minutiae reliably determines the uniqueness of fingerprint. Therefore, the local and global structures of minutiae together provide a solid basis for reliable and robust minutiae matching. The proposed minutiae matching scheme is suitable for an on line processing due to its high processing speed. Experimental results show the performance of the proposed technique.

In [50], a hybrid matching algorithm that uses both minutiae (point) information and texture (region) information is presented for matching the fingerprints. Results obtained shows that a combination of the texture-based and minutiae-based matching scores leads to a substantial improvement in the overall matching performance. This work was motivated by the small contact area sensors provide for the fingertip and, therefore, sense only a limited portion of the fingerprint. Thus multiple impressions of the same fingerprint may have only a small region of overlap. Minutiae-based matching algorithms, which consider ridge activity only in the vicinity of minutiae points, are not likely to perform well on these images due to the insufficient number of corresponding points in the input and template images.

In [51], a new and efficient method for minutiae-based fingerprint matching is proposed, which is invariant to translation, rotation and distortion effects of fingerprint patterns. The algorithm is separated from a prior feature extraction and uses a compact description of minutiae features in fingerprints. The matching process consists of three major steps:

- Finding pairs of possibly corresponding minutiae in both fingerprint patterns,
- Combining these pairs to valid tuples of four minutiae each, containing two minutiae from each pattern.
- The third step is the matching itself.
  It is realized by a monotonous tree search that finds consistent combinations of tuples with a maximum number of different minutiae pairs. The approach has low and scalable memory requirements and it is computationally inexpensive.

In [45], three ideas are introduced along the following three aspects:

- Introduction of ridge information into the minutiae matching process in a simple but effective way, which solves the problem of reference point pair selection with low computational cost;
- Use of a variable sized bounding box to make our algorithm more robust to non-linear deformation between fingerprint images;
- Use of a simpler alignment method in our algorithm.
  Experiments using the Fingerprint Verification Competition 2000 (FVC2000) databases with the FVC2000 performance evaluation show that these ideas are effective.

In [52], a novel minutiae indexing method is proposed to speed up fingerprint matching, which narrows down the searching space of minutiae to reduce the expense of computation. An orderly sequence of features is extracted to describe each minutia and the indexing score is defined to select minutiae candidates from the query fingerprint for each minutia in the input fingerprint. The proposed method can be applied in both minutiae structure-based verification and fingerprint identification. Experiments are performed on a large-distorted fingerprint database (FVC2004 DB1) to approve the validity of the proposed method.

In most existing minutiae-based matching methods, a reference minutia is chosen from the template fingerprint and the query fingerprint, respectively. When matching the two sets of minutiae, the template and the query, firstly reference minutiae pair is aligned coordinately and directionally, and secondly the matching score of the rest minutiae is evaluated. This method guarantees satisfactory alignments of regions adjacent to the reference minutiae. However, the alignments of regions far away from the reference minutiae are usually not so satisfactory. In [53], a minutia matching method based on global alignment of multiple pairs of reference minutiae is proposed. These reference minutiae are commonly distributed in various fingerprint regions. When matching, these pairs of reference minutiae are to be globally aligned, and those region pairs far away from the original reference minutiae will be aligned more satisfactorily. Experiment shows that this method leads to improvement in system identification performance.

In [54], the design and implementation of an on-line fingerprint verification system is described. This system operates in two stages: minutia extraction and minutia matching. An improved version of the minutia extraction algorithm proposed by Ratha et al. [7], which is much faster and more reliable, is implemented for extracting features from an input fingerprint image captured with an on-line inkless scanner. For minutia matching, an alignment-based elastic matching algorithm has been developed. This algorithm is capable of finding the correspondences between minutiae in the input image and the stored template without resorting to exhaustive search and has the ability of adaptively compensating for the nonlinear deformations and inexact pose transformations between fingerprints. The system has been tested on two sets of fingerprint images captured with inkless scanners. The verification accuracy is found to be acceptable. Typically, a complete fingerprint verification procedure takes, on an average, about eight seconds on a SPARC 20 workstation. These experimental results show that our system meets the response time requirements of on-line verification with high accuracy.

In [55], a minutia matching algorithm which modified Jain et al.'s [54] algorithm is proposed The algorithm can better distinguish two images from different fingers and is more robust to nonlinear deformation. Experiments done on a set of fingerprint images captured with an inkless scanner shows that the algorithm is fast and has high accuracy.

In [56], a new fingerprint minutiae matching algorithm is proposed, which is fast, accurate and suitable for the real time fingerprint identification system. In this algorithm, the core point is used to determine the reference point and a round bounding box is used for matching. Experiments done on a set of fingerprint images captured with a scanner showed that the algorithm is faster and more accurate than Xiping Luo's [55] algorithm.

There are two major shortcomings of the traditional approaches to fingerprint representation[15]:
1. For a considerable fraction of population, the representations based on explicit detection of complete ridge structures in the fingerprint are difficult to extract automatically. The widely used minutiae-based representation does not utilize a significant component of the rich discriminatory information available in the fingerprints. Local ridge structures cannot be completely characterized by minutiae.
2. Further, minutiae-based matching has difficulty in quickly matching two fingerprint images containing different number of unregistered minutiae points.

The filter-based algorithm in [15] uses a bank of Gabor filters, explained in section 2.1.5, to capture both local and global details in a fingerprint as a compact fixed length FingerCode. The fingerprint matching is based on the Euclidean distance between the two corresponding FingerCodes and hence is extremely fast. Verification accuracy achieved is only marginally inferior to the best results of minutiae-based algorithms published in the open literature [57]. Proposed system performs better than a state-of-the-art minutiae-based system when the performance requirement of the application system does not demand a very low false acceptance rate. Finally, it is shown that the matching performance can be improved by combining the decisions of the matchers based on complementary (minutiae-based and filter-based) fingerprint information.

Motivated by this analysis, a new algorithm is proposed in this chapter. This novel algorithm is minutiae-based matching algorithm. The proposed matching algorithm is described in section 4.1, the advantages are drawn in section 4.2, and finally, implementation, performance evaluation of the algorithm and conclusion are explained in section 4.3.

## 4.1. Formal description of fingerprint identification system:

If we define:

X=fingerprint image
T= fingerprint template
PID= person identity
R=result of matching

We can define:

True match as:
$$\forall X \in image(PID) \wedge X \equiv T \rightarrow R = True\,match$$

False match as:
$$\exists X \notin image(PID) \wedge X \equiv T \rightarrow R = false\,match$$

True nonmatch as:
$$\exists X \notin image(PID) \wedge X \neq T \rightarrow R = true\,non\,match$$

False nonmatch as:

$$\forall X \in image(PID) \wedge X \neq T \rightarrow R = false\,non\,match$$

# 4.2. The proposed matching algorithm

Any Fingerprint Identification System (FIS) has two phases, fingerprint enrollment and fingerprint matching (identification or verification), as shown in Figure 2.2.

### 4.2.1. Enrollment phase

Figure 4.1 shows the steps of the enrollment phase of the proposed matching algorithm, which is divided into the following steps:

1. Get the core point location of the fingerprint to be enrolled after applying enhancement process.
2. Extract all minutiae from the fingerprint image.
3. From output data of step2, get the minutiae locations (x, y coordinates) together with their type: type1 for termination minutiae and type2 for bifurcation minutiae.
4. Construct tracks of 10 pixels wide centered at the core point.
5. In each track, count the number of minutiae of type1 and the number of minutiae of type2.
6. Construct a table of two columns, column 1 for type1 minutiae and column 2 for type2 minutiae, having number of rows equal number of found tracks.

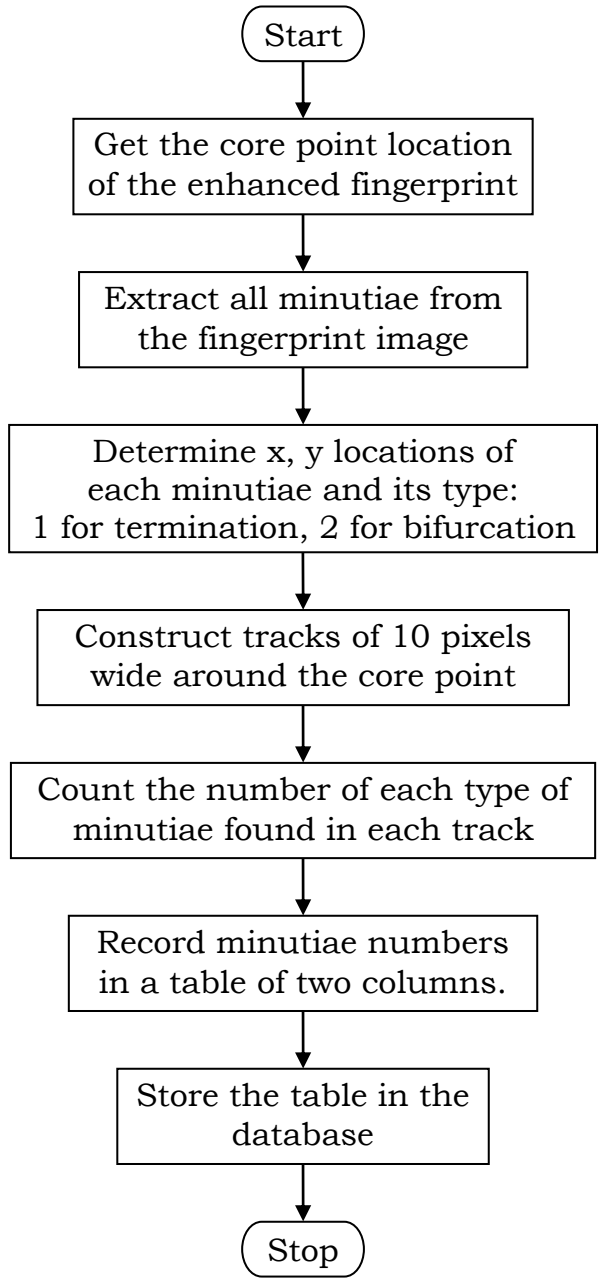

*Figure 4.1. Flowchart of the enrollment phase of the proposed matching algorithm.*

7. In the first row, record the number of minutiae of type1 found in the first track in the first column, and the number of minutiae of type2 found in the first track in the second column.

8. Repeat step 7 for the remaining tracks of the fingerprint, and then store the table in the database.

This enrollment phase will be repeated for all prints of the same user's fingerprint. Number of prints depends on the application requirements at which the user registration takes place. For FVC2000 [30], described in section 2.1.6, there are 8 prints for each fingerprint. So, eight enrollments will be required for each user to be registered in the application. Finally, eight tables will be available in the database for each user.

### 4.2.2. Verification phase

For authenticating a user, verification phase should be applied on the user's fingerprint to be verified at the application.

Figure 4.2 shows the steps of the verification phase of the proposed matching algorithm, which is divided into the following steps:

1. Capture the fingerprint of the user to be verified.
2. Apply the steps of enrollment phase, described in section 4.1.1, on the captured fingerprint to obtain its minutiae table T.
3. Get all the minutiae tables corresponding to the different prints of the claimed fingerprint from the database.
4. Get the absolute differences cell by cell between minutiae table T and all minutiae tables of the claimed fingerprint taken from the database, now we have eight difference tables.
5. Get the summations of all cells in each of column1 (type1) and column2 (type2) for each difference table, now we have sixteen summations.
6. Get the geometric mean of the eight summations of type1 columns (gm1), and the geometric mean of the eight summations of type2 columns (gm2).

7. Check: if gm1<= threshold1 and gm2<=threshold2 then the user is genuine and accept him; else the user is imposter and reject him.

## 4.3. Advantages of the proposed matching algorithm

The proposed minutiae-based matching algorithm has the following advantages:

1. Since all cells in each minutiae table, representing the fingerprint in database, contain *just* the number of minutiae of type1 or type2 in each track around the core point of the fingerprint, neither position (x or y) nor orientation (θ) of the minutiae is considered, the algorithm is *rotation* and *translation invariant*.

2. The numbers of minutiae to be stored in the database *need less storage* than traditional minutiae-based matching algorithms which store position and orientation of each minutia. Experiments show that nearly 50% reduction in storage size is obtained.

3. Matching phase itself *takes less time* which as will be shown in section 4.3.3.8 reaches 0.00134 sec.

## 4.4. Implementation of the proposed matching algorithm

Using MATLAB Version 7.9.0.529 (R2009b), both proposed enrollment and verification phases are implemented in Appendix B, as described in next two subsections:

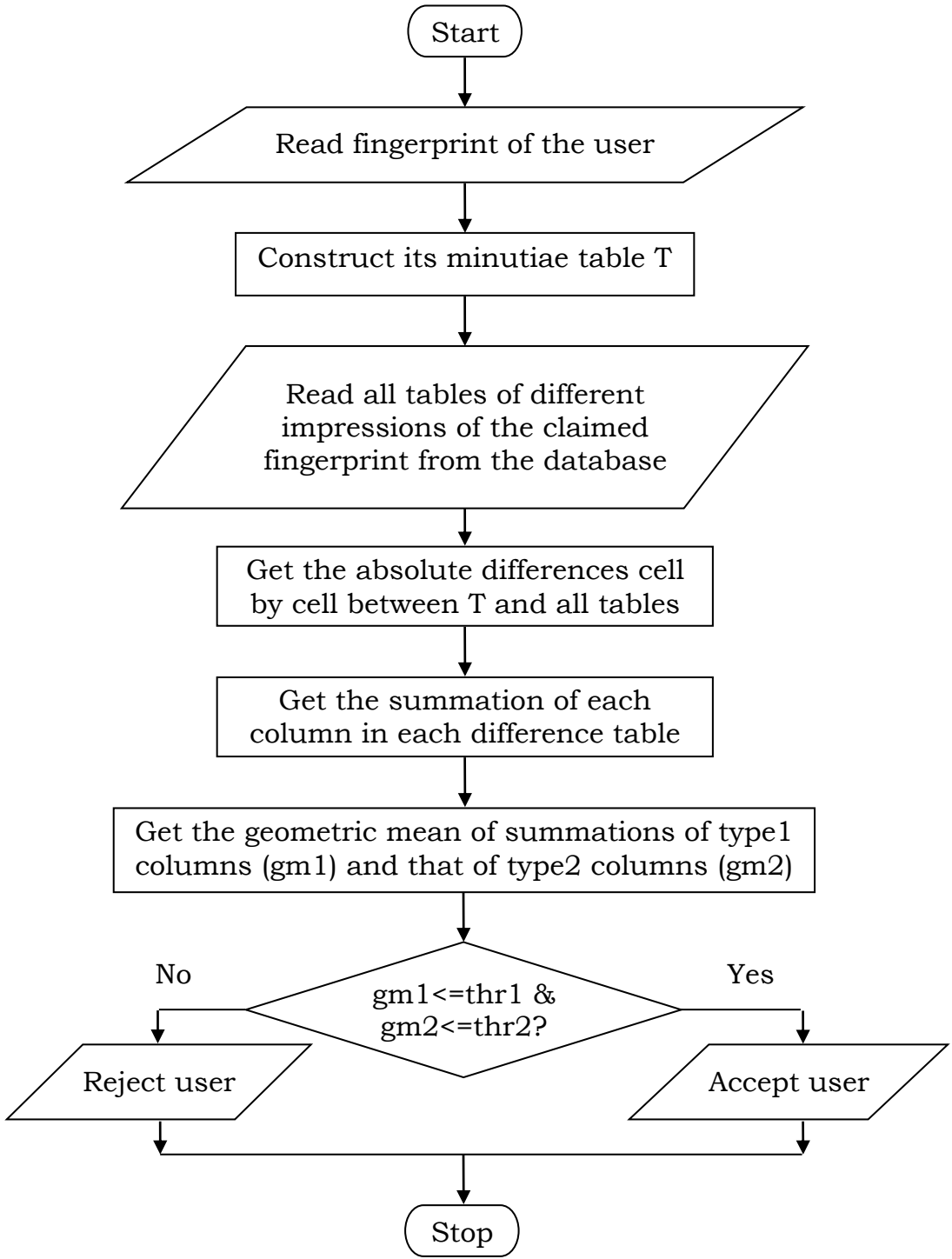

*Figure 4.2. Flowchart of the proposed verification phase.*

### 4.4.1.    **Enrollment phase**

4.4.1.1. Enhancement of the fingerprint image

The first step is to enhance the fingerprint image using Short Time Fourier Transform STFT analysis [2]. The performance of a fingerprint matching algorithm depends critically upon the quality of the input fingerprint image. While the quality of a fingerprint image may not be objectively measured, it roughly corresponds to the clarity of the ridge structure in the fingerprint image, and hence it is necessary to enhance the fingerprint image. Since the fingerprint image may be thought of as a system of oriented texture with non-stationary properties, traditional Fourier analysis is not adequate to analyze the image completely as the STFT analysis does [21].

Fingerprint enhancement MATLAB code is available at [58].

The algorithm for image enhancement consists of two stages as summarized below:

**Stage 1**: STFT analysis
1. For each overlapping block in an image, generate and reconstruct a ridge orientation image by computing gradients of pixels in a block, and a ridge frequency image through obtaining the FFT value of the block, and an energy image by summing the power of FFT value;
2. Smoothen the orientation image using vector average to yield a smoothed orientation image, and generate a coherence image using the smoothed orientation image;
3. Generate a region mask by thresholding the energy image;

**Stage 2**: Apply Enhancement
For each overlapping block in the image, the next five sub-steps are applied.

4. Generate an angular filter Fa centered on the orientation in the smoothed orientation image with a bandwidth inversely proportional to coherence image;

5. Generate a radial filter Fr centered on frequency image;

6. Filter a block in the FFT domain, F=F×Fa×Fr;

7. Generate the enhanced block by inverse Fourier transform IFFT(F);

8. Reconstruct the enhanced image by composing enhanced blocks, and yield the final enhanced image with the region mask.

The result of the enhancement process is shown in Figure 4.3, where Figure 4.3.a is taken from FVC2000 DB1_B (108_5) and Figure 4.3.b is the enhanced version of Figure 4.3.a.

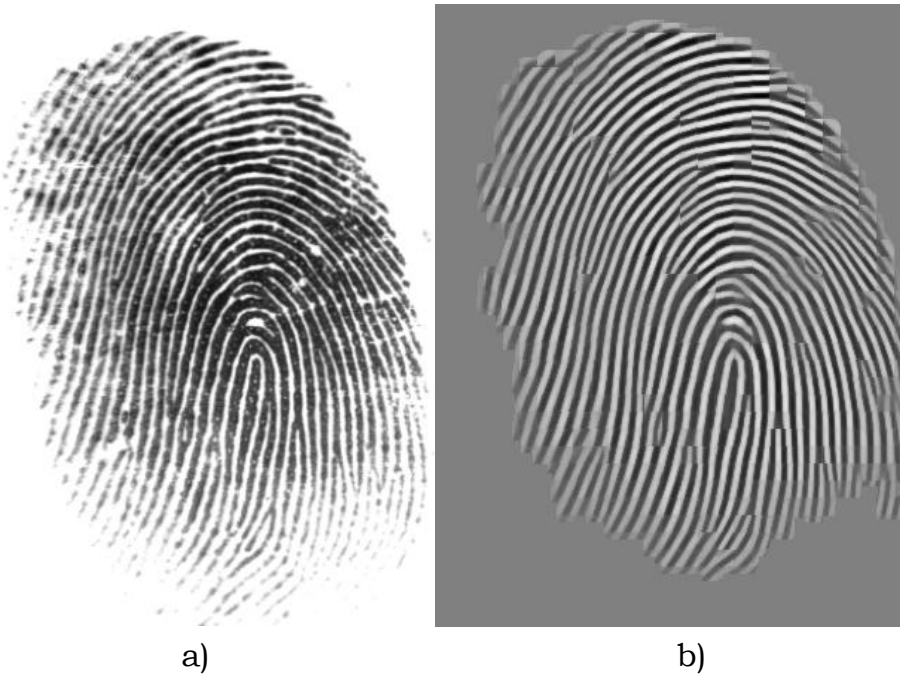

a)                                    b)

*Figure 4.3. a) Fingerprint image 108_5 from DB1_B in FVC2000,*
*b) Enhanced version of fingerprint image 108_5.*

## 4.4.1.2.    Get core point of the enhanced fingerprint

Core point MATLAB code is available at [58] and translated from Italian to English in (Appendix A), where the idea of determining the reference point is taken from [21], which is described as follows:

The reference point is defined as "the point of the maximum curvature on the convex ridge [59]" which is usually located in the central area of fingerprint. The reliable detection of the position of a reference point can be accomplished by detecting the maximum curvature using complex filtering methods [60].

They apply complex filters to ridge orientation field image generated from original fingerprint image. The reliable detection of reference point with the complex filtering methods is summarized below:

1. For each overlapping block in an image;
   a) Generate a ridge orientation image with the same method in STFT analysis;
   b) Apply the corresponding complex filter $h = (x + iy)^m g(x, y)$ centered at the pixel orientation in the orientation image, where m and $g(x, y) = \exp\{-((x^2 + y^2)/2\sigma^2))\}$ indicate the order of the complex filter and a Gaussian window, respectively;
   c) For $m = 1$ , the filter response of each block can be obtained by a convolution,

$$h * O(x, y) = g(y) * ((xg(x))^t * O(x,y)) + ig(x)^t * ((yg(y) * O(x, y)))$$
   where $O(x, y)$ represents the pixel orientation image;
2. Reconstruct the filtered image by composing filtered blocks.

The maximum response of the complex filter in the filtered image can be considered as the reference point. Since there is only one output, the unique output point is taken as the reference point (core point).

### 4.4.1.3.    Minutiae extraction

To extract minutiae from the enhanced fingerprint image, the minutiae extraction method [26], explained in section 2.1.6, is used. Hence we have three information for each minutia: x and y coordinates of its location, type of minutia (type1 if it is a termination, type2 if it is a bifurcation).

The result of this minutiae extraction stage is shown in Figure 4.4, where Figure 4.4.a is the same as Figure 4.3.b. Figure 4.4.b shows the termination minutiae in circles and the bifurcation minutiae in diamonds together with the core point of the fingerprint with an asterisk.

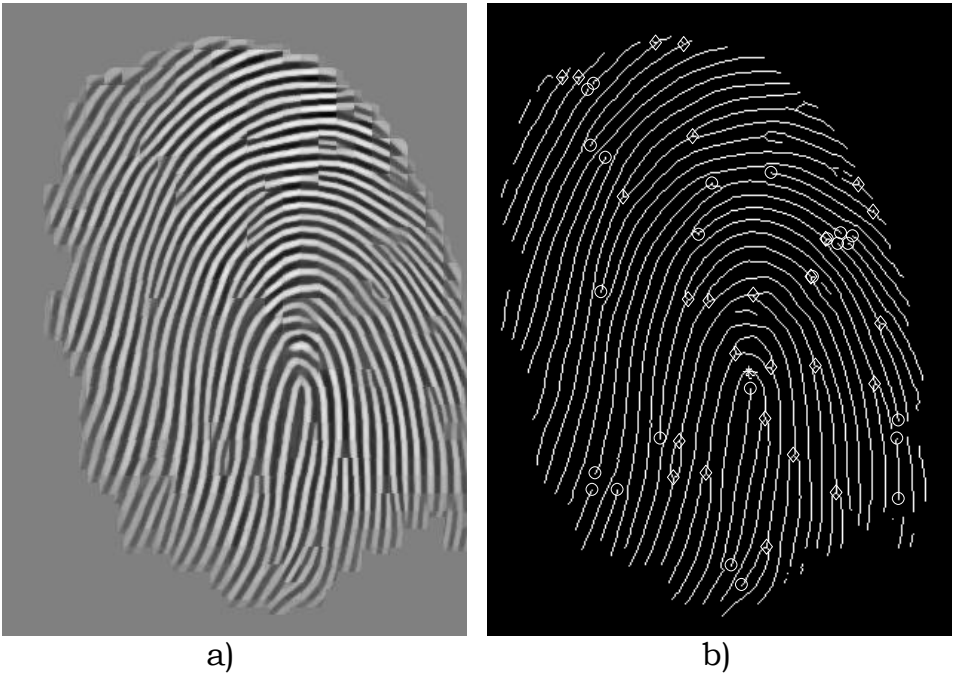

a)                                                    b)

*Figure 4.4. a) Enhanced Fingerprint image 108_5 from DB1_B in FVC2000, b) core point (asterisk), terminations (circles) and bifurcations (diamonds)*

### 4.4.1.4.    Construct the minutiae table

From the output of the minutiae extraction step, the proposed minutiae table is constructed as following:

    a. Get all minutiae locations together with their types.

    b. Using Euclidean distances, get the distances between all the minutiae and the core point of the fingerprint: if the core location is at $(x_c, y_c)$ and a minutia location is at $(x, y)$, the Euclidean distance between them will be:

$$\sqrt{(x-x_c)^2+(y-y_c)^2}$$

    c. Construct tracks of 10×n pixels wide (where n=1...max_distance/10) centered at the core point until all minutiae are exhausted, the track width is chosen to be 10 as the average distance (in pixels) between two consecutive ridges is 10 pixels, this is achieved in the fingerprint 108_5 in Figure 4.4.a. where it is of 96dpi resolution.

    d. In each track, count the number of existed minutiae of type1 and the number of existed minutiae of type2.

    e. Construct a table of two columns, column 1 for type1 minutiae and column 2 for type2 minutiae, having number of rows equal number of found tracks.

    f. In the first row, record the number of minutiae of type1 found in the first track in the first column, and record the number of minutiae of type2 found in the first track in the second column.

    g. Repeat last step for the remaining tracks of the fingerprint until all tracks are processed, and then store the minutiae table in the database.

The resulted minutiae table from Figure 4.4.b is as shown in Table 4.1. The first column is for illustration only but in MATLAB, it is not existed and it is used as the index for each row of the minutiae table consisting of just two columns.

To validate the table's data, the total number of the minutiae of both types: termination and bifurcation, in Figure 4.4.b is found to be the total summation of both columns of minutiae table which is equal to 51 minutiae.

| Track number | # of type1 minutiae | # of type2 minutiae |
|---|---|---|
| 1 | 0 | 0 |
| 2 | 1 | 2 |
| 3 | 0 | 0 |
| 4 | 0 | 1 |
| 5 | 0 | 1 |
| 6 | 0 | 2 |
| 7 | 0 | 3 |
| 8 | 2 | 2 |
| 9 | 0 | 2 |
| 10 | 1 | 2 |
| 11 | 6 | 1 |
| 12 | 3 | 1 |
| 13 | 3 | 0 |
| 14 | 4 | 1 |
| 15 | 1 | 2 |
| 16 | 0 | 1 |
| 17 | 0 | 0 |
| 18 | 1 | 0 |
| 19 | 1 | 0 |
| 20 | 0 | 0 |
| 21 | 0 | 0 |
| 22 | 1 | 0 |
| 23 | 1 | 4 |
| 24 | 0 | 1 |

*Table 4.1. The minutiae table of fingerprint 108_5 from DB1_B in FVC2000*

### 4.4.2.    Verification phase

4.4.2.1.            Capture the fingerprint to be verified

The user who is claiming that he is (e.g. M) will put his finger on the scanner to be captured at the application he wants to access. Now, a fingerprint will be available to be verified to check if he is actually M so that he will be accepted or he is not so that he will be rejected.

### 4.4.2.2.    Construct the minutiae table of that fingerprint

The same steps of the enrollment explained in section 4.4.1 will be applied on the fingerprint obtained from the previous step. Now, a minutiae table T corresponding to the fingerprint under test will be built.

### 4.4.2.3.    Get all corresponding minutiae tables from the database

In FVC2000 [30], each fingerprint has eight prints, so to verify a certain input fingerprint, all the corresponding minutiae tables of different prints of that claimed fingerprint stored in the database must be fetched.

Taking as an example the fingerprint 108_7 (see Figure 4.5.a) taken from DB1_B in FVC2000 to be verified and applying the same steps of enrollment, the results are shown in (Figure 4.5) where Figure 4.5.b shows the enhanced version of Figure 4.5.a, and Figure 4.5.c shows its thinned version together with the terminations in circles, the bifurcations in diamonds and finally the core point in asterisk. As can be shown, because the image is of poor quality, the number of minutiae is so different.

Now the minutiae table is ready to be constructed as shown in Table 4.2. Summing all the minutiae of both types in Table 4.2, it results in 73 minutiae which is different from the number before for fingerprint 108_5 which was 51 minutiae.

Fingerprints 108_1, 108_2, 108_3, 108_4, 108_6, and 108_8 taken from DB1_B in FVC2000 are shown in Figure 4.6, and their corresponding minutiae tables are shown in Table 4.3, Table 4.4, Table 4.5, Table 4.6, Table 4.7, and Table 4.8 respectively. Fingerprint 108_5 is already shown in Figure 4.3.a and its corresponding minutiae table is shown in Table 4.1.

## 4.4.2.4.    Calculate the absolute differences between minutiae table of the input fingerprint and all minutiae tables of the claimed fingerprint

Now, the absolute differences between minutiae table corresponding to fingerprint 108_7 and all minutiae tables corresponding to fingerprints 108_1, 108_2, 108_3, 108_4, 108_5, 108_6, and 108_8 are calculated. Because the sizes (number of rows) of minutiae tables are not equal, the minimum size must be determined to be able to perform the absolute subtraction on the same size for different tables. The minimum number of tracks found in DB1_B is 14.

| Track number | # of type1 minutiae | # of type2 minutiae |
|---|---|---|
| 1 | 1 | 0 |
| 2 | 1 | 2 |
| 3 | 1 | 0 |
| 4 | 0 | 0 |
| 5 | 1 | 2 |
| 6 | 1 | 1 |
| 7 | 6 | 1 |
| 8 | 5 | 1 |
| 9 | 2 | 1 |
| 10 | 4 | 0 |
| 11 | 3 | 1 |
| 12 | 2 | 0 |
| 13 | 5 | 1 |
| 14 | 1 | 1 |
| 15 | 3 | 1 |
| 16 | 2 | 1 |
| 17 | 0 | 2 |
| 18 | 2 | 0 |
| 19 | 4 | 1 |
| 20 | 4 | 1 |
| 21 | 3 | 2 |
| 22 | 1 | 0 |
| 23 | 1 | 1 |

*Table 4.2. The minutiae table of fingerprint 108_7 from DB1_B in FVC2000*

So, only the first 14 rows of each minutiae table will be considered during the absolute differences calculation. Table 4.9, Table 4.10, Table 4.11, and Table 4.12 shows the absolute differences between the minutiae table of fingerprint 108_7 and the minutiae tables of fingerprints 108_1, 108_2, 108_3, 108_4, 108_5, 108_6 and 108_8.

## 4.4.2.5. Get the summation of each column in each difference table

At the bottom of each of Table 4.9, Table 4.10, Table 4.11, and Table 4.12, in the rows titled "sum", the summation of each column in each difference table is drawn, now we have seven summations for type1 columns, and other seven summations for type2 columns.

## 4.4.2.6. Get the geometric mean of the resulted summations for both of type1 and type2

Get the geometric mean of the summations of type1 columns in all difference tables: Table 4.9, Table 4.10, Table 4.11, and Table 4.12.

The geometric mean, in mathematics, is a type of mean or average, which indicates the central tendency or typical value of a set of numbers [61]. It is similar to the arithmetic mean, which is what most people think of with the word "average", except that the numbers are multiplied and then the nth root (where n is the count of numbers in the set) of the resulting product is taken.

$$gm1 = \sqrt[7]{25 \times 24 \times 24 \times 26 \times 27 \times 17 \times 69} = 27.488$$

Get the geometric mean of the summations of type2 columns in all difference tables.

$$gm2 = \sqrt[7]{9 \times 9 \times 10 \times 15 \times 11 \times 7 \times 6} = 9.2082$$

Check the values of gm1 and gm2:

If gm1 <= threshold1 and gm2 <= threshold2 then
      the user is genuine and accept him
else
      the user is imposter and reject him

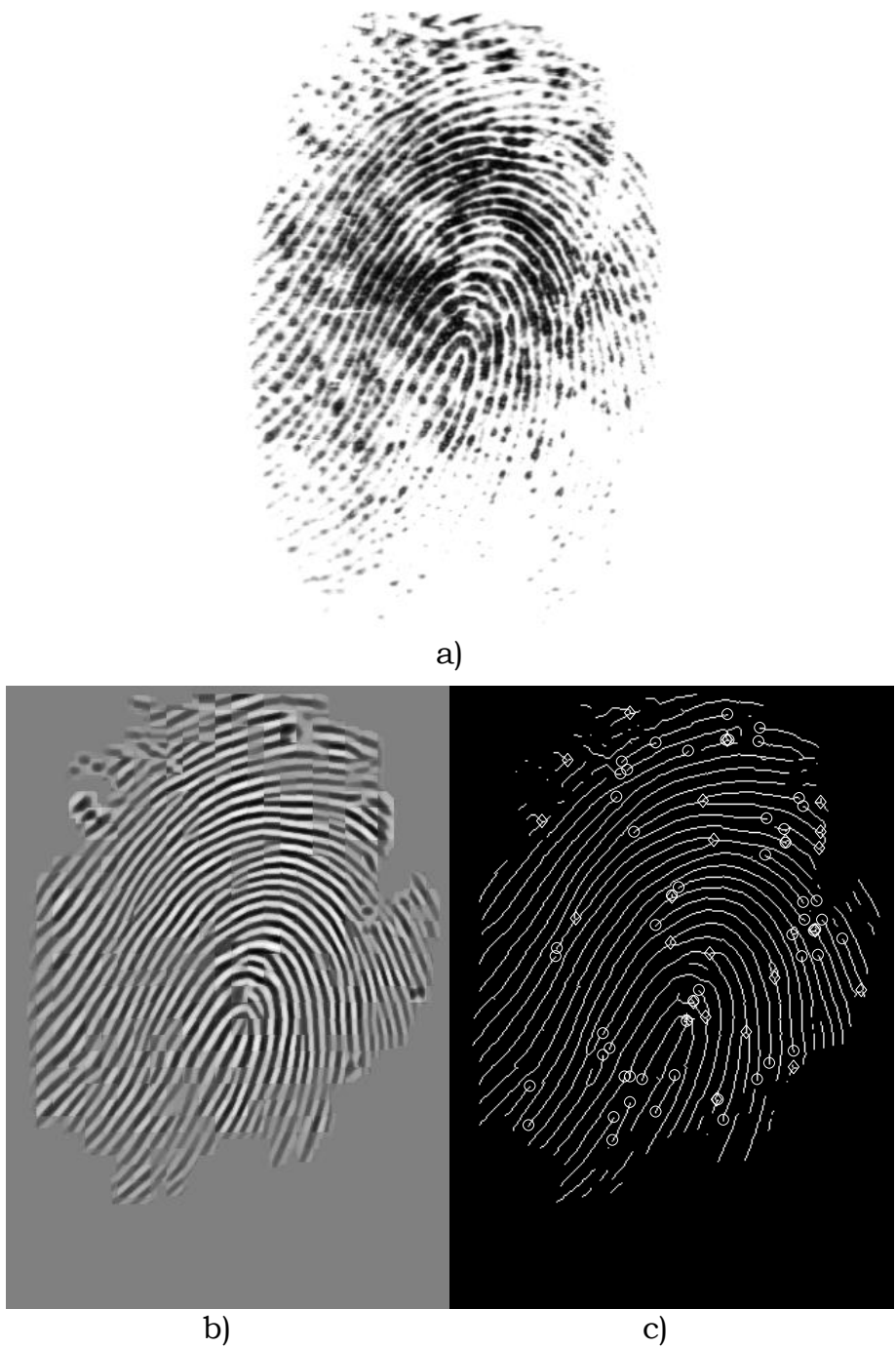

a)

b)                          c)

*Figure 4.5. a) Fingerprint image 108_7 from DB1_B in FVC2000,*
*b) Enhanced version of fingerprint image 108_7, c) core point (asterisk),*
*terminations (circles) and bifurcations (diamonds).*

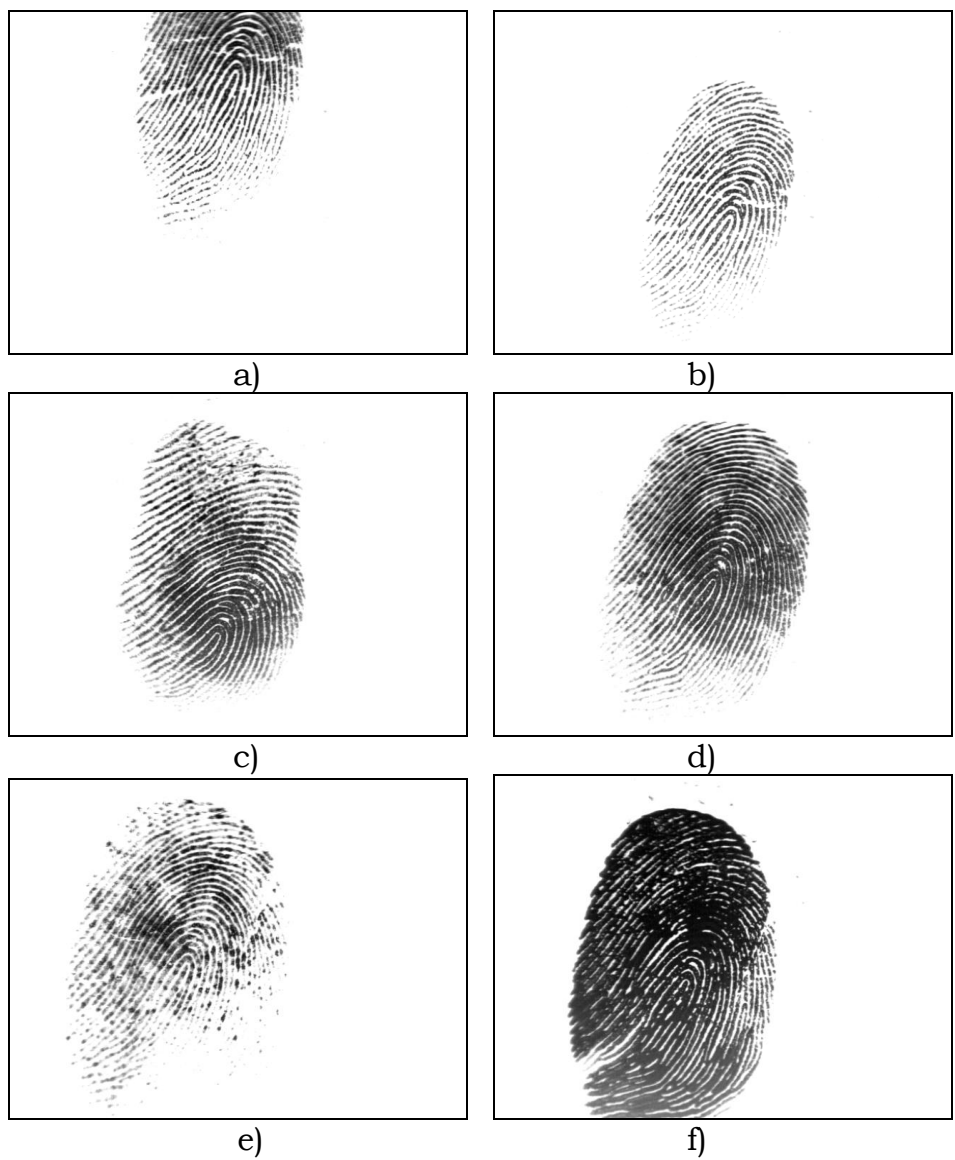

*Figure 4.6. Fingerprints taken from DB1_B in FVC2000: a) 108_1,*
*b)108_2, c)108_3, d)108_4, e)108_6, and f)108_8*

| Track number | # of type1 minutiae | # of type2 minutiae |
|---|---|---|
| 1 | 1 | 0 |
| 2 | 1 | 0 |
| 3 | 1 | 0 |
| 4 | 0 | 1 |
| 5 | 3 | 1 |
| 6 | 3 | 0 |
| 7 | 4 | 0 |
| 8 | 0 | 1 |
| 9 | 0 | 1 |
| 10 | 0 | 0 |
| 11 | 1 | 0 |
| 12 | 1 | 1 |
| 13 | 1 | 0 |
| 14 | 2 | 1 |
| 15 | 1 | 0 |
| 16 | 2 | 0 |

*Table 4.3. Minutiae table corresponding to the fingerprint 108_1.*

| Track number | # of type1 minutiae | # of type2 minutiae |
|---|---|---|
| 1 | 0 | 0 |
| 2 | 2 | 1 |
| 3 | 1 | 0 |
| 4 | 2 | 0 |
| 5 | 1 | 1 |
| 6 | 0 | 0 |
| 7 | 3 | 0 |
| 8 | 1 | 0 |
| 9 | 1 | 0 |
| 10 | 2 | 0 |
| 11 | 0 | 0 |
| 12 | 1 | 0 |
| 13 | 1 | 0 |
| 14 | 2 | 0 |
| 15 | 1 | 0 |
| 16 | 1 | 0 |
| 17 | 0 | 1 |
| 18 | 0 | 0 |

*Table 4.4. Minutiae table corresponding to the fingerprint 108_2.*

| Track number | # of type1 minutiae | # of type2 minutiae |
|---|---|---|

| 1 | 3 | 0 |
|---|---|---|
| 2 | 1 | 0 |
| 3 | 0 | 0 |
| 4 | 0 | 1 |
| 5 | 1 | 3 |
| 6 | 1 | 1 |
| 7 | 2 | 0 |
| 8 | 2 | 1 |
| 9 | 3 | 2 |
| 10 | 0 | 0 |
| 11 | 0 | 0 |
| 12 | 0 | 2 |
| 13 | 2 | 1 |
| 14 | 2 | 0 |
| 15 | 0 | 0 |
| 16 | 0 | 0 |
| 17 | 0 | 0 |
| 18 | 0 | 1 |
| 19 | 1 | 0 |
| 20 | 0 | 0 |
| 21 | 0 | 0 |
| 22 | 4 | 0 |
| 23 | 1 | 0 |
| 24 | 0 | 0 |
| 25 | 1 | 1 |
| 26 | 1 | 1 |
| 27 | 3 | 1 |
| 28 | 1 | 1 |

*Table 4.5. Minutiae table corresponding to the fingerprint 108_3.*

| Track number | # of type1 minutiae | # of type2 minutiae |
|---|---|---|
| 1 | 0 | 0 |
| 2 | 3 | 0 |
| 3 | 3 | 0 |
| 4 | 2 | 0 |
| 5 | 3 | 1 |
| 6 | 5 | 2 |

| 7 | 3 | 0 |
|---|---|---|
| 8 | 3 | 0 |
| 9 | 4 | 2 |
| 10 | 5 | 2 |
| 11 | 5 | 2 |
| 12 | 2 | 1 |
| 13 | 2 | 3 |
| 14 | 1 | 3 |
| 15 | 0 | 2 |
| 16 | 1 | 1 |
| 17 | 1 | 0 |

*Table 4.6. Minutiae table corresponding to the fingerprint 108_4.*

| Track number | # of type1 minutiae | # of type2 minutiae |
|---|---|---|
| 1 | 1 | 0 |
| 2 | 1 | 1 |
| 3 | 0 | 0 |
| 4 | 2 | 1 |
| 5 | 2 | 1 |
| 6 | 1 | 1 |
| 7 | 2 | 0 |
| 8 | 8 | 1 |
| 9 | 3 | 0 |
| 10 | 3 | 0 |
| 11 | 1 | 0 |
| 12 | 3 | 1 |
| 13 | 4 | 1 |
| 14 | 1 | 1 |
| 15 | 4 | 0 |
| 16 | 1 | 1 |
| 17 | 2 | 1 |
| 18 | 3 | 0 |
| 19 | 0 | 1 |
| 20 | 0 | 0 |
| 21 | 1 | 0 |
| 22 | 1 | 1 |

*Table 4.7. Minutiae table corresponding to the fingerprint 108_6.*

| Track number | # of type1 minutiae | # of type2 minutiae |
|---|---|---|
| 1 | 2 | 0 |
| 2 | 3 | 2 |
| 3 | 0 | 0 |
| 4 | 2 | 0 |
| 5 | 5 | 0 |
| 6 | 7 | 1 |
| 7 | 6 | 0 |
| 8 | 6 | 0 |
| 9 | 7 | 1 |
| 10 | 10 | 0 |
| 11 | 12 | 1 |
| 12 | 18 | 2 |
| 13 | 15 | 1 |
| 14 | 7 | 1 |
| 15 | 13 | 2 |
| 16 | 9 | 5 |
| 17 | 6 | 3 |
| 18 | 7 | 0 |
| 19 | 4 | 2 |
| 20 | 2 | 3 |
| 21 | 6 | 1 |
| 22 | 1 | 0 |
| 23 | 0 | 1 |
| 24 | 0 | 1 |

*Table 4.8. Minutiae table corresponding to the fingerprint 108_8.*

| Track number | abs(108_7-108_1) | | abs(108_7-108_2) | |
|---|---|---|---|---|
| | # of type1 minutiae | # of type2 minutiae | # of type1 minutiae | # of type2 minutiae |
| 1 | 0 | 0 | 1 | 0 |
| 2 | 0 | 2 | 1 | 1 |
| 3 | 0 | 0 | 0 | 0 |
| 4 | 0 | 1 | 2 | 0 |
| 5 | 2 | 1 | 0 | 1 |
| 6 | 2 | 1 | 1 | 1 |

| 7 | 2 | 1 | 3 | 1 |
| 8 | 5 | 0 | 4 | 1 |
| 9 | 2 | 0 | 1 | 1 |
| 10 | 4 | 0 | 2 | 0 |
| 11 | 2 | 1 | 3 | 1 |
| 12 | 1 | 1 | 1 | 0 |
| 13 | 4 | 1 | 4 | 1 |
| 14 | 1 | 0 | 1 | 1 |
| Sum | 25 | 9 | 24 | 9 |

*Table 4.9. Absolute differences between minutiae table of fingerprint*
*108_7 and both minutiae tables of fingerprints 108_1, 108_2.*

| Track number | abs(108_7-108_3) | | abs(108_7-108_4) | |
|---|---|---|---|---|
| | # of type1 minutiae | # of type2 minutiae | # of type1 minutiae | # of type2 minutiae |
| 1 | 2 | 0 | 1 | 0 |
| 2 | 0 | 2 | 2 | 2 |
| 3 | 1 | 0 | 2 | 0 |
| 4 | 0 | 1 | 2 | 0 |
| 5 | 0 | 1 | 2 | 1 |
| 6 | 0 | 0 | 4 | 1 |
| 7 | 4 | 1 | 3 | 1 |
| 8 | 3 | 0 | 2 | 1 |
| 9 | 1 | 1 | 2 | 1 |
| 10 | 4 | 0 | 1 | 2 |
| 11 | 3 | 1 | 2 | 1 |
| 12 | 2 | 2 | 0 | 1 |
| 13 | 3 | 0 | 3 | 2 |
| 14 | 1 | 1 | 0 | 2 |
| Sum | 24 | 10 | 26 | 15 |

*Table 4.10. Absolute differences between minutiae table of fingerprint*
*108_7 and both minutiae tables of fingerprints 108_3, 108_4.*

| Track number | abs(108_7-108_5) | | abs(108_7-108_6) | |
|---|---|---|---|---|
| | # of type1 minutiae | # of type2 minutiae | # of type1 minutiae | # of type2 minutiae |
| 1 | 1 | 0 | 0 | 0 |

| | | | |
|---:|---:|---:|---:|
| 2 | 0 | 0 | 0 | 1 |
| 3 | 1 | 0 | 1 | 0 |
| 4 | 0 | 1 | 2 | 1 |
| 5 | 1 | 1 | 1 | 1 |
| 6 | 1 | 1 | 0 | 0 |
| 7 | 6 | 2 | 4 | 1 |
| 8 | 3 | 1 | 3 | 0 |
| 9 | 2 | 1 | 1 | 1 |
| 10 | 3 | 2 | 1 | 0 |
| 11 | 3 | 0 | 2 | 1 |
| 12 | 1 | 1 | 1 | 1 |
| 13 | 2 | 1 | 1 | 0 |
| 14 | 3 | 0 | 0 | 0 |
| Sum | 27 | 11 | 17 | 7 |

*Table 4.11. Absolute differences between minutiae table of fingerprint 108_7 and both minutiae tables of fingerprints 108_5, 108_6.*

| Track number | abs(108_7-108_8) | |
|---:|---:|---:|
| | # of type1 minutiae | # of type2 minutiae |
| 1 | 1 | 0 |
| 2 | 2 | 0 |
| 3 | 1 | 0 |
| 4 | 2 | 0 |
| 5 | 4 | 2 |
| 6 | 6 | 0 |
| 7 | 0 | 1 |
| 8 | 1 | 1 |
| 9 | 5 | 0 |
| 10 | 6 | 0 |
| 11 | 9 | 0 |
| 12 | 16 | 2 |
| 13 | 10 | 0 |
| 14 | 6 | 0 |
| Sum | 69 | 6 |

*Table 4.12. Absolute differences between minutiae table of fingerprint 108_7 and minutiae table of fingerprint 108_8.*

### 4.4.3.    **Performance Evaluation**

In section 2.3.6, the recognition rate is briefly analyzed, where to evaluate any matching algorithm performance, some important quantities have to be measured such as [30]:

- False NonMatch Rate (FNMR) often referred to as False Rejection Rate (FRR)
- False Match Rate (FMR) often referred to as False Acceptance Rate (FAR)
- Equal Error Rate (EER)
- ZeroFNMR
- ZeroFMR
- Average enroll time
- Average match time

Because the presence of the fingerprint cores and deltas is not guaranteed in FVC2000 since no attention was paid on checking the correct finger position on the sensor [30], and the core point detection is the first step in the proposed matching algorithm, another group of fingerprints have been captured experimentally; this group contains the right forefinger of 20 different persons, each is captured three times, having 60 different fingerprint images. They are numbered as follows: 101_1, 101_2, 101_3, 102_1, ...., 120_1, 120_2, 120_3.  All these fingerprints have a core point. This group will be tested first and then the four databases DB1, DB2, DB3, and DB4 (from FVC2000) will be tested afterwards. All steps used to evaluate the performance of the proposed algorithm are implemented and shown in the last M-file in Appendix B.

4.4.3.1.    Calculate False NonMatch Rate (FNMR) or False Rejection Rate (FRR)

Each fingerprint template (minutiae table) $T_{ij}$, i=1...20, j=1...3, is matched against the fingerprint images (minutiae tables) of $F_i$, and the corresponding Genuine Matching Scores (GMS) are stored. The number of matches (denoted as NGRA – Number of Genuine Recognition Attempts [30]) is $20 \times 3 = 60$.

Now FRR(t) curve will be easily computed from GMS distribution for different threshold values. Given a threshold $t$, FRR($t$) denotes the percentage of GMS $\geq t$. Here, because the input fingerprint is verified if it gives less difference values between corresponding minutiae tables, lower scores are associated with more closely matching images. This is the opposite of most fingerprint matching algorithms in fingerprint verification, where higher scores are associated with more closely matching images. So, the FRR(t) (or FNMR(t)) curve will start from the left not from the right as seen in Figure 2.21. Also, it is worth to be noted that the curve of FRR(t) will be a 2D surface (FRR($t_1$, $t_2$)) because there are two thresholds as mentioned in previous section.

For example, consider the fingerprint 101_1, or any slightly different version of it, is to be matched with its other prints 101_2, 101_3, this is considered as a genuine recognition attempt because they are all from the same fingerprint. Figure 4.7 shows the fingerprint 101_1 together with its enhanced thinned version where core point is shown in a solid circle, terminations are shown with circles, and bifurcations are shown with diamonds.

Also, Figure 4.8 and Figure 4.9 show fingerprints 101_2 and 101_3 respectively together with their enhanced thinned versions. As an example, some noise is applied on the minutiae table of fingerprint 101_1, to act as a new user's fingerprint. Noise is a sequence of Pseudorandom integers from a uniform discrete distribution used to randomly select tracks from the minutiae table that will be changed by adding '1' to values under termination (or bifurcation) column and subtracting '1' to values under bifurcation (or termination) column in each selected track. The sequence of numbers, produced using Matlab function called "randi", is determined by the internal state of the uniform pseudorandom number generator. The number of random selected tracks is a constant ratio (30%) from the overall number of tracks in each database.

As described in Section 4.3.2, minutiae tables of fingerprints 101_1, 101_2, and 101_3 will be fetched from the database. The minimum number of rows (tracks) in all the minutiae tables in the database under study is found to be 13, so only the first 13 rows of any minutiae table are considered during calculations. Table 4.13 shows the first 13 rows of minutiae table of fingerprint 101_1 together with the noisy version of this fingerprint.

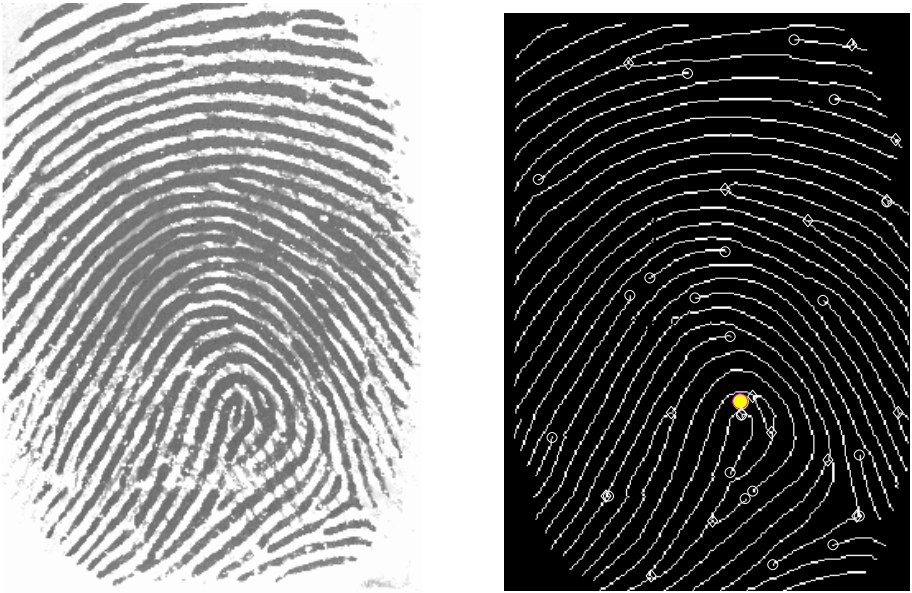

*Figure 4.7. Fingerprint 101_1 and its enhanced thinned version.*

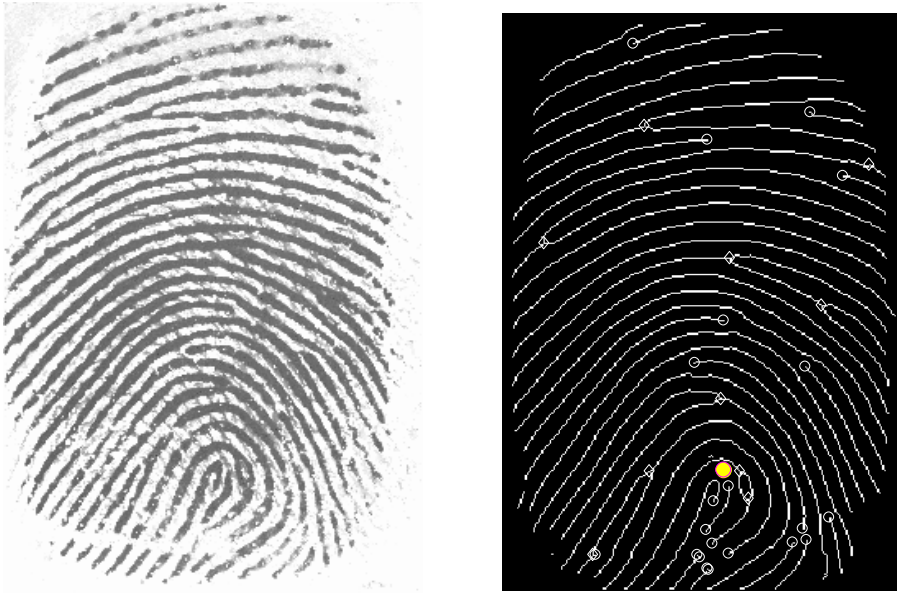

*Figure 4.8. Fingerprint 101_2 and its enhanced thinned version.*

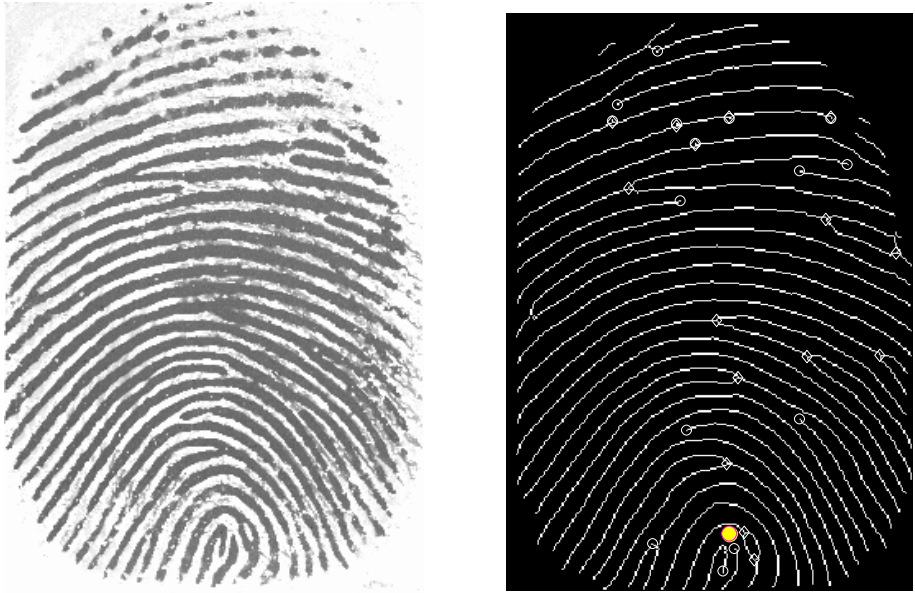

*Figure 4.9. Fingerprint 101_3 and its enhanced thinned version.*

| | Fgp 101_1 | | Noisy version of Fgp 101_1 (NFGP 101_1) | |
|---|---|---|---|---|
| Track number | # of type1 minutiae | # of type2 minutiae | # of type1 minutiae | # of type2 minutiae |
| 1 | 1 | 2 | 1 | 2 |

| 2 | 0 | 0 | 0 | 0 |
|---|---|---|---|---|
| 3 | 0 | 1 | 0 | 2 |
| 4 | 1 | 0 | 1 | 0 |
| 5 | 1 | 1 | 1 | 1 |
| 6 | 2 | 0 | 2 | 0 |
| 7 | 1 | 1 | 0 | 2 |
| 8 | 2 | 1 | 2 | 1 |
| 9 | 0 | 0 | 0 | 0 |
| 10 | 5 | 3 | 6 | 2 |
| 11 | 2 | 0 | 2 | 0 |
| 12 | 1 | 2 | 1 | 2 |
| 13 | 0 | 1 | 0 | 1 |

*Table 4.13. The minutiae table of fingerprint 101_1and its noisy version.*

Table 4.14 shows the minutiae tables of fingerprints 101_2, and 101_3.

| Track number | Fgp 101_2 | | Fgp 101_3 | |
|---|---|---|---|---|
| | # of type1 minutiae | # of type2 minutiae | # of type1 minutiae | # of type2 minutiae |
| 1 | 1 | 1 | 1 | 1 |
| 2 | 1 | 0 | 0 | 0 |
| 3 | 0 | 1 | 1 | 1 |
| 4 | 1 | 0 | 0 | 0 |
| 5 | 1 | 2 | 1 | 1 |
| 6 | 6 | 0 | 0 | 0 |
| 7 | 4 | 0 | 1 | 0 |
| 8 | 1 | 0 | 0 | 0 |
| 9 | 0 | 0 | 1 | 0 |
| 10 | 2 | 1 | 0 | 1 |
| 11 | 0 | 0 | 0 | 0 |
| 12 | 0 | 1 | 0 | 1 |
| 13 | 0 | 1 | 0 | 1 |

*Table 4.14. The minutiae tables of fgps 101_2 and 101_3.*

Now, the absolute differences tables are calculated and shown in Table 4.15 and Table 4.16. Summations of all cells of each column are shown in the last row.

| Track number | abs(N101_1-101_1) | | abs(N101_1-101_2) | |
|---|---|---|---|---|
| | # of type1 minutiae | # of type2 minutiae | # of type1 minutiae | # of type2 minutiae |
| 1 | 0 | 0 | 0 | 1 |
| 2 | 0 | 0 | 1 | 0 |
| 3 | 0 | 1 | 0 | 1 |
| 4 | 0 | 0 | 0 | 0 |
| 5 | 0 | 0 | 0 | 1 |
| 6 | 0 | 0 | 4 | 0 |
| 7 | 1 | 1 | 4 | 2 |
| 8 | 0 | 0 | 1 | 1 |
| 9 | 0 | 0 | 0 | 0 |
| 10 | 1 | 1 | 4 | 1 |
| 11 | 0 | 0 | 2 | 0 |
| 12 | 0 | 0 | 1 | 1 |
| 13 | 0 | 0 | 0 | 0 |
| Sum | 2 | 3 | 17 | 8 |

Table 4.15. Absolute differences between noisy version of minutiae table of fingerprint 101_1 and both minutiae tables of fingerprints 101_1, and 101_2.

| Track number | abs(N101_1-101_3) | |
|---|---|---|
| | # of type1 minutiae | # of type2 minutiae |
| 1 | 0 | 1 |
| 2 | 0 | 0 |
| 3 | 1 | 1 |
| 4 | 1 | 0 |
| 5 | 0 | 0 |
| 6 | 2 | 0 |
| 7 | 1 | 2 |
| 8 | 2 | 1 |
| 9 | 1 | 0 |
| 10 | 6 | 1 |
| 11 | 2 | 0 |
| 12 | 1 | 1 |
| 13 | 0 | 0 |
| Sum | 17 | 7 |

*Table 4.16. Absolute difference between noisy version of minutiae table of fingerprint 101_1 and minutiae table of fingerprint 101_3.*

The second step is to calculate the geometric mean of the sum of each of type1 and type2 absolute differences:

$$\text{gm1} = \sqrt[3]{2 \times 17 \times 17} = 8.33,$$
$$\text{gm2} = \sqrt[3]{3 \times 8 \times 7} = 5.52$$

Then, because a user is accepted if the two geometric means satisfy that:

$$\text{gm1} <= \text{threshold1}(t_1) \wedge \text{gm2} <= \text{threshold2}(t_2)$$

Using Demorgan's law, the (FNMR) or false rejection rate will be computed as follows:

$$\text{FNMR}(t_1, t_2) = \frac{\text{NGMS1s} > t_1 \vee \text{NGMS2s} > t_2}{\text{NGRA}}$$

Where NGMS1s is the number of genuine matching scores computed from gm1 values, NGMS2s is the number of genuine matching scores computed from gm2 values, and NGRA is the number of genuine recognition attempts which is 60. The threshold values, $t_1$ and $t_2$, vary from 1 to 100.

The same steps are performed for the remaining fingerprints, all 60 instances. The previous example is considered as a genuine recognition attempt as the comparison is held between a noisy version of the first of the three prints and the three true versions of them fetched from the database.

### 4.4.3.2.    Calculate False Match Rate (FMR) or False Acceptance Rate (FAR)

Each fingerprint template (minutiae table) $T_{ij}$, $i = 1...20$, $j = 1...3$ in the database is matched against the other fingerprint images (minutiae tables) $F_k$, $k \neq i$ from different fingers and the corresponding *Imposter Matching Scores* ims are stored. Number of Impostor Recognition Attempts is $(20 \times 3) \times (20 - 1) = 60 \times 19 = 1140$.

Now FAR($t_1$, $t_2$) surface will be easily computed from IMS distribution for different threshold values. Given thresholds $t_1$ and $t_2$, FAR($t_1$, $t_2$) denotes the percentage of IMS1s $<= t_1$ and IMS2s$<=t_2$. Here, because the input fingerprint is rejected if it gives high difference values between corresponding minutiae tables; higher scores are associated with mismatching images. This is the opposite of most fingerprint matching algorithms in fingerprint verification, where lower scores are associated with mismatching images. So, the FAR($t_1$, $t_2$) (or FMR($t_1$, $t_2$)) surface will start from the right not from the left as seen in Figure 2.21.

For example, consider the noisy version of fingerprint 101_1 is to be matched with another fingerprint like 103, this is considered as an imposter recognition attempt because they are from different fingers. Figure 4.7 shows the fingerprint 101_1 together with its enhanced thinned version. Also, Figure 4.10, Figure 4.11, and Figure 4.12 show fingerprints 103_1, 103_2 and 103_3 respectively together with their enhanced thinned versions.

As described in Section 4.3.2, all minutiae tables of fingerprints 103_1, 103_2, and 103_3 have to be fetched from the database. As before, because the minimum number of rows is 13, so only the first 13 rows of any minutiae table are considered during calculations. Table 4.17 shows the first 13 rows of minutiae table of fingerprint 103_1.

Table 4.18 shows the minutiae tables of fingerprints 103_2, 103_3.

Now, the absolute differences tables are calculated and shown in Table 4.19 and Table 4.20, where the noisy minutiae table of fgp 101_1 is shown in Table 4.13.

Geometric mean gm1 and gm2 are calculated as follows:

$$\text{gm}1 = \sqrt[3]{26 \times 23 \times 16} = 21.23 ,$$
$$\text{gm}2 = \sqrt[3]{14 \times 13 \times 13} = 13.33$$

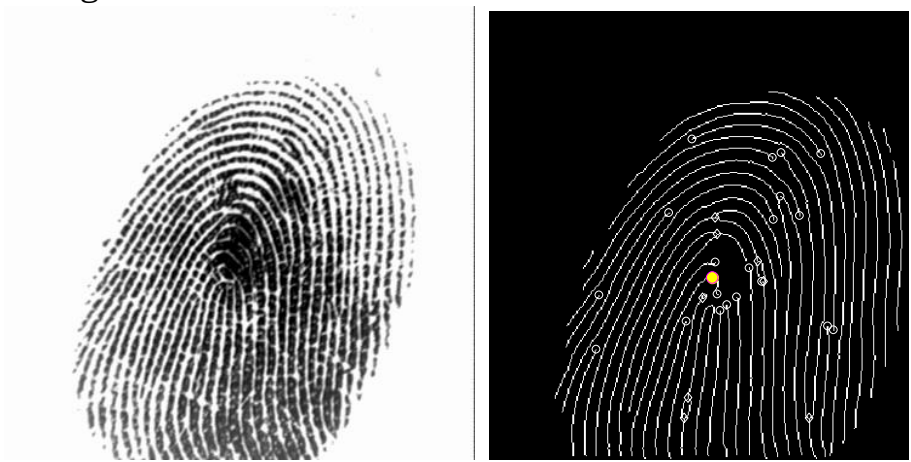

*Figure 4.10. Fingerprint 103_1 and its enhanced thinned version*

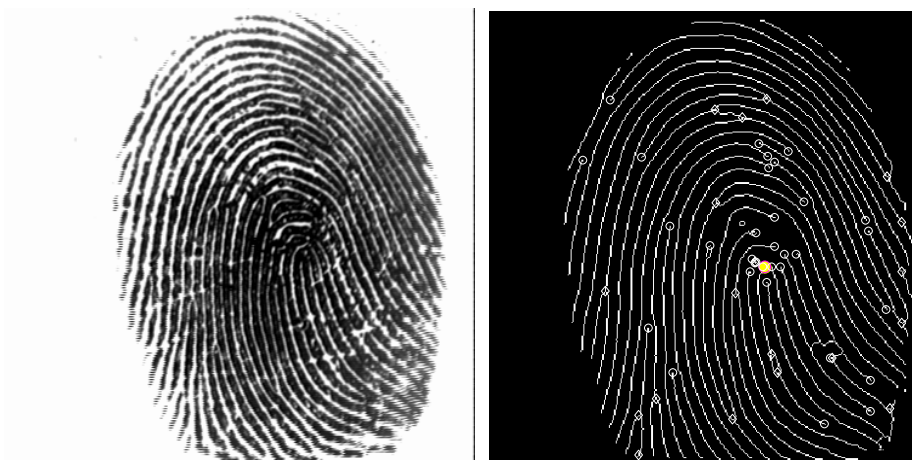

*Figure 4.11. Fingerprint 103_2 and its enhanced thinned version.*

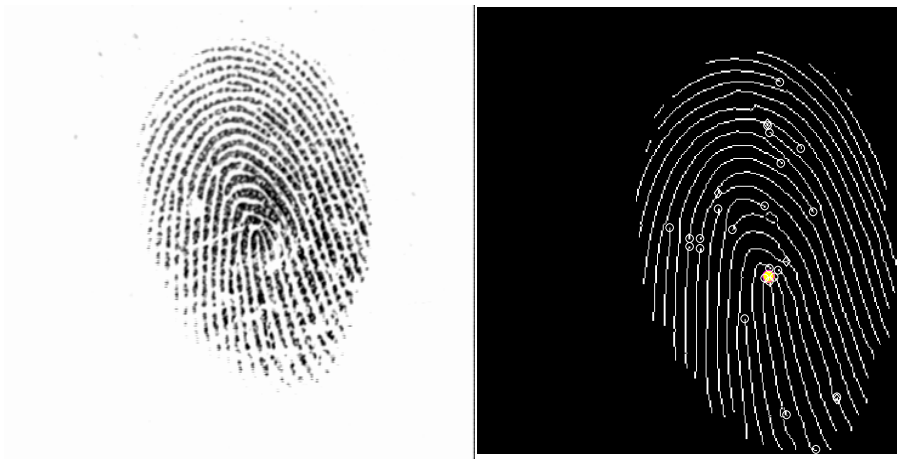

| Track number | # of type1 minutiae | # of type2 minutiae |
|---|---:|---:|
| 1 | 0 | 0 |
| 2 | 2 | 1 |
| 3 | 4 | 0 |
| 4 | 2 | 3 |
| 5 | 0 | 1 |
| 6 | 0 | 0 |
| 7 | 2 | 0 |
| 8 | 0 | 0 |
| 9 | 3 | 0 |
| 10 | 1 | 1 |
| 11 | 4 | 0 |
| 12 | 1 | 1 |
| 13 | 1 | 0 |

*Figure 4.12. Fingerprint 103_3 and its enhanced thinned version.*

*Table 4.17. The minutiae tables of fgp 103_1.*

| Track number | Fgp 103_2 | | Fgp 103_3 | |
|---|---|---|---|---|
| | # of type1 minutiae | # of type2 minutiae | # of type1 minutiae | # of type2 minutiae |
| 1 | 4 | 1 | 5 | 0 |
| 2 | 6 | 0 | 0 | 1 |
| 3 | 2 | 1 | 0 | 0 |
| 4 | 1 | 0 | 1 | 0 |

| 5 | 1 | 0 | 2 | 0 |
|---|---|---|---|---|
| 6 | 1 | 0 | 5 | 0 |
| 7 | 0 | 2 | 1 | 1 |
| 8 | 2 | 0 | 2 | 0 |
| 9 | 5 | 2 | 1 | 0 |
| 10 | 3 | 0 | 3 | 1 |
| 11 | 2 | 1 | 1 | 1 |
| 12 | 0 | 4 | 0 | 0 |
| 13 | 2 | 3 | 1 | 0 |

*Table 4.18. The minutiae tables of fgp 103_2 and 103_3.*

| | abs(N101_1-103_1) | |
|---|---|---|
| Track number | # of type1 minutiae | # of type2 minutiae |
| 1 | 1 | 2 |
| 2 | 2 | 1 |
| 3 | 4 | 2 |
| 4 | 1 | 3 |
| 5 | 1 | 0 |
| 6 | 2 | 0 |
| 7 | 2 | 2 |
| 8 | 2 | 1 |
| 9 | 3 | 0 |
| 10 | 5 | 1 |
| 11 | 2 | 0 |
| 12 | 0 | 1 |
| 13 | 1 | 1 |
| Sum | 26 | 14 |

*Table 4.19. Absolute differences between noisy minutiae table of fingerprint 101_1 and minutiae table of fingerprint 103_1.*

| | abs(N101_1-103_2) | | abs(N101_1-103_3) | |
|---|---|---|---|---|
| Track number | # of type1 minutiae | # of type2 minutiae | # of type1 minutiae | # of type2 minutiae |
| 1 | 3 | 1 | 4 | 2 |
| 2 | 6 | 0 | 0 | 1 |
| 3 | 2 | 1 | 0 | 2 |
| 4 | 0 | 0 | 0 | 0 |

| 5 | 0 | 1 | 1 | 1 |
|---|---|---|---|---|
| 6 | 1 | 0 | 3 | 0 |
| 7 | 0 | 0 | 1 | 1 |
| 8 | 0 | 1 | 0 | 1 |
| 9 | 5 | 2 | 1 | 0 |
| 10 | 3 | 2 | 3 | 1 |
| 11 | 0 | 1 | 1 | 1 |
| 12 | 1 | 2 | 1 | 2 |
| 13 | 2 | 2 | 1 | 1 |
| Sum | 23 | 13 | 16 | 13 |

*Table 4.20. Absolute differences between noisy minutiae table of fingerprint 101_1 and both minutiae tables of fingerprints 103_2, 103_3.*

The same steps are performed for the remaining fingerprints; all 60 instances (20 fingerprints, each having 3 impressions) will be matched against the other 19 fingerprints, so a total of 1140 IMSs.

FMR($t_1$, $t_2$) will be calculated as follows:

$$\text{FMR}(t_1, t_2) = \frac{\text{NIMS1s} \leq t_1 \wedge \text{NIMS2s} \leq t_2}{\text{NIRA}}$$

Where NIMS1s is the number of imposter matching scores computed from gm1 values, NIMS2s is the number of imposter matching scores computed from gm2 values, and NIRA is the number of imposter recognition attempts which is 1140. The threshold values, $t_1$ and $t_2$, vary from 1 to 70.

Both surfaces FRR($t_1$, $t_2$) and FAR($t_1$, $t_2$) are drawn in Figure 4.13. with blue and red colors respectively. The intersection between the two surfaces is drawn with a solid line used in the next section.

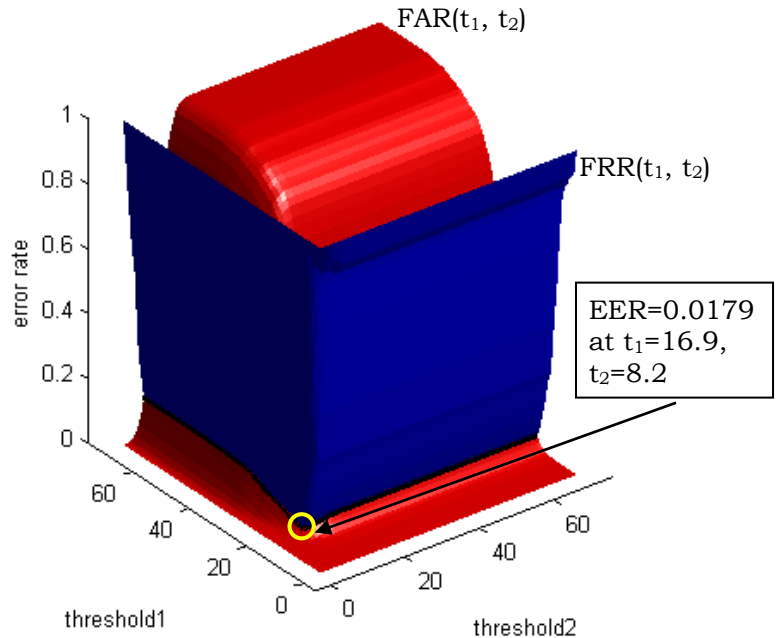

*Figure 4.13. FRR and FAR surfaces where the intersection between the two surfaces is drawn with a solid line.*

### 4.4.3.3.    Equal Error Rate EER

The Equal Error Rate is computed as the point where $FMR(t) = FNMR(t)$. From Figure 4.13., to determine the equal error rate, the intersection line between the two surfaces is drawn and then the minimum value of error rates along this line is the EER from where the values of thresholds $t_1$ and $t_2$ can be determined.

In Figure 4.13., it is shown that EER = 0.0179, where it corresponds to the threshold values of $t_1$ = 16.92 and $t_2$ = 8.21, values of thresholds are not always integers because it is not necessary for the two surfaces to intersect at integer values of thresholds.

Now to determine the integer values of thresholds that corresponds to error rates FRR and FAR, the four possible combinations of thresholds around the two given before are tested and the two values combination that gives the minimum difference between FRR and FAR (because EER is defined as the point where FRR and FAR are equal) are considered as the thresholds $t_1$ and $t_2$ that will be used for that database for any later fingerprint recognition operation.

So, the four possible combinations that threshold values $t_1$ and $t_2$ can take are: (16, 8), (16, 9), (17, 8), and (17, 9). It is found by experiment that the combination (17, 8) gives the minimum difference between FAR and FRR. So, when these thresholds are used in the proposed matching algorithm, the result is that FRR = 0.0167 and FAR = 0.0184.

True Acceptance Rate is
$$TAR = 1\text{-}FAR = 1\text{-}0.0184 = 0.9816$$

And the True Rejection Rate is
$$TRR = 1\text{-}FRR = 1\text{-}0.0167 = 0.9833$$

So, the recognition accuracy is $\approx 98\%$ when thresholds values are $t_1 = 17$ and $t_2 = 8$.

4.4.3.4.      Chi-square Quality test

Because EER is the error where false match rate equal false nonmatch rate chi-square test is
$$X^2 = FMR^2 + FNMR^2 = 2(EER)^2$$
So from Figure 4.13:
$$X^2 = 2*(0.0179)^2 = 0.00064$$

4.4.3.5.      ZeroFMR and ZeroFNMR

ZeroFMR is defined as the lowest FNMR at which no False Matches occur and ZeroFNMR as the lowest FMR at which no False Non-Matches occur[30]:

$$\text{ZeroFMR}(t) = \min_{t} \{\text{FNMR}(t) \mid \text{FMR}(t) = 0\},$$

$$\text{ZeroFNMR}(t) = \min_{t} \{\text{FMR}(t) \mid \text{FNMR}(t) = 0\}.$$

Because now the FRR(FNMR) and FAR(FMR) are drawn as 2D surfaces, all locations of FAR points having zero values are determined and the minimum value of the corresponding FRR values at these locations is the ZeroFAR. Also, to calculate the ZeroFAR value, all locations of FRR points having zero values are determined and the minimum value of the corresponding FAR values at these locations is the ZeroFRR.

From Figure 4.13., following values are drawn:
ZeroFMR = 0.3167 at $t_1$ = 14 and $t_2$ = 5,
ZeroFRR = 0.0316  at $t_1$ = 16 and $t_2$ = 10.

## 4.4.3.6.    Drawing ROC curve

A ROC (Receiving Operating Curve) is given where FNMR is plotted as a function of FMR; the curve is drawn in log-log scales for better comprehension[30].

To draw the curve in the positive portions of x- and y-axis, FMR and FNMR values are multiplied by 100 before applying the logarithm on them. Figure 4.14 shows the ROC curve of the proposed matching algorithm. To get one curve, only one column of the FAR matrix is drawn against one column of the FRR matrix, after multiplying with 100 and applying the logarithmic on both. As can be shown, the recognition performance is good by comparison with the curve of a good recognition performance system seen in Figure 2.22. It is noted that the curve in Figure 4.14 is going to the top right portion of the plotting area whereas the good recognition performance curve of Figure 2.22 is going to the bottom left portion of the plotting area, this is because in the proposed matching algorithm, lower scores are associated with matching fingerprints and higher scores are associated with mismatching fingerprints. This is the opposite of most fingerprint matching algorithms in fingerprint verification.

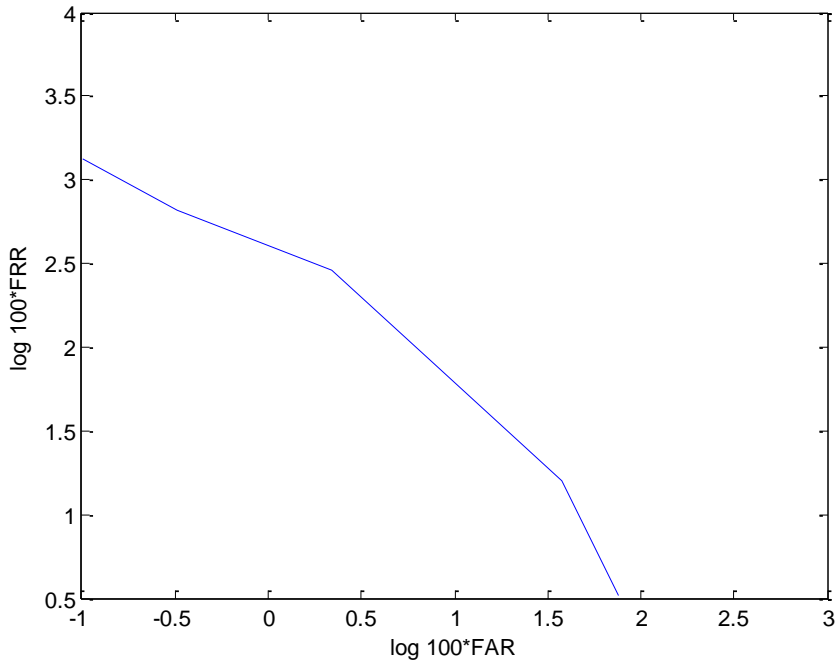

*Figure 4.14. ROC curve*

### 4.4.3.7.    Applying the proposed matching algorithm on FVC2000

Applying the proposed matching algorithm and all above steps in previous sections on the database FVC2000, it is not expected to get good results comparing with the results obtained in the previous section. This is due to the reasons mentioned in section 4.3.3.

Table 4.21 and Table 4.22 show the results of the proposed matching algorithm on FVC2000.

As shown, the recognition accuracy ranges from (1-0.2315 for DB2_B) 77% to (1 – 0.0882 for DB3_B) 91%.

Figure 4.15 and Figure 4.16 show the FRR and FAR surfaces at the left side and the ROC curves at the right side for databases DB1_A, DB1_B, DB2_A, DB2_B, DB3_A, DB3_B, DB4_A, and DB4_B respectively.

| Database | EER | $t_1$ | $t_2$ |
|---|---|---|---|
| DB1_A | 0.2109 | 18 | 6.99 |
| DB1_B | 0.1988 | 31.68 | 10 |
| DB2_A | 0.1649 | 18.48 | 8 |
| DB2_B | 0.2315 | 24.096 | 14 |
| DB3_A | 0.1454 | 28 | 12.55 |
| DB3_B | 0.0882 | 28 | 12.85 |
| DB4_A | 0.1815 | 10 | 4.88998 |
| DB4_B | 0.1206 | 14.35 | 9 |

*Table 4.21. Results of EER and its corresponding thresholds after applying the proposed matching algorithm on FVC2000*

| Database | FAR | FRR | $t_1$ | $t_2$ |
|---|---|---|---|---|
| DB1_A | 0.2113 | 0.2105 | 18 | 7 |
| DB1_B | 0.2049 | 0.1944 | 32 | 10 |
| DB2_A | 0.1835 | 0.15 | 18 | 9 |
| DB2_B | 0.2403 | 0.2375 | 24 | 15 |
| DB3_A | 0.1325 | 0.1525 | 29 | 12 |
| DB3_B | 0.0944 | 0.075 | 28 | 13 |
| DB4_A | 0.1844 | 0.1788 | 10 | 5 |
| DB4_B | 0.1153 | 0.125 | 14 | 10 |

*Table 4.22. Results of FAR and FRR and their corresponding thresholds after applying the proposed matching algorithm on FVC2000*

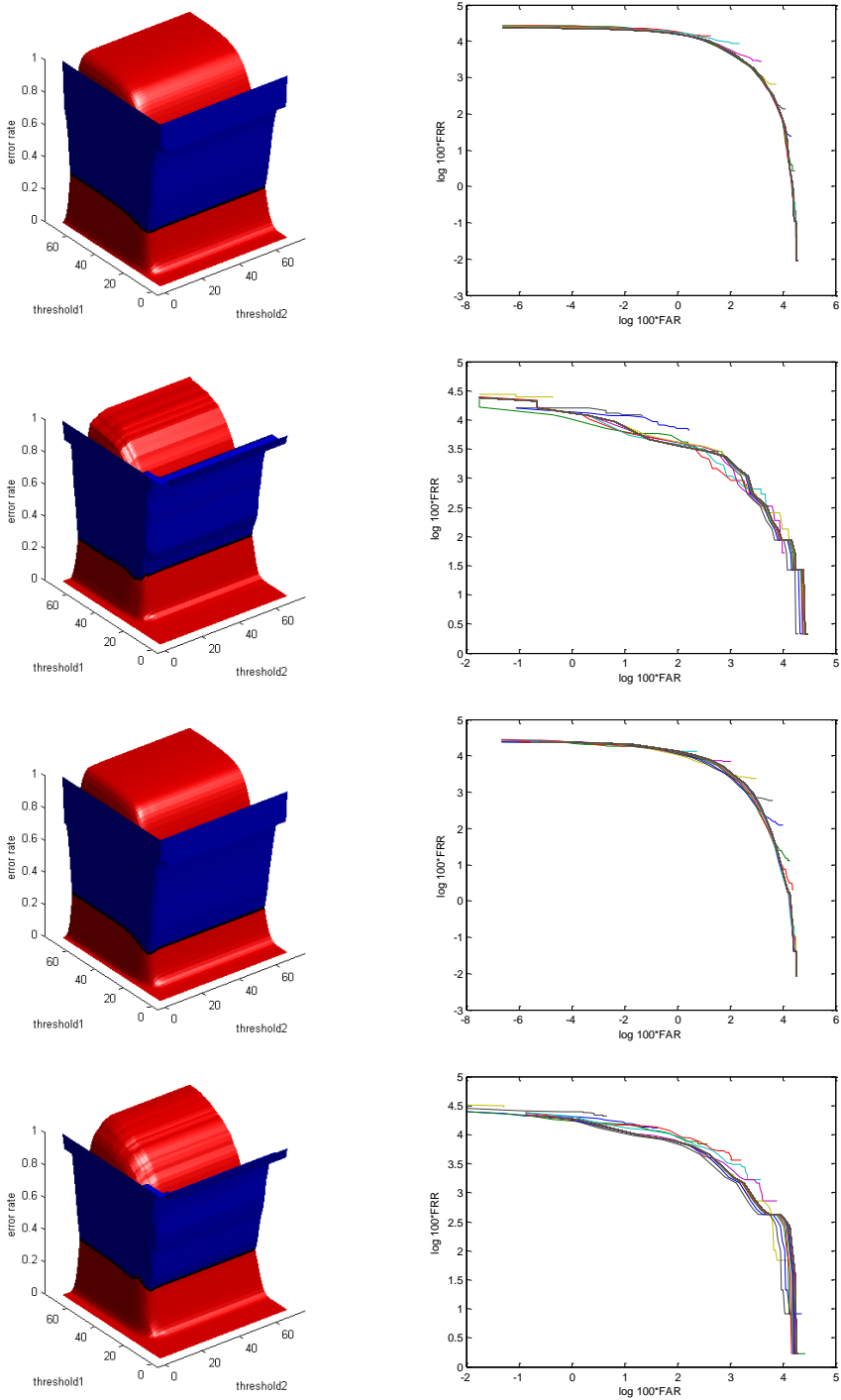

*Figure 4.15. FAR, FRR and ROC curves for DB1_A, DB1_B, DB2_A, and DB2_B respectively.*

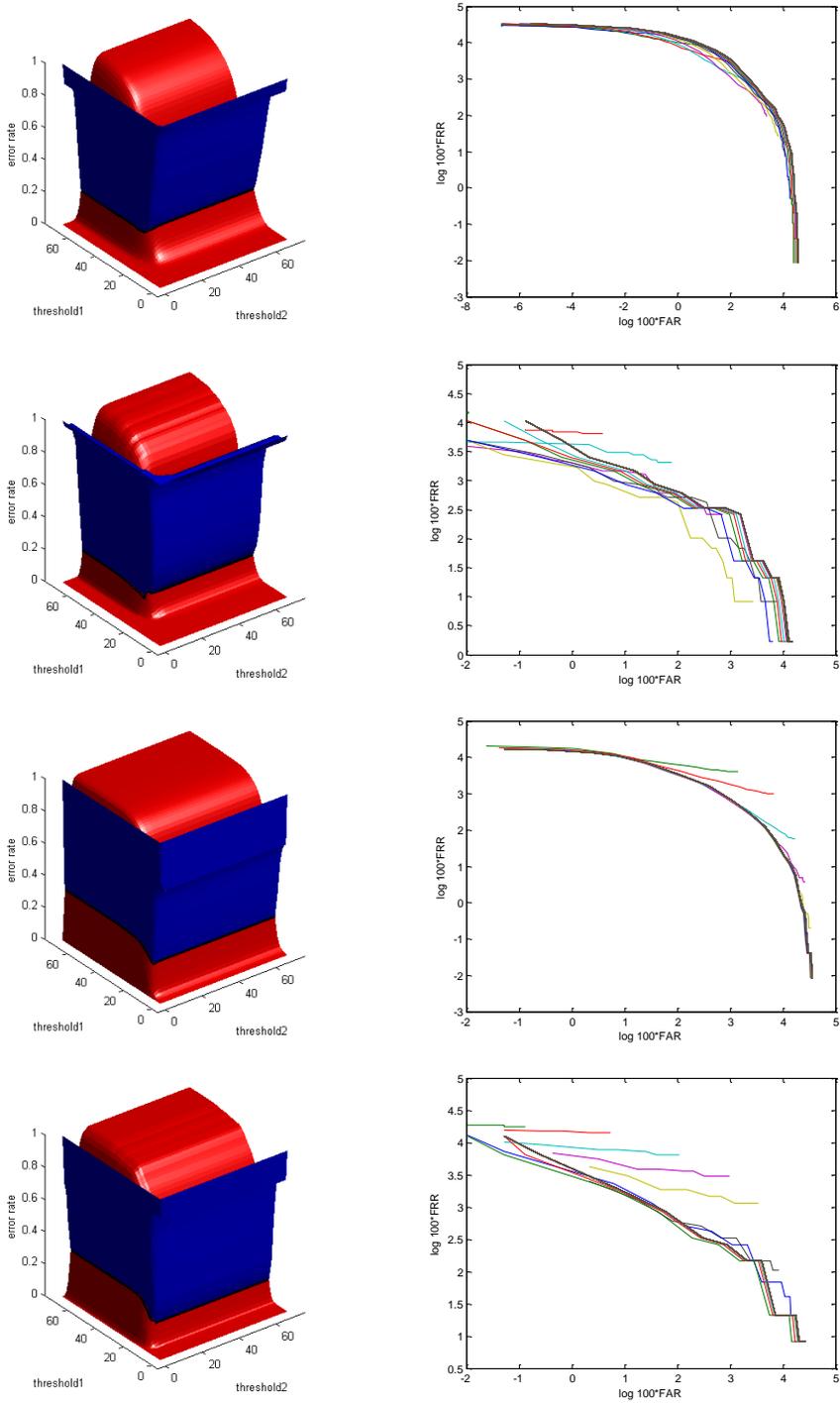

*Figure 4.16. FAR, FRR and ROC curves for DB3_A, DB3_B, DB4_A, and DB4_B respectively.*

4.4.3.8.      Calculating average enroll time

The average enroll time is calculated as the average CPU time taken by a single enrollment operation [30].

The steps of enrollment are discussed in section 4.3.1. Table 4.23 shows a detailed timing for each step in the enrollment phase. These results were implemented using MATLAB version 7.9.0529 (R2009b) as the programming platform. Programs were tested on a 2.00GHz personal computer with 1.99 GB of RAM.

Total enroll time is found to be 6.043 sec

| Step | Average time taken (sec) |
|---|---|
| Enhancement of the fingerprint | 3.7 |
| Core point detection | 0.54 |
| Thinning and minutiae extraction | 1.8 |
| Minutiae table construction | 0.003 |
| Total enroll time | 6.043 |

*Table 4.23. Enroll timing details*

4.4.3.9.      Calculating average match time

The average match time is calculated as the average CPU time taken by a single match operation between a template and a fingerprint image [30].

The steps of matching are discussed in section 4.3.2. Table 4.24 shows a detailed timing for each step in the matching phase after the construction of the minutiae table corresponding to the input fingerprint, which has been already estimated to be 6.043sec from section 4.3.3.7.

Total match time is found to be 0.00134 sec

| Step | Average time taken (sec) |
|---|---|
| Get all minutiae tables of the claimed fingerprint stored in the | 0.0011 |
| database | |
| Calculate absolute differences between the input fgp minutiae table and all minutiae tables obtained from the previous step and get the two geometric means | 0.0002 |
| Compare the resulting means with the two thresholds and decide if the user is accepted or rejected | 0.00004 |
| Total match time | 0.00134 |

*Table 4.24. Match timing details*

### 4.4.4.    Comparison with Whuzhili method

Comparing the results obtained from the proposed matching method with whuzhili matching method (reference), shows that the proposed method is more efficient.

| | Proposed | Whuzhili |
|---|---|---|
| EER | 0.0179 | 0.25 |
| X$^2$ | 0.00064 | 0.125 |
| Match time | 0.00134sec | 1.17sec |
| Data storage | 21bytes | 85bytes |
| Translation and rotation invariant | Yes | No |

Comparison between proposed and whuzhili matching methods

### 4.4.5.    Conclusion

As shown, the time for matching is extremely small as all the process is taking geometric mean of absolute differences.

There is no need for any pre-alignment which is a very complicated and time consuming process. As a result, it is translation and rotation invariant.

Also, the space needed to store any minutiae table is in the average = 21(as the average number of tracks in all database) $\times$ 2 $\times$ 4 = 168bits = 168/8bytes = 21bytes which is small in comparison with the size of 85 bytes as in [62] where the tradition method is storing locations and orientation for each minutia as a tuple <x, y, $\theta$>.

# Chapter 5
# Securing Fingerprint Systems

## 5.1. Introduction

Consider that a facility is secured with a lock. Usually, the sturdier the lock, the higher is the perceived level of security. However, even if a facility is equipped with the strongest possible lock, it is not absolutely certain that the facility cannot be broken into [26].

This is not to say that a system designer should not try her best to guard against all possible security threats. What it implies is that the type of security needed depends upon the requirements of the application. A *threat model* for an application can be defined based on what needs to be protected and from whom.

The typical threats in an application may include the following [26]:

- Denial of service (DoS): An adversary damages the system such that legitimate users can no longer access the system.
- Circumvention: An unauthorized user illegitimately gains access to the system and data. Circumvention could either be a privacy attack or a subversive attack. In a privacy attack, the intruder gets access to data that he may not be authorized to access. In a subversive attack, an intruder may manipulate the system to use it for illegal activities. For example, an intruder may break into a person's bank account and withdraw/transfer all the money.
- Repudiation: A legitimate user denies having accessed the system.
- Contamination or covert acquisition: It is possible that the means of recognition could be compromised without

the knowledge of a legitimate user and be subsequently abused.

- Collusion: In any application, some users of the system will have a super-user status that allows them to bypass the recognition component and to overrule the decision made by the system. This facility is incorporated in the system workflow to permit handling of exceptional situations, for example, processing of individuals with no fingers in a fingerprint-based recognition system. This could potentially lead to an abuse of the system by way of collusion between the super-users and the other users.
- Coercion: The genuine users could be potentially coerced (forced) to identify themselves to the system. The recognition measurements could be forcibly extracted from a genuine user to gain access to the system with concomitant privileges.

## 5.2. Points of Attack

Many of the possible attacks in fingerprint recognition systems were identified and systematically documented by Bolle, Connell, and Ratha [63]. A number of other types of attack points have been documented by Anderson [64] and Schneier [65]. Anderson studied the technical aspect of fraud involved in using ATMs and found that most of the failures were due to poor design and administration. Schneier pointed out a number of system vulnerabilities due to errors in design, implementation, and installation. Figure 5.1 shows the major modules and the dataflow paths in a fingerprint verification system. The eight potential attack points are also marked in Figure 5.1 and are elaborated below.

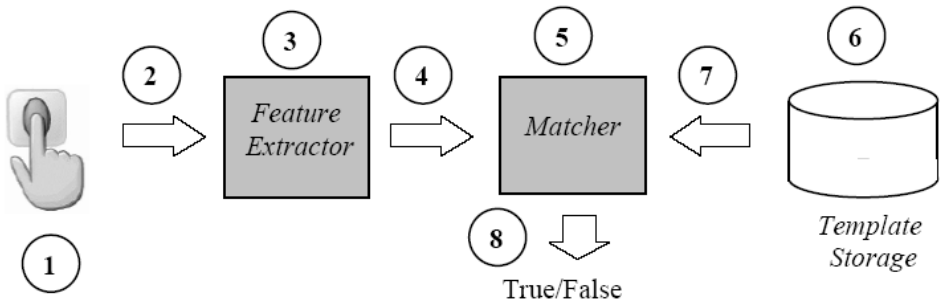

*Figure 5.1. The design of a fingerprint verification system is shown to illustrate the possible security attack points which are marked with numbers from 1 to 8.*

1. Attack at the scanner,
2. Attack on the channel between the scanner and the feature extractor,
3. Attack on the feature extraction module,
4. Attack on the channel between the scanner and the feature extractor and the matcher module,
5. Attack on the matcher module,
6. Attack on the system database,
7. Attack on the channel between the system database and matcher module, and
8. Attack on the channel between the matcher module and the application requesting verification.

Attacks 2, 4, 7, and 8 are launched against communication channels; they are similar in nature and can be collectively called "replay" attacks. The attacks 1, 3, 5, and 6 are launched against system modules; they are also similar in nature and can be collectively called "Trojan horse" attacks. Furthermore, the replay attacks are similar to those present in token- and knowledge-based authentication systems whereas the fake finger attack is unique to the fingerprint recognition system.

## 5.3. Trojan Horse Attacks

A Trojan horse attack can be launched at the scanner (e.g., a sensor emulator), feature extractor, matcher, or the system database. These attacks are launched against "entities." A Trojan horse program can disguise itself as one or more of these modules and generate false results (such as a fingerprint image, a fingerprint template, or a matching score). A Trojan horse attack at the feature extractor (see Figure 5.2) is almost the same as the attack on the system database and very similar to the attacks at the scanner and matcher.

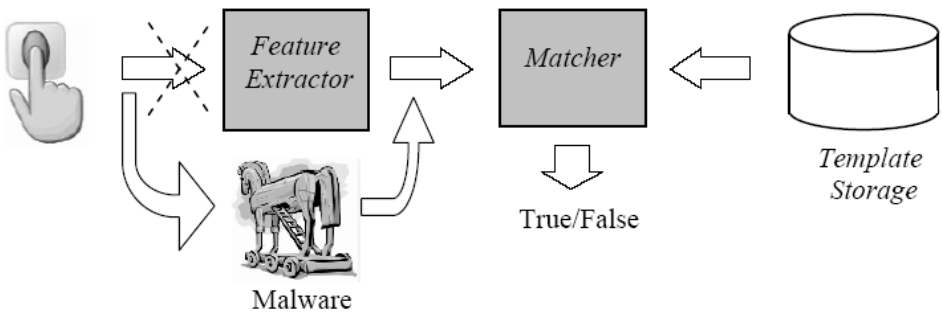

*Figure 5.2. A Trojan horse attack against the feature extraction module is shown.*

First of all, it is extremely important that the feature extraction, matcher, and system database reside at a secure and trusted location. Furthermore, the fingerprint scanner should be trustworthy and have some security capability (such as encryption) built into it. The trust between the scanner and the server should be established both ways (i.e., server trusts the scanner and the scanner trusts the server). In a remote recognition application, a client cannot always be trusted and it only acts as a medium that passes the encrypted fingerprint image from the scanner to the server for feature extraction and matching. A two-way mutual trust is established when the scanner authenticates the server and the server authenticates the scanner that is necessary in a network environment.

Smartcards have received much attention from developers/integrators of fingerprint recognition systems because of their internally protected storage and computational resources. In fact, embedding a fingerprint recognition system inside a smartcard is an effective way to prevent unauthorized users from installing Trojan horses and to avoid eavesdropping of critical information in transit. Although an entire fingerprint recognition system can be implemented onto a smartcard (including sensing element, feature extraction, matching and template database), most of the solutions proposed to date implement only some parts of the recognition system "on-card," due to the limited computational power of the onboard CPU.

## 5.4. Replay Attacks

In addition to safeguarding a fingerprint recognition system against fake finger attacks and Trojan horse attacks, protection against any replay attack is required to ensure data integrity and non-repudiation. The integrity of the data transmission must be secure all the way from the scanner to the application. This is typically achieved by cryptographic methods. In particular, encryption, digital signature, and challenge-response provide sufficient security for data integrity, non-repudiation, and replay attacks, respectively. Appendix D gives a brief description of both encryption, digital signatures, timestamp and challenge-response. In the next section, an introduction to digital watermarking is given.

### 5.4.1.    Digitally    watermarking    fingerprint images/features

A few image-based techniques similar to digital signatures (where the key is constant) and challenge-response (when a different key is sent by the server before each verification) have also been proposed specifically for fingerprints. Digital watermarking [66] is a technique that hides a secret digital pattern (called a digital watermark) in a digital image or data. A digital watermark may be visible or invisible. In a visible watermark, a visible pattern or image is embedded in the original image, but the invisible watermark does not change the visual appearance of the image. The existence of an invisible watermark can be determined only through a watermark extraction or detection algorithm. Furthermore, a digital watermarking algorithm may be robust (the watermark can withstand attacks such as compression and enhancement of an image) or fragile (the watermark is "broken" under the slightest change to the image). Typically, robust watermarking is used for copyright protection and fragile watermarking is used for data integrity.

Many watermarking techniques for gray scale and color images are studied and implemented in [67], also two different watermarking algorithms are proposed and implemented using MATLAB.

A general watermarking algorithm includes the following two parts: one is watermark embedding, the other is watermark extraction [67].

Figure 5.3 illustrates the general embedding process. Given the code function, we can get the watermarked image. The general detection process is depicted in Figure 5.4.

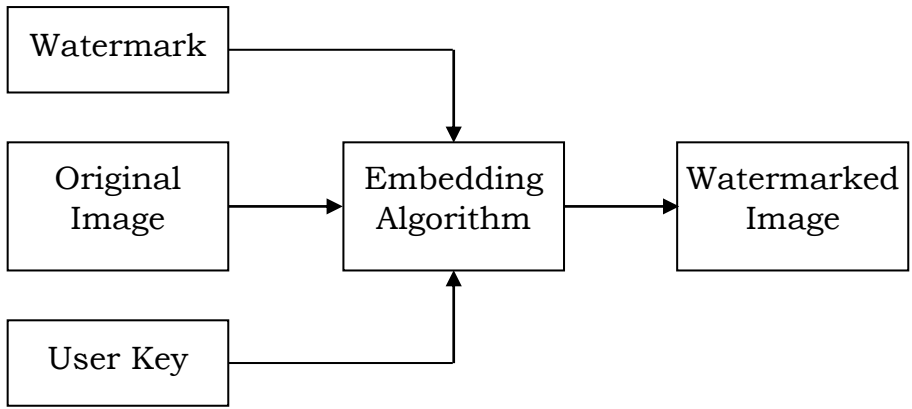

*Figure 5.3. General watermark embedding process.*

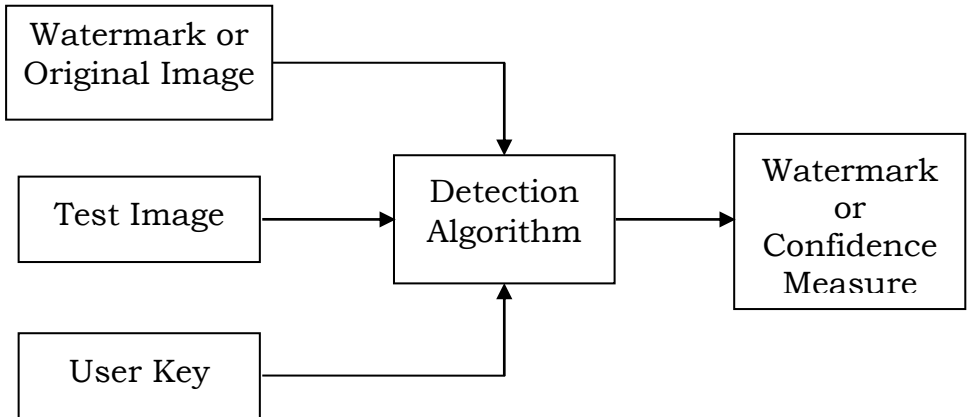

*Figure 5.4. General watermark detection process.*

Given the original image or watermark, the questioned image and the key, the watermark or some confidence measure can be obtained which indicates how likely it is for a given mark at the input to be present in the image under inspection. This kind of scheme is the simplest and most popular frame. Some other techniques do not need the original image to extract the watermark; these are called "blind" methods.

## 5.5. Previous work

Many watermarking techniques are applied to secure communication channels in fingerprint recognition systems, attacks number 2, 4, 7, and 8 in Figure 5.1.

In [68], the inherent strengths of biometrics-based authentication are outlined, the weak links in systems employing biometrics-based authentication are identified, and new solutions for eliminating some of these weak links are presented.

In [62], an application of steganography and watermarking is introduced to enable secure biometric data (e.g., fingerprints) exchange. They hide fingerprint minutiae data in a host image, which can be a synthetic fingerprint image, a face image or an arbitrary image. It is this carrier image that is transferred to the receiving party in this exchange, instead of the actual minutiae data. The hidden biometric data are extracted accurately from the carrier image using a secret key. Furthermore, when the host is a face image, this method provides an additional cue in authenticating the user. Data are hidden in the host image in an adaptive way to minimize possible degradations to that image. The proposed method can also tolerate several attacks on the carrier image.

In [69], two spatial methods are introduced in order to embed watermark data into fingerprint images, without corrupting their features. The first method inserts watermark data after feature extraction, thus preventing watermarking of regions used for fingerprint classification. The method utilizes an image adaptive strength adjustment technique which results in watermarks with low visibility. The second method introduces a feature adaptive watermarking technique for fingerprints, thus applicable before feature extraction. For both of the methods, decoding does not require original fingerprint image. Unlike most of the published spatial watermarking methods, the proposed methods provide high decoding accuracy for fingerprint images. High data hiding and decoding performance for color images is also observed.

In [70], an application of wavelet-based watermarking method to hide the fingerprint minutiae data in fingerprint images is introduced. The application provides a high security to both hidden data (i.e. fingerprint minutiae) that have to be transmitted and the host image (i.e. fingerprint). The original unmarked fingerprint image is not required to extract the minutiae data. The method is essentially introduced to increase the security of fingerprint minutiae transmission and can also used to protect the original fingerprint image.

In [71], a scheme based on the fingerprint watermark is proposed, and they tried to introduce biometrics in the watermark system. They integrated the digital watermarking technology with the fingerprint identification technology. The scheme depends upon the spatial domain, and DCT domain of multi-bits embedded watermark methods to embed and extract the information of the fingerprint characteristic. The experiment has showed that their scheme is feasible and effective.

In [72], two techniques that protects fingerprint biometric data are considered using digital watermarking techniques. Both techniques discussed are based on Discrete Wavelet Transformation (DWT). From experimental results, it can be concluded that both techniques provide adequate security to the fingerprint data without degrading visual quality. Further, the verification performance after dewatermarking is also analyzed.

Most of the above approaches are based on the secrecy/obscurity of the algorithms; security based on obscurity is weak and often unsustainable. While standard cryptographic techniques have been proven to be secure in an information-theoretic sense, the security of watermarking, obfuscation, and other similar techniques that have been specifically designed for fingerprint images have not been shown to possess proven security. So it is debatable whether they provide any benefits over standard cryptographic techniques in preventing injection of fingerprint data into the communication channel [1].

## 5.6. Application of watermarking on fingerprint recognition systems

In this section, the watermarking technique used in [62] will be described and enhanced.

### 5.6.1.    Data hiding method

The amplitude modulation-based watermarking method described in [62] is an extension of the blue channel watermarking method of Kutter *et al.* [73]. The used method includes image adaptivity and watermark strength controller along with the basic method in [73]. An earlier version of the method is presented in [16], in which the increase in data decoding accuracy related to these extensions is analyzed.

For *watermark embedding*, the minutiae data to be hidden is first converted into a bit stream. Every field of individual minutia is converted to a 9-bit binary representation. Such a representation can code integers between [0, 511] and this range is adequate for x-coordinate ([0, #rows-1]), y-coordinate ([0, #columns-1]) and orientation ([0, 359]) of a minutia. A random number generator initialized with the secret key generates locations of the host image pixels to be watermarked.

These pixels are changed according to the following equation:

$$P_{WM}(i,j) = P(i,j) + (2s-1)P_{AV}(i,j)q\left(1 + \frac{P_{SD}(i,j)}{A}\right)\left(1 + \frac{P_{GM}(i,j)}{B}\right) \quad (5.1)$$

where $P_{WM}(i,j)$ and $P(i,j)$ are values of the watermarked and original pixels at location $(i,j)$, respectively. The value of watermark bit is denoted as $s$ and watermark embedding strength is denoted as $q$, $s \in [0, 1]$, $q > 0$. $P_{AV}(i,j)$ and $P_{SD}(i,j)$ denote the average and standard deviation of pixel values in a neighborhood of location $(i,j)$ and $P_{GM}$ denotes the gradient magnitude at $(i,j)$. A and B are weights for the standard deviation and gradient magnitude, respectively.

$P_{AV}$ is calculated in a 5×5 square neighborhood and $P_{SD}$ is calculated in a 5×5 cross-shaped neighborhood. The gradient magnitude is computed via the 3×3 Sobel operator [5]. These image adaptivity terms adjust the magnitude of watermarking, by utilizing several properties of the human visual system (HVS). Using $P_{AV}(i,j)$ in modulating watermark magnitude conforms to amplitude nonlinearity of HVS. As Equation (5.1) shows, the magnitude of the change in the value of pixel $(i,j)$ caused by watermarking is higher when the $P_{AV}(i,j)$ value is high. Standard deviation and gradient magnitude terms utilize contrast/texture masking properties of HVS. Masking is the reduction of the visibility of an image component (masked signal) due to the presence of another component [74]. The changes in pixel values in highly textured and high contrast image areas are masked more strongly than changes in smooth image areas. These image adaptivity terms increase the magnitude of watermarking in image areas where such an increase does not become very visible to a human observer. This leads to more accurate decoding of the hidden data, especially in the case of attacks on host images.

Every watermark bit with value s in Equation (5.1) is embedded at multiple locations (depends on no. of embedding locations and the size of the watermark) in the host image. This redundancy increases the correct decoding rate of the embedded information. The amount of this redundancy is limited by image capacity (size) and visibility of the changes in pixel values.

In addition to the binary minutiae data, two reference bits, 0 and 1, are also embedded in the image. These reference bits help in calculating an adaptive threshold to determine the minutiae bit values during extraction.

*Watermark extraction* starts with finding the data embedding locations in the watermarked image, via the secret key used during the watermark encoding stage. For every bit embedding location, $(i, j)$, its value is estimated as the linear combination of pixel values in a 5×5 cross-shaped neighborhood of the watermarked pixels as in Equation (5.2).

$$\hat{P}(i,j) = \frac{1}{8}\left( \sum_{k=-2}^{2} P_{WM}(i+k,j) + \sum_{k=-2}^{2} P_{WM}(i,j+k) - 2P_{WM}(i,j) \right) \qquad (5.2)$$

The difference between the estimated and watermarked pixel values is calculated by Equation (5.3) as

$$\delta = P_{WM}(i,j) - \hat{P}(i,j) \qquad (5.3)$$

These differences are averaged over all the embedding locations associated with the same bit, to yield $\bar{\delta}$. For finding an adaptive threshold, these averages are calculated separately for the reference bits, 0 and 1, as $\bar{\delta}_{R0}$ and $\bar{\delta}_{R1}$, respectively. Finally, the watermark bit value $\hat{s}$ is estimated as

$$\hat{s} = \begin{cases} 1 & \text{if } \bar{\delta} > \dfrac{\bar{\delta}_{R0} + \bar{\delta}_{R1}}{2} \\ 0 & \text{otherwise.} \end{cases} \qquad (5.4)$$

The watermark extraction process can produce erroneous bits since extraction is based on an estimation procedure, which may fail to find the exact original pixel values. In order to increase the extraction accuracy, the embedding uses a controller block. This block adjusts the strength of watermarking, q, on a pixel-by-pixel basis, if there is a possibility of incorrect bit extraction. From extracted watermark bits, the minutiae data hidden in the host image is extracted. In the second application scenario, which typically includes a smart card containing a digital fingerprint image marked with the minutiae data of the fingerprint depicted in this image, an estimate of the original fingerprint image is also found via replacing the watermarked pixel values with the $\hat{P}(i, j)$ estimate calculated by Equation (5.2).

As shown, the watermark extraction method did not refer to the original image, i.e. it is a blind watermarking algorithm.

### 5.6.2.    The proposed watermarking method

Some modifications have been applied on the data hiding method used in [62] and described in section 5.6.1 above. These modifications are as follows:

1.  Instead of using *one* key for determining embedding locations in the host image, *two* keys are used to increase the security level: the first key is used for the same purpose, i.e. for determining the embedding locations; the second key is used to permutate the watermark before using it for embedding. Both keys are used as a seed for a random number generator.

2. The host image used for embedding was the *face* of the user in the method in [62], whereas the host image in the proposed method will be the *fingerprint* of the user.

3. The watermark in the previous method was the minutiae data tuples; i.e. <x, y, θ> of each minutia, now the watermark will be the minutiae table, which have been described in the proposed matching algorithm in Chapter 4, where the minutiae table was a table of two columns: the number of terminations minutiae existed in each track around the core point is recorded in the first column and the number of bifurcation minutiae existed in each track around the core point is recorded in the second column.

### 5.6.2.1. Watermark embedding algorithm

Figure 5.5 shows a flowchart of the proposed watermark embedding algorithm. In the following, a detailed explanation of the steps of the algorithm is given. The database used in matching in Chapter 4 is the same used in this section also.

First of all, the watermark signal must be ready to be embedded in the fingerprint image. As mentioned, the minutiae table will be used as the watermark signal. So, the minutiae table is converted to a bit stream. By seeking the maximum value of minutiae number at any cell in the table to know how many binary digits are needed for representing it, it is found to be "12" in the database under study. So, 4 binary digits are enough to represent any number in the minutiae table. Now watermark embedding steps can start as follows:

1. Every field in the minutiae table is converted to a 4-bit binary representation.
2. In addition to the binary minutiae table, two reference bits, 0 and 1, are appended to the watermark data to be embedded in the image.
3. Permute the watermark bits using a secret key $k_1$.

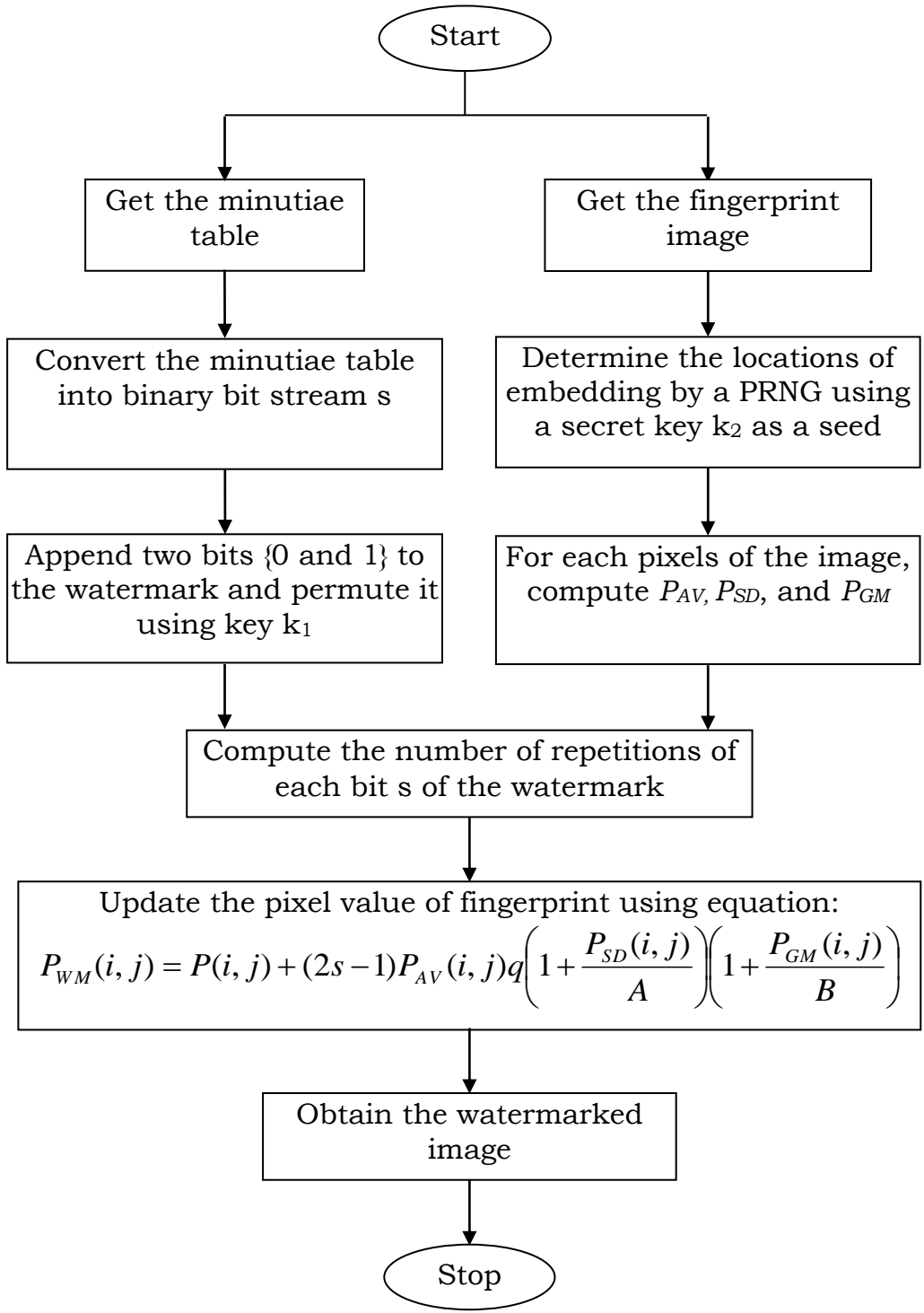

*Figure 5.5. Proposed watermark embedding algorithm*

4. A random number generator initialized with secret key $k_2$ generates locations of the fingerprint image pixels to be watermarked.
5. These pixels are changed according to Equation (5.1).
6. Every watermark bit with value s in Equation (5.1) is embedded at multiple locations in the fingerprint image.

### 5.6.2.2.  Watermark extraction algorithm

Figure 5.6 shows a flowchart of the watermark extraction algorithm. In the following, the steps of the algorithm are given.

1. Extraction starts with finding the data embedding locations in the watermarked image, via the secret key $k_2$ used during the watermark embedding stage.
2. For every bit embedding location, (i, j), its value is estimated as the linear combination of pixel values in a 5×5 cross-shaped neighborhood of the watermarked pixels as in Equation (5.2).
3. The difference between the estimated and watermarked pixel values is calculated by Equation (5.3).
4. These differences are averaged over all the embedding locations associated with the same bit, to yield $\bar{\delta}$ .
5. For finding an adaptive threshold, these averages are calculated separately for the reference bits, 0 and 1, as $\bar{\delta}_{R0}$ and $\bar{\delta}_{R1}$, respectively.
6. Finally, the watermark bit value $\hat{s}$ is estimated as Equation (5.4).
7. Inverse permute the resulted watermark bits to obtain the true watermark bits using the secret key $k_1$.

Both watermark embedding and extraction algorithms are implemented using MATLAB as shown in Appendix E.

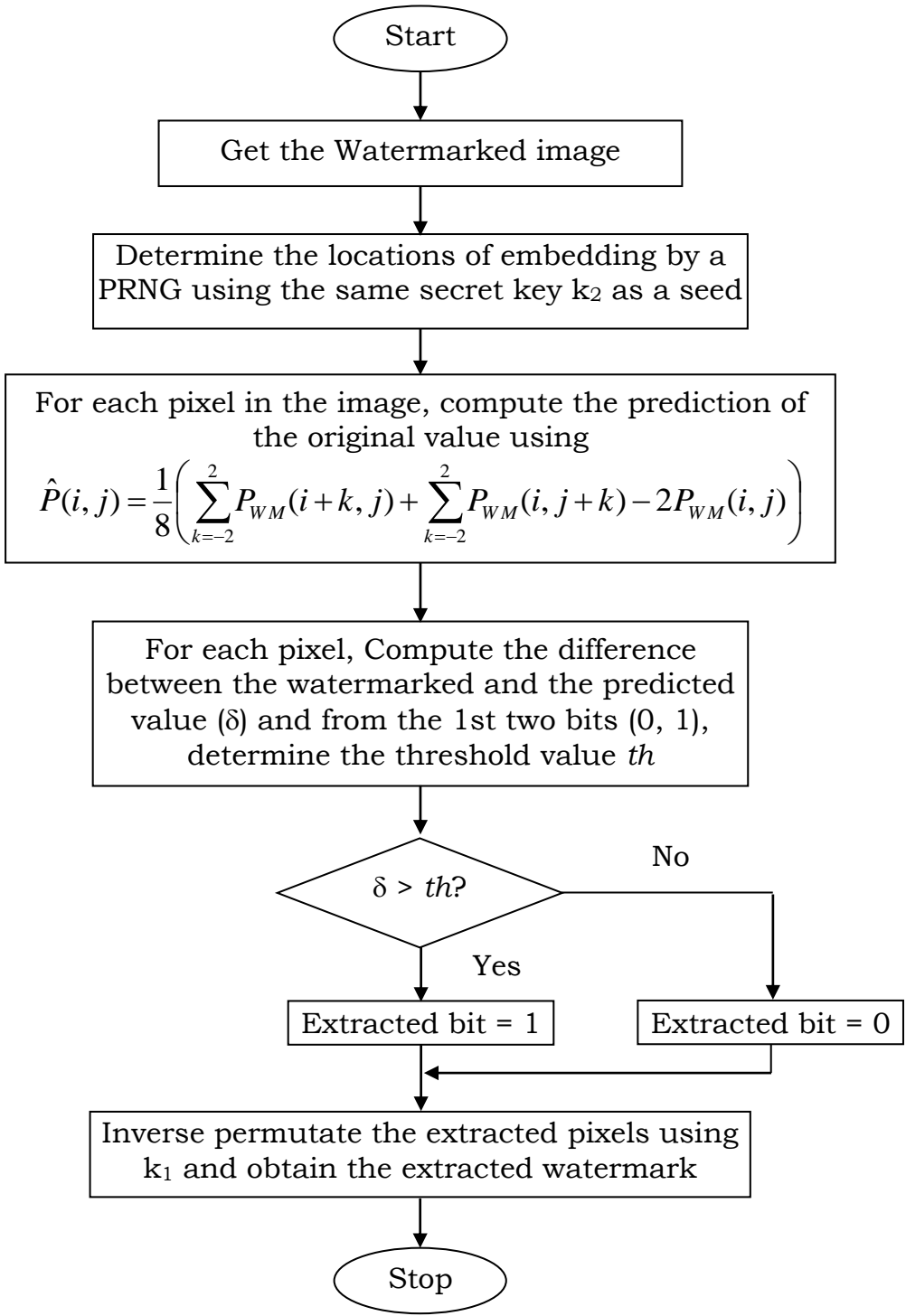

*Figure 5.6. Watermark extraction algorithm*

### 5.6.3.    Implementation of the proposed watermarking algorithm

Using the database used before for matching in Chapter 4, fingerprint 101_1, shown in Figure 5.7, will be watermarked by its minutiae table, and then the watermark will be extracted from the watermarked fingerprint again without the need of the original fingerprint.

#### 5.6.3.1.    Implementation of watermark embedding

As mentioned in Section 5.6.2, the first step is to convert the minutiae table into a bit stream. Because the max number is "12" in all minutiae tables in the database under study, 4 binary digits are adequate to represent these numbers. So, minutiae table of fingerprint 101_1, shown in Table 5.1, will be converted into its binary representation as shown in Table 5.2.

Append two bits 0, and 1 to the watermark signal to be used as a reference bits helping in calculating an adaptive threshold in determining the minutiae bit values during watermark extraction.

The secret key $K_1$ is used as a seed to initialize a Pseudo Random Number Generator (PRNG) to generate a random range of numbers (where its minimum value is one and maximum value is the length of watermark signal) to scramble (permute) the watermark bits. The hidden data size is approximately 23×8=23 bytes, whereas in the method in [62], it was 85 bytes (25 minutiae in average for each fingerprint times 9-bit for each of x, y, and θ fields per minutia = 25 × 9 × 3 = 675, so the size is 675/8 ≈ 85 bytes).

Now, the watermark signal is ready to be used. Then taking the fingerprint having this minutiae table to be watermarked, the second secret key $k_2$ is used as a seed to initialize a PRNG to generate locations of the fingerprint image to be watermarked. The exact value of the keys does not affect the performance of the method. This is to enhance the security issue of the watermarking method.

By using a density parameter ρ, the probability of any single pixel being used for embedding can be given. This value thus lies between 0 and 1, where 0 means that no information is embedded and 1 means that information is embedded in every pixel. If the number of pixels used for embedding is N, we have:

N= ρ × total number of pixels in the image [73].

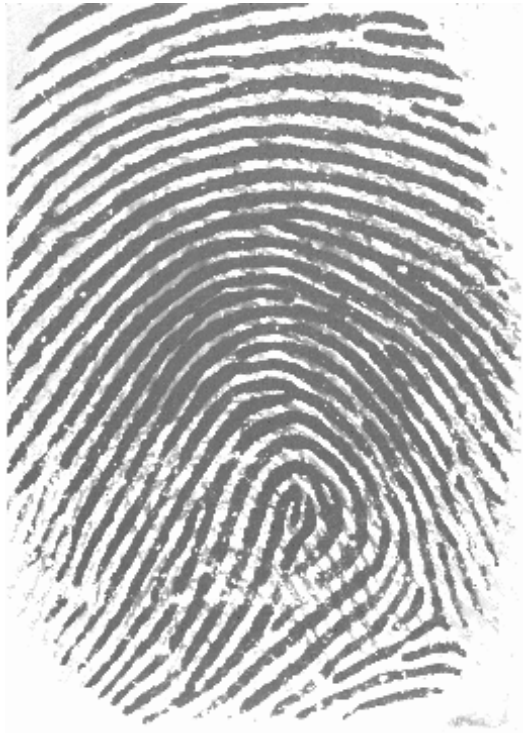

*Figure 5.7. Fingerprint image 101_1*

| Track no. | #Type1 | #Type2 |
|---|---|---|
| 1 | 1 | 2 |
| 2 | 0 | 0 |
| 3 | 0 | 1 |
| 4 | 1 | 0 |
| 5 | 1 | 1 |

| 6 | 2 | 0 |
| 7 | 1 | 1 |
| 8 | 2 | 1 |
| 9 | 0 | 0 |
| 10 | 5 | 3 |
| 11 | 2 | 0 |
| 12 | 1 | 2 |
| 13 | 0 | 1 |
| 14 | 0 | 0 |
| 15 | 1 | 0 |
| 16 | 0 | 1 |
| 17 | 0 | 0 |
| 18 | 0 | 0 |
| 19 | 1 | 1 |
| 20 | 1 | 0 |
| 21 | 1 | 0 |
| 22 | 0 | 1 |
| 23 | 1 | 1 |

*Table 5.1. Minutiae table of fingerprint 101_1*

| #Type1 | | | | #Type2 | | | |
|---|---|---|---|---|---|---|---|
| 0 | 0 | 0 | 1 | 0 | 0 | 1 | 0 |
| 0 | 0 | 0 | 0 | 0 | 0 | 0 | 0 |
| 0 | 0 | 0 | 0 | 0 | 0 | 0 | 1 |
| 0 | 0 | 0 | 1 | 0 | 0 | 0 | 0 |
| 0 | 0 | 0 | 1 | 0 | 0 | 0 | 1 |
| 0 | 0 | 1 | 0 | 0 | 0 | 0 | 0 |
| 0 | 0 | 0 | 1 | 0 | 0 | 0 | 1 |
| 0 | 0 | 1 | 0 | 0 | 0 | 0 | 1 |
| 0 | 0 | 0 | 0 | 0 | 0 | 0 | 0 |
| 0 | 1 | 0 | 1 | 0 | 0 | 1 | 1 |
| 0 | 0 | 1 | 0 | 0 | 0 | 0 | 0 |
| 0 | 0 | 0 | 1 | 0 | 0 | 1 | 0 |
| 0 | 0 | 0 | 0 | 0 | 0 | 0 | 1 |
| 0 | 0 | 0 | 0 | 0 | 0 | 0 | 0 |
| 0 | 0 | 0 | 1 | 0 | 0 | 0 | 0 |
| 0 | 0 | 0 | 0 | 0 | 0 | 0 | 1 |
| 0 | 0 | 0 | 0 | 0 | 0 | 0 | 0 |

| 0 | 0 | 0 | 0 | 0 | 0 | 0 | 0 |
|---|---|---|---|---|---|---|---|
| 0 | 0 | 0 | 1 | 0 | 0 | 0 | 1 |
| 0 | 0 | 0 | 1 | 0 | 0 | 0 | 0 |
| 0 | 0 | 0 | 1 | 0 | 0 | 0 | 0 |
| 0 | 0 | 0 | 0 | 0 | 0 | 0 | 1 |
| 0 | 0 | 0 | 1 | 0 | 0 | 0 | 1 |

*Table 5.2. Binary representation of minutiae table of fingerprint 101_1.*

The locations for embedding are determined as follows: for each pixel of the image, a pseudo-random number $x$ is generated. If $x$ is smaller than $\rho$, then information is embedded into the pixel. Otherwise the pixel is left intact. The value of $\rho$ used is 0.18 in experiments [62], as this value is limited by visibility of the changes in pixel values.

These pixels are changed using Equation (5.1), where the values of q, A, and B are as follows [62]: q=0.1, A=100, and B=1000. A higher q value increases the visibility of the hidden data. Increasing A or B decreases the effect of standard deviation and gradient magnitude in modulating watermark embedding strength, respectively.

Every watermark bit will be embedded at multiple locations in the fingerprint image. This repetition of each bit in the watermark signal increases the correct extraction rate of the embedded information. The repetition number of each watermark bit depends on the image capacity (size) and the size of the watermark and is calculated as follows:

No. of repetitions = N / watermark length ≈ 80.

A post-processing step includes eliminating the dominance of white background, and replacing this background with a uniform gray level distribution between [225, 235], with a mean of 230. This background transformation helps in hiding the minutiae data more invisibly.

The result of the watermark embedding algorithm is shown in Figure 5.8, where as shown, the watermark invisibility requirement is achieved. The time taken to perform the watermark embedding algorithm was 12.5 sec.

Using the following similarity equation to compare between two vectors $X$ and $X^*$ [75]*:*

$$Sim(X, X^*) = \frac{X \cdot X^*}{\sqrt{X \cdot X} \cdot \sqrt{X^* \cdot X^*}} \tag{5.5}$$

where $X$ and $X^*$ can represent the original and the watermarked fingerprint images respectively, and the operator (.) means the dot product. It is used as when both $X$ and $X^*$ are typical the value estimated will reach 1, then the result value is multiplied by 100, which gives a good indication of how both $X$ and $X^*$ are similar to each other.

It is found that the similarity between the original and the watermarked image is equal to 99.81%.

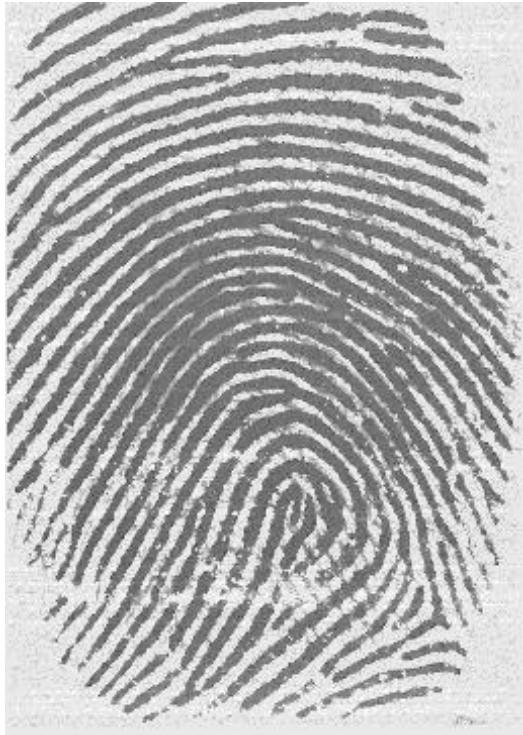

*Figure 5.8. Fingerprint 101_1 after watermarking*

### 5.6.3.2.    Implementation of watermark extraction

The same secret key $k_2$ used during watermark embedding is now used to find the data embedding locations in the watermarked image.

For every bit embedding location, $(i, j)$, its value is predicted as the linear combination of pixel values in a $5 \times 5$ neighborhood of the watermarked pixels as in Equation (5.2).

The difference between the estimated and watermarked pixel values is calculated by Equation (5.3). These differences are averaged over all embedding locations associated with the same bit, to yield $\bar{\delta}$.

Using the secret key $k_1$ used in watermark embedding, to generate the same sequence of permutation numbers to know the locations of previously appended two bits 0 and 1. Now, separately calculate the averages $\bar{\delta}_{R0}$ and $\bar{\delta}_{R1}$ for the reference bits, 0 and 1 respectively to find an adaptive threshold

$$th = \frac{\bar{\delta}_{R0} + \bar{\delta}_{R1}}{2} = 0.0035 \text{ in our experiment}$$

Finally, the watermark bit value $\hat{s}$ is estimated as Equation (5.4). Then, inverse permute the resulted watermark bits to obtain the true watermark bits using the secret key $k_1$.

The time taken to perform extraction was 0.9 sec. The similarity between the extracted watermark and the original watermark is 100%, i.e. the extracted minutiae table is identical to the embedded data.

To see how well the adaptive threshold can separate zero bits from one bits, Figure 5.9 shows the extracted watermark signal, where bits having $\bar{\delta}$ above the adaptive threshold are considered one bits, and bits having $\bar{\delta}$ below the adaptive threshold are considered zero bits.

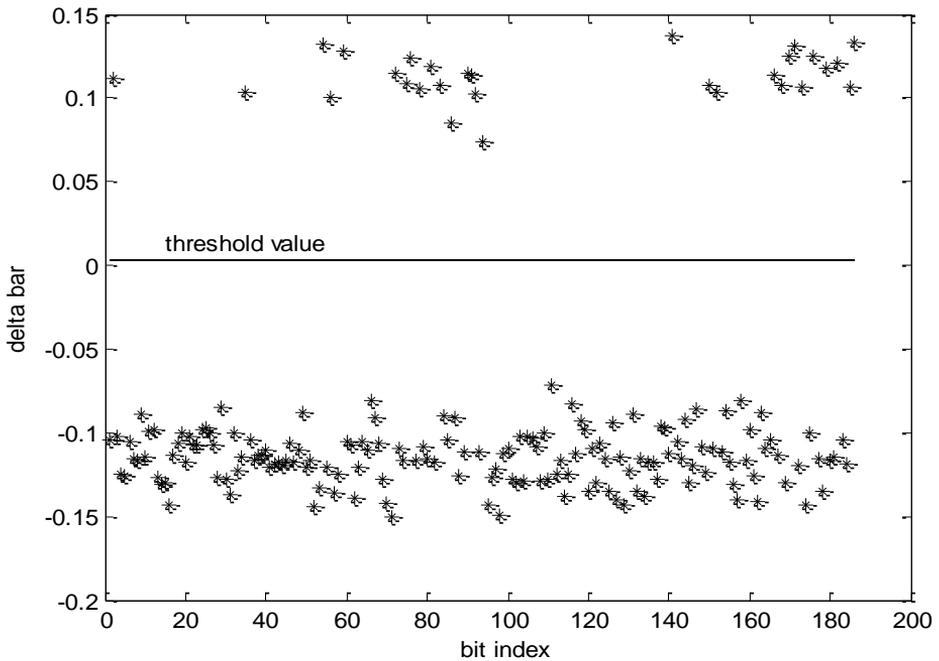

*Figure 5.9. Extracted Watermark representing the minutiae table of fgp 101_1.*

Also, an estimate of the original fingerprint image is also found by replacing the watermarked pixel values with the $\hat{P}(i, j)$ estimate calculated by Equation (5.2). The similarity between the reconstructed fingerprint image and the original fingerprint image was= 98.96%. Figure 5.10 shows the reconstructed fingerprint image from its watermarked version.

Watermark embedding is implemented on the entire database and Table 5.3 shows a sample of the results of these experiments, where EmbT stands for the embedding time, ExtT stands for the extraction time, OW stands for original watermark, EW stands for Extracted watermark, WI stands for watermarked image, OI stands for original image, and RCI stands for reconstructed image.

| Fgp | No.of Tracks | EmbT sec | ExtT sec | Similarity WI and OI | Similarity EW and OW | Similarity RCI and OI |
|---|---|---|---|---|---|---|
| 101_1 | 23 | 12.12 | 1.77 | 99.81 | 100 | 98.96 |
| 101_2 | 27 | 12.98 | 1.3 | 99.83 | 100 | 99.09 |
| 101_3 | 30 | 12.77 | 1.3 | 99.82 | 100 | 99 |
| 102_1 | 19 | 15.4 | 1.48 | 99.85 | 100 | 99.52 |

*Table 5.3. The results of applying watermarking algorithm on fingerprints 101_1, 101_2, 101_3, and 102_1.*

### 5.6.4.    **Applications of fingerprint watermarking**

Two applications are considered in this section, the basic watermarking method is the same in both of applications, but it differs in the characteristics of the host image carrying the minutiae table and the medium of data transfer. Each application will be described in the following two subsections.

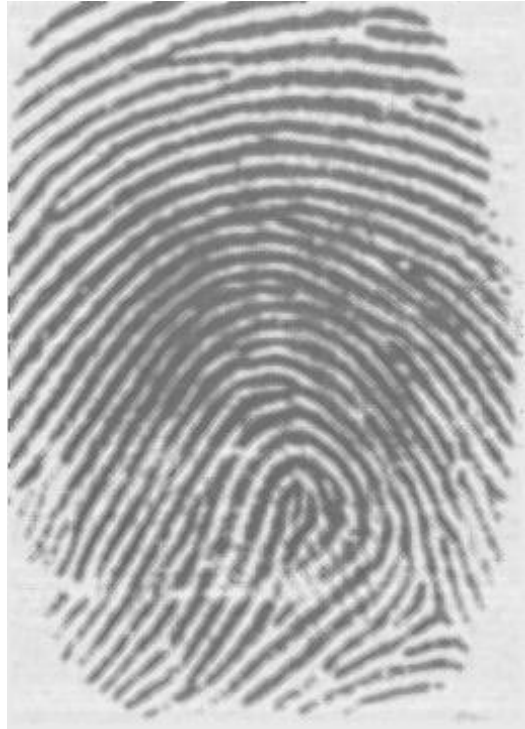

*Figure 5.10. Reconstructed fingerprint image from its watermarked version using Equation (5.2).*

### 5.6.4.1.    Application #1: Steganography based minutiae hiding

The first application scenario involves a steganography-based application (see Figure 5.11): the data (fingerprint minutiae table) that need to be transmitted (possibly via a non-secure communication channel) is hidden in a host (i.e., cover) image, whose only function is to carry the data. The host image is the fingerprint having this minutiae table. Using such a fingerprint image to carry fingerprint minutiae table provides an increased level of security since a person who intercepts the communication channel and obtains the carrier image (fingerprint image) would treat this image as if it does not contain any hidden data.

This application can be used to counter the attack on the communication channel between the database and the fingerprint matcher. An attacker will most probably not suspect that a cover image is carrying the minutiae information.

Two secret keys are utilized in watermark embedding to increase the security of the hidden data (see Figure 5.11). The image with embedded data (watermarked image) is sent through the channel that may be subject to interceptions. At the extraction site, using the same keys that were used by the embedding method (which can be delivered to the decoder using a secure channel prior to watermarked image transfer), the hidden data is recovered from the watermarked image.

So, this application can be used to guarantee secure transmission of acquired fingerprint images from intelligence agencies to a central image database.

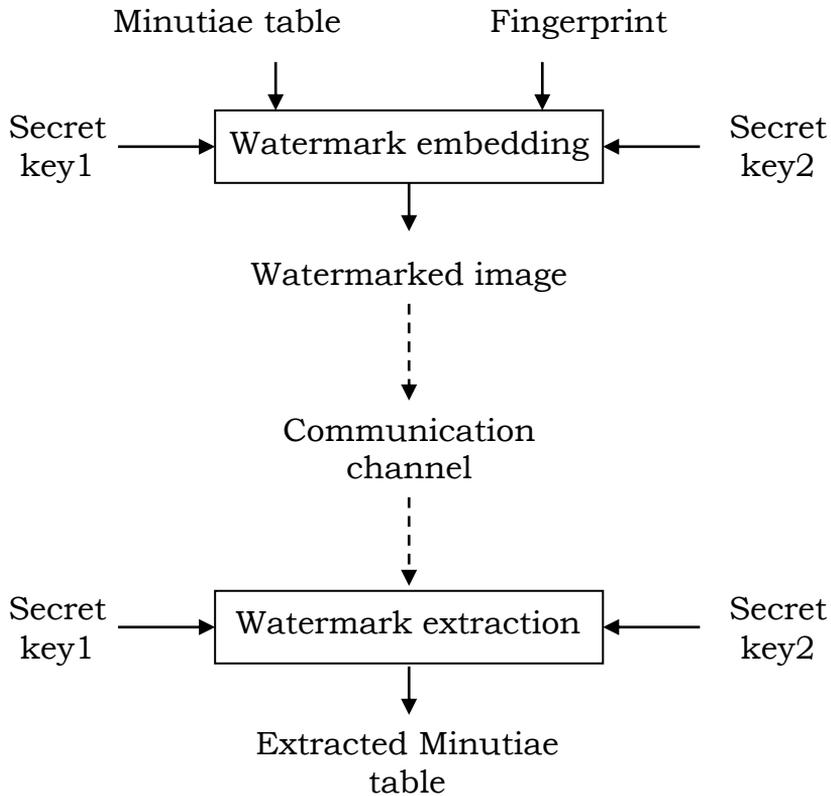

*Figure 5.11. Steganography-based minutiae hiding.*

Also, it can be used to eliminate several types of biometric system attacks described in [76] such as:

- A user who succeeds in inserting a new template into the database will not be authenticated at the access site since this new template will not contain the minutiae data.

- Fraud detection in fingerprint images by means of fragile watermarks (which do not resist to any operations on the data and get lost, thus indicating possible tampering of the data) [70].

5.6.4.2.     Application #2: Minutiae hiding for fingerprint
    verification

The second application scenario aims at increasing the
security of fingerprint images. In this scenario, a person's
fingerprint image, which also carries that person's
fingerprint minutiae data, is encoded in a smart card (see
Figure 5.12). At a controlled access site, this image will be
read from the smart card and the original fingerprint image
will be reconstructed. The extracted minutiae table will be
compared to the minutiae obtained from the user at the
access site. These two minutiae data sets and the
reconstructed fingerprint image will be used to accept or
reject the user.

This application can be used to eliminate other types of
biometric system attacks such as:

- Fake biometric submission via a smart card that
  contains a non-authentic image will be useless since
  that image will not contain the true minutiae data.

- Resubmission of digitally stored biometrics data (e.g.,
  via a stolen but authentic smart card) will not be
  feasible since the system authenticates every user by
  using this data along with the minutiae data obtained
  online at a controlled access site.

## 5.6.5.     Watermark robustness analysis

Watermarked fingerprints are susceptible to other types of
unintentional attacks such as Gaussian noise addition,
JPEG compression,  and different types of filtering such as
median filtering, trimmed mean filtering, and Wiener
filtering. A brief discussion of some of these attacks is given
in the following subsections. To apply these attacks and
many other types, a MATLAB source code is available at
[77].

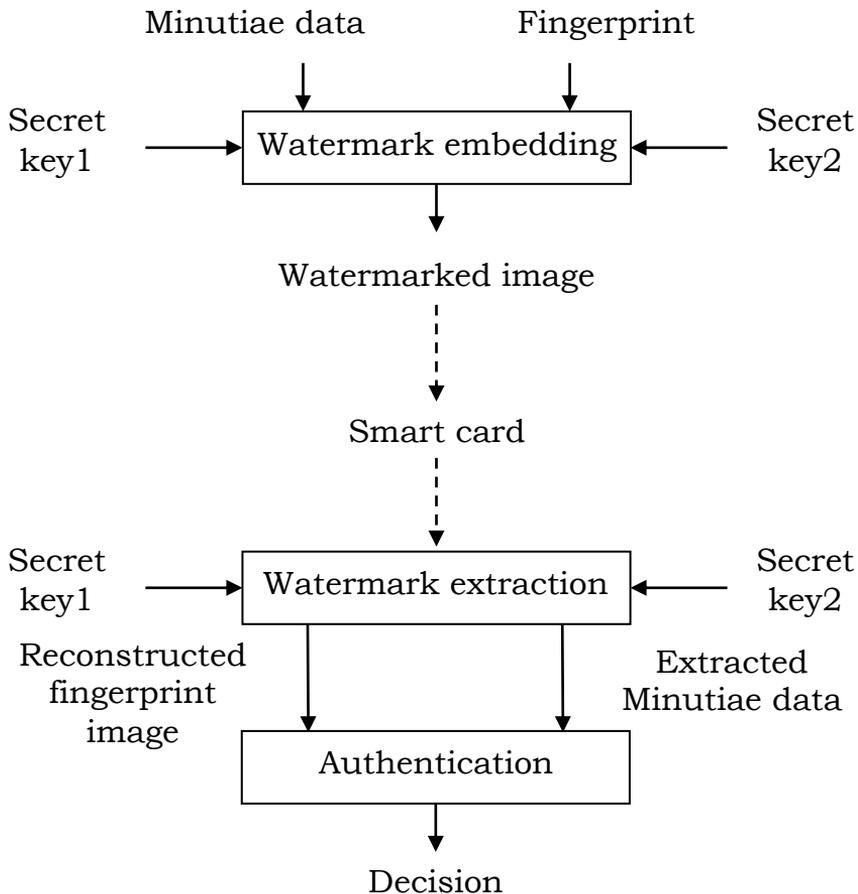

*Figure 5.12. Minutiae hiding for fingerprint verification*

**JPEG Compression**

JPEG (pronounced "jay-peg") is a standardized image compression mechanism. JPEG stands for Joint Photographic Experts Group, the original name of the committee that wrote the standard [78].

JPEG is designed for compressing full-color or gray-scale images of natural, real-world scenes. It works well on photographs, naturalistic artwork, and similar material, and not so well on lettering, simple cartoons, or line drawings. JPEG handles only still images, but there is a related standard called MPEG for motion pictures.

JPEG is "lossy", meaning that the decompressed image isn't quite the same as the original one. There are lossless image compression algorithms, but JPEG achieves much greater compression than is possible with lossless methods. JPEG is designed to exploit known limitations of the human eye, notably the fact that small color changes are perceived less accurately than small changes in brightness.

**Median filtering**

Smoothing filters are used for blurring and for noise reduction. Blurring is used in preprocessing steps, such as removal of small details from an image prior to (large) object extraction, and bridging of small gaps in lines or curves. Noise reduction can be accomplished by blurring with a linear filter and also by nonlinear filtering [5].

When the objective is to achieve noise reduction rather than blurring, an approach called *median filters* is used. That is, the gray level of each pixel is replaced by the median of the gray levels in a neighborhood of that pixel. This method is particularly effective when the noise pattern consists of strong, spikelike components and the characteristic to be preserved is edge sharpness. The median $m$ of a set of values is such that half the values in the set are less than $m$ and half are greater than $m$. In order to perform median filtering in a neighborhood of a pixel, the values of the pixel and its neighbors are first sort, the median is determined, and this value is assigned to the pixel.

**Trimmed Mean Filtering**

It is both an order and a mean filter because it relies on ordering the pixel values, but then is calculated by an averaging process. The trimmed mean filter is the average of the pixel values within the window, but with some of the endpoint-ranked values excluded. This filter is useful for images containing multiple types of noise, such as gaussian and salt-and-pepper noise. So, the trimmed mean filter uses order statistics such as the median filter but produces the trimmed version of the mean centered about the median point [79].

**Wiener Filtering**

The Wiener filter is the MSE-optimal stationary linear filter for images degraded by additive noise and blurring. Calculation of the Wiener filter requires the assumption that the signal and noise processes are second-order stationary. Wiener filters are usually applied in the frequency domain [5].
To study the effect of these image processing techniques, the watermarked images are to be processed with those techniques, and then the extracted watermarks are to be tested if they are robust (still valuable) or not against those attacks.

Table 5.4 shows the similarity values between the original watermarks (minutiae tables) and the extracted watermarks from the watermarked fingerprint images 101_1, 101_2, and 101_3 after applying the previous attacks. As shown from Table 5.4, the watermark is robust against Gaussian noise addition, JPEG compression, and Wiener filtering, whereas it is not robust against median and trimmed mean filtering.

| Applied attack | Fgp101_1 | Fgp101_2 | Fgp101_3 |
|---|---|---|---|
| No Attack | 100.00 | 100.00 | 100.00 |
| Gaussian noise | 100.00 | 100.00 | 100.00 |
| Jpegcompression_100 | 100.00 | 100.00 | 100.00 |
| Jpegcompression_80 | 100.00 | 100.00 | 100.00 |
| Jpegcompression_60 | 100.00 | 100.00 | 100.00 |
| Jpegcompression_40 | 100.00 | 100.00 | 100.00 |
| Jpegcompression_25 | 95.20 | 100.00 | 96.08 |
| Median filter | 56.45 | 79.24 | 48.18 |
| Trimmed mean filter | 61.71 | 82.75 | 47.62 |
| Wiener filter | 100.00 | 97.80 | 100.00 |

*Table 5.4. Similarity values between original and extracted watermarks for different fingerprint images.*

Figure 5.14 till Figure 5.23 show the attacked versions of watermarked fingerprint image 101_1 together with their extracted watermarks from each of them.

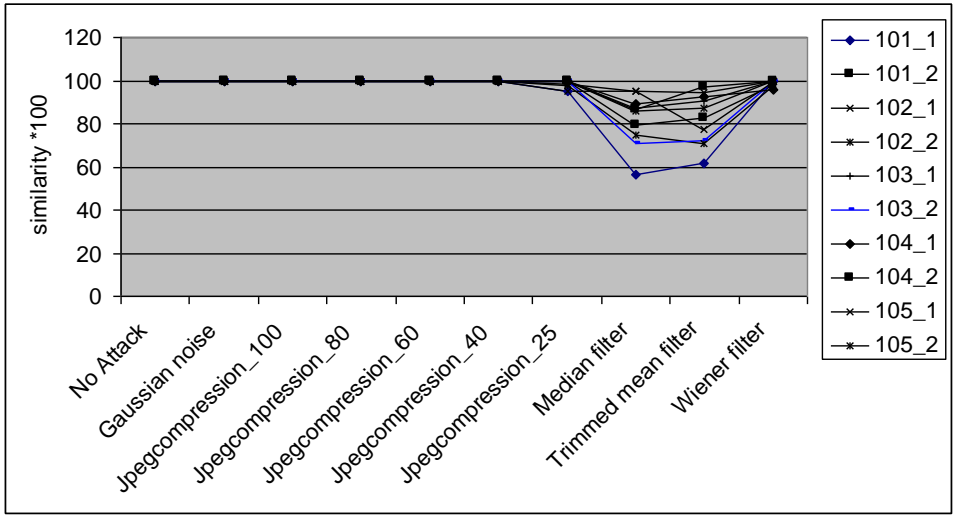

*Figure 5.13. Similarity values between original and extracted watermarks after applying different attacks for different images.*

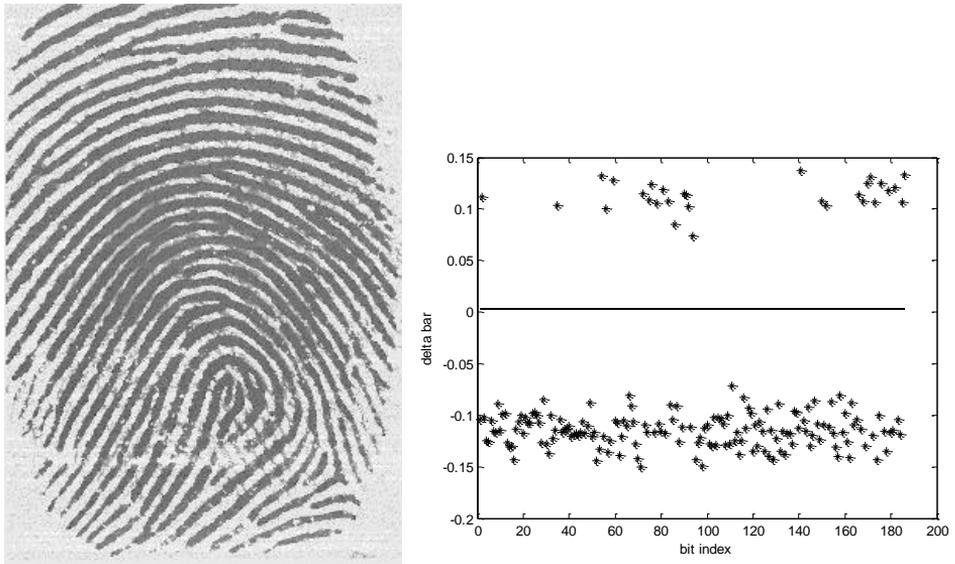

*Figure 5.14. a) The watermarked fingerprint without any attack, b) The extracted watermark signal.*

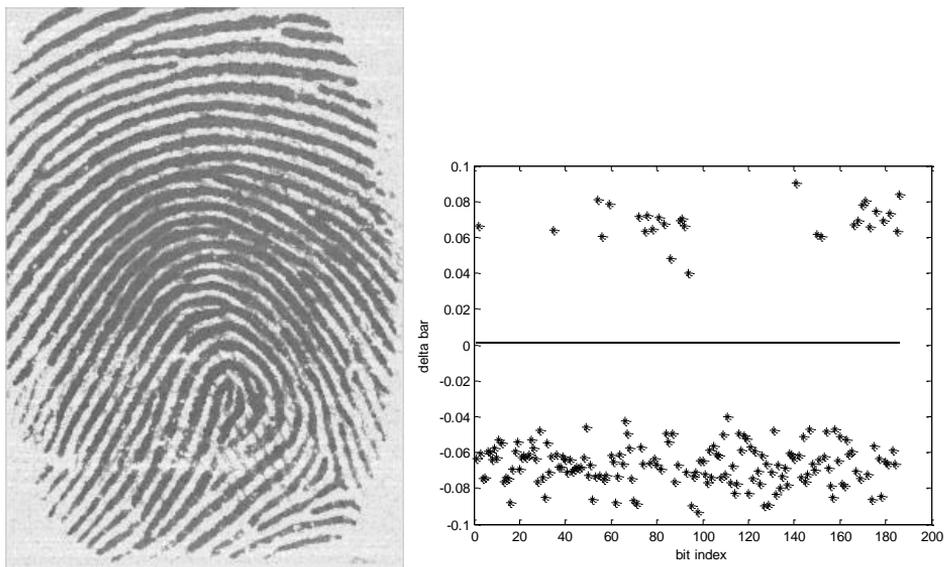

*Figure 5.15. a) The watermarked fingerprint after gaussian noise addition, b) The extracted watermark signal.*

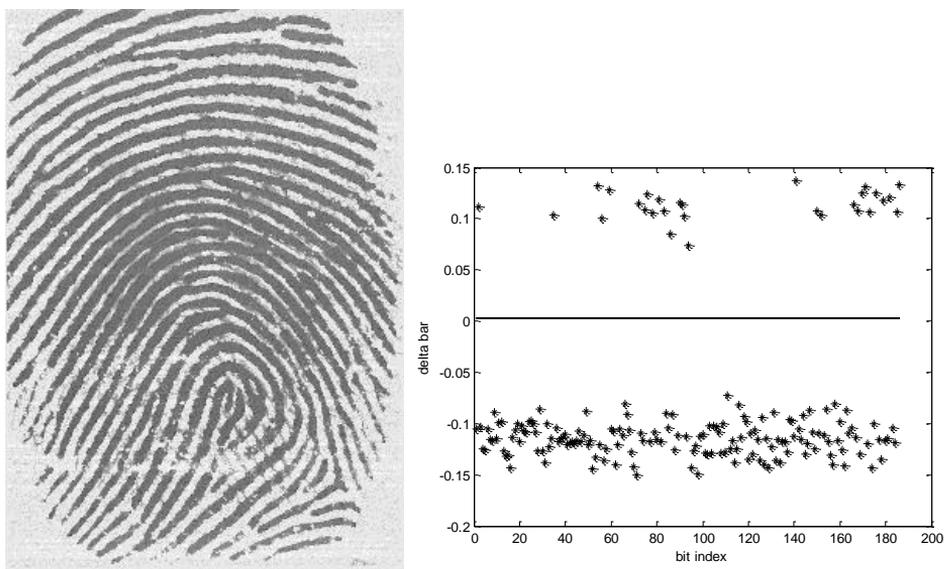

*Figure 5.16. a) The watermarked fingerprint after JPEG compression with Q=100, b) The extracted watermark signal.*

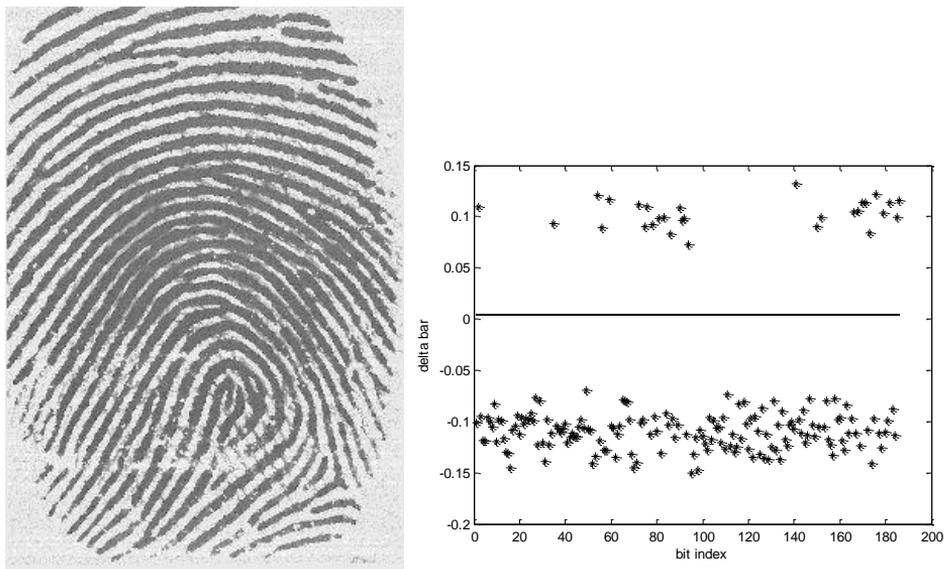

*Figure 5.17. a) The watermarked fingerprint after JPEG compression with Q=80, b) The extracted watermark signal.*

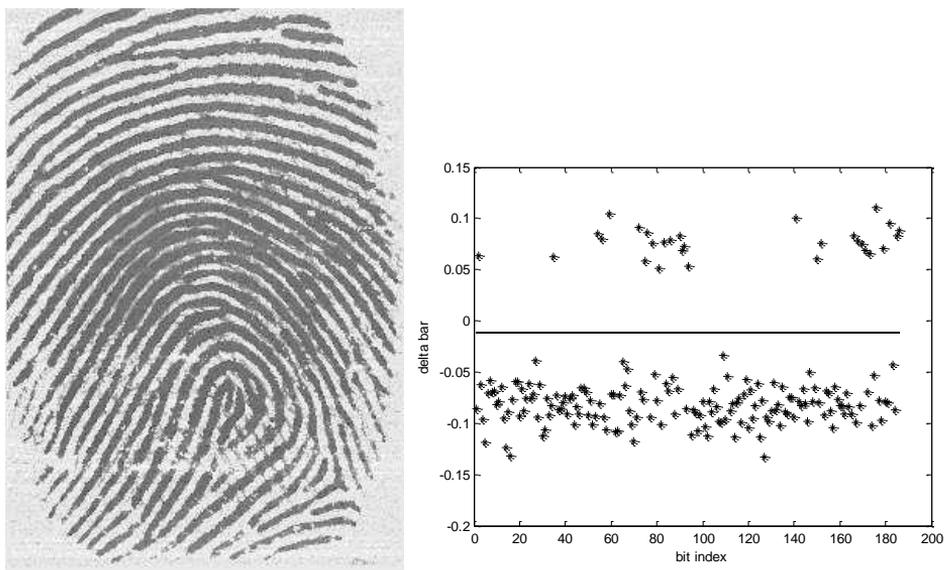

*Figure 5.18. a) The watermarked fingerprint after JPEG compression with Q=60, b) The extracted watermark signal.*

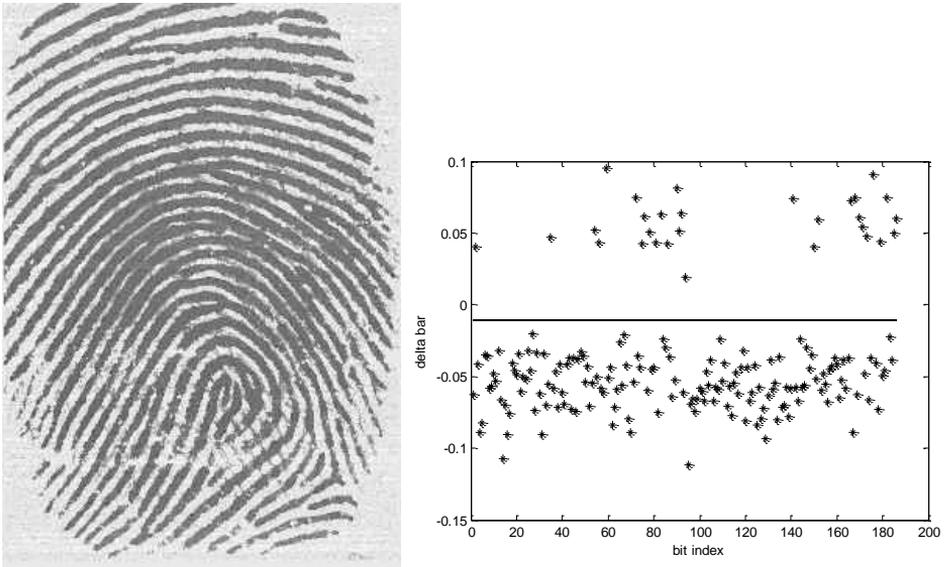

*Figure 5.19. a) The watermarked fingerprint after JPEG compression with Q=40, b) The extracted watermark signal.*

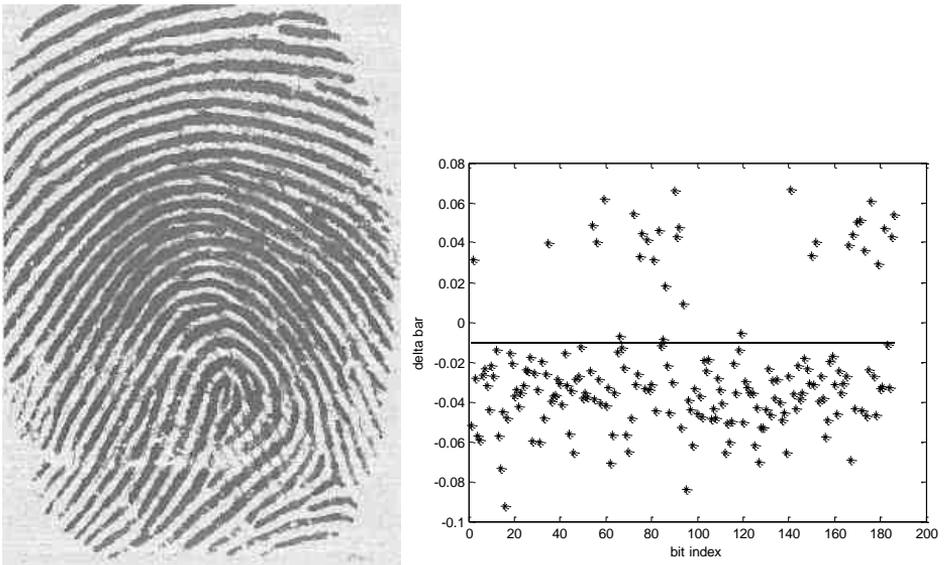

*Figure 5.20. a) The watermarked fingerprint after JPEG compression with Q=25, b) The extracted watermark signal.*

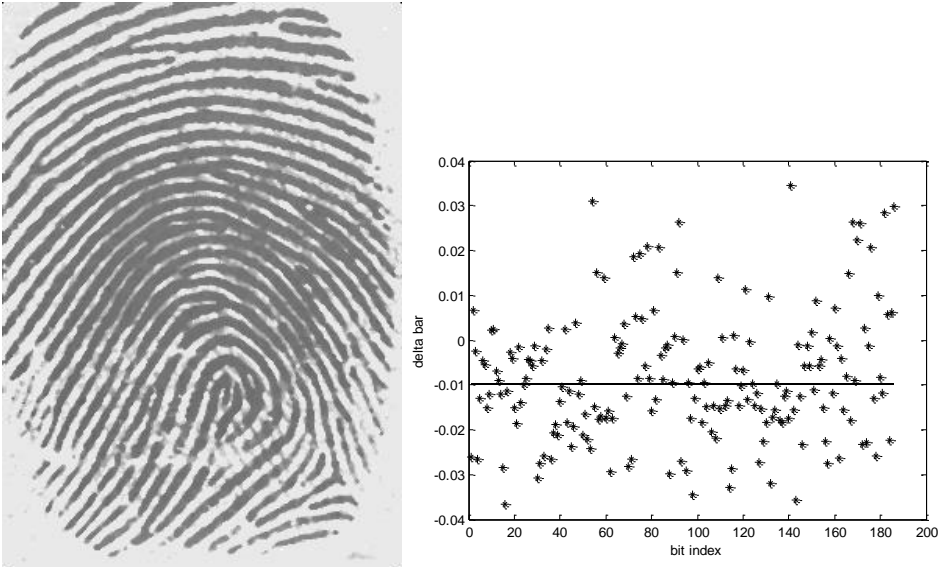

*Figure 5.21. a) The watermarked fingerprint after median filtering, b) The extracted watermark signal.*

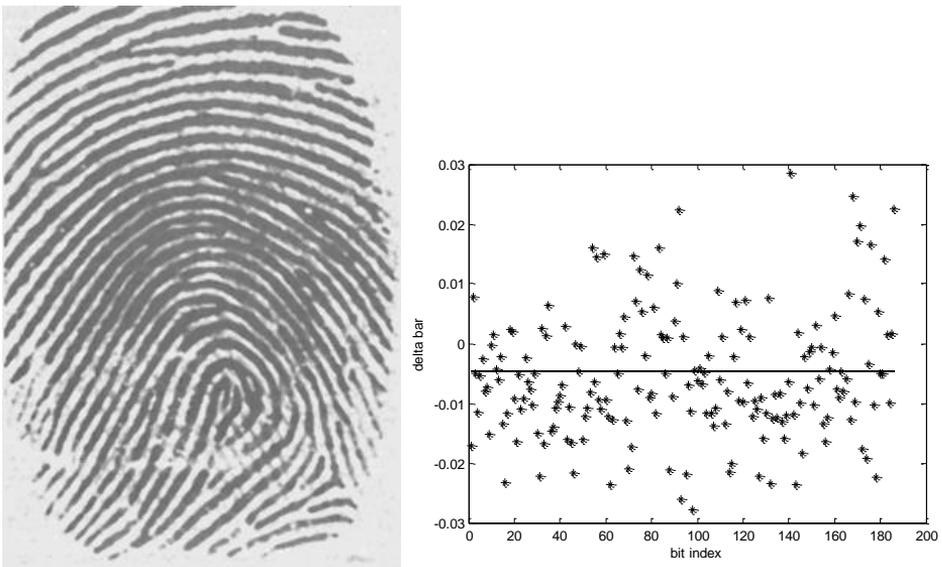

*Figure 5.22. a) The watermarked fingerprint after trimmed mean filtering, b) The extracted watermark signal.*

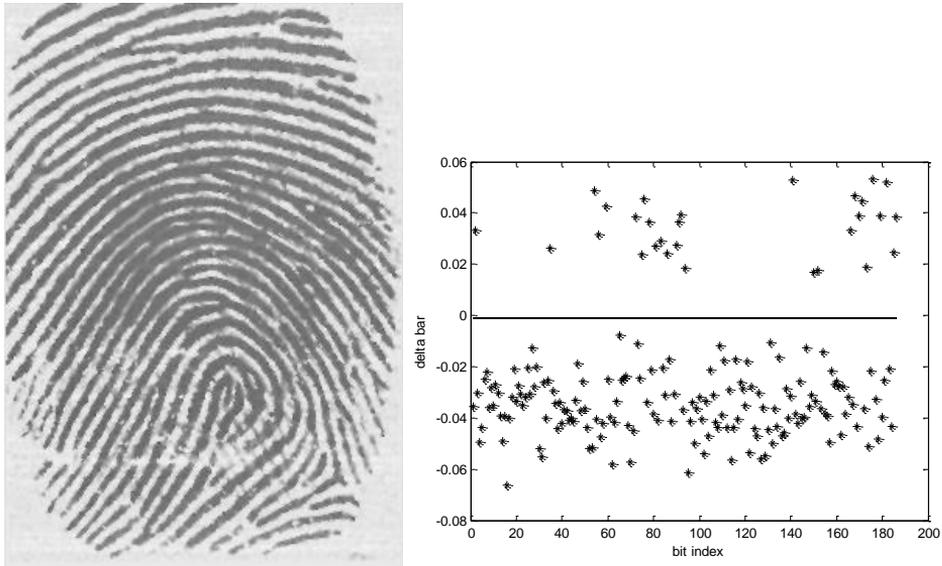

*Figure 5.23. a) The watermarked fingerprint after Wiener filtering, b) The extracted watermark signal.*

As can be shown from Figure 5.13, the watermark is robust against Gaussian noise addition, JPEG compression with quality factors Q =100, 80, 60, 40, and Wiener filtering.

## 5.6.6.    Recognition accuracy analysis

To study the effect of watermarking on recognition accuracy, there are two cases:

- Without applying any attacks on the watermarked image:

After watermarking all the fingerprints in the database (20 fingerprints each having 3 impressions for a total of 60 images), the watermarks (minutiae tables) extracted from all these watermarked images are all the same as the embedded watermarks. So, the recognition accuracy will be the same without any changes which was 98% from results drawn in Chapter 4.

- After applying different attacks on the watermarked image:

After having 60 different watermarked images, applying the nine different attacks described above, together with their unattacked version, we will have 600 different images. It is hoped that all attacked versions of a certain watermarked fingerprint will give the same matching score as if they do not have been attacked.

Extracting the watermark (minutiae table) from the above 600 different watermarked images, and applying the verification algorithm between each extracted watermark and its corresponding minutiae tables stored in the database, two numbers will be obtained. These numbers are compared with the two estimated thresholds before in Chapter 4 which were $t_1=17$ and $t_2=8$. Table 5.5 shows the results of recognition accuracy.

|  | All attacks on DB | remove median and trimmed mean filters | remove all filters | remove all filters + JPEG 25 | remove all filters + JPEG 25,40 | remove all filters+ JPEG 25,40,60 |
|---|---|---|---|---|---|---|
| FRR | 24.43 | 12.62 | 7.29 | 4.54 | 3.75 | 3.63 |
| 100-FRR | 75.57 | 87.38 | 92.71 | 95.5 | 96.25 | 96.37 |

*Table 5.5. Recognition accuracy after applying different attacks on watermarked images.*

As shown from Table 5.5, the recognition accuracy felt down to 75.57% when all attacks are considered on all watermarked fingerprints of database.

By removing the two filters attacks: median and trimmed mean filters applied, the recognition accuracy rose again to reach 87.38%.

If all the three types of filters are removed, the recognition accuracy reaches 92.71%.

Finally, trying to remove the JPEG compression attack having quality factors Q=25 and 40 together with other three filters, the reached recognition accuracy was ≈ 96%.

Figure 5.24 shows FRR and (100-FRR) curves for fingerprint database after applying different attacks on their watermarked versions.

So, we can conclude that to reach recognition accuracy near the one without any applied attacks (which was 98%), the watermarking algorithm is robust only against Gaussian noise addition and JPEG compression with high and moderate quality factors.

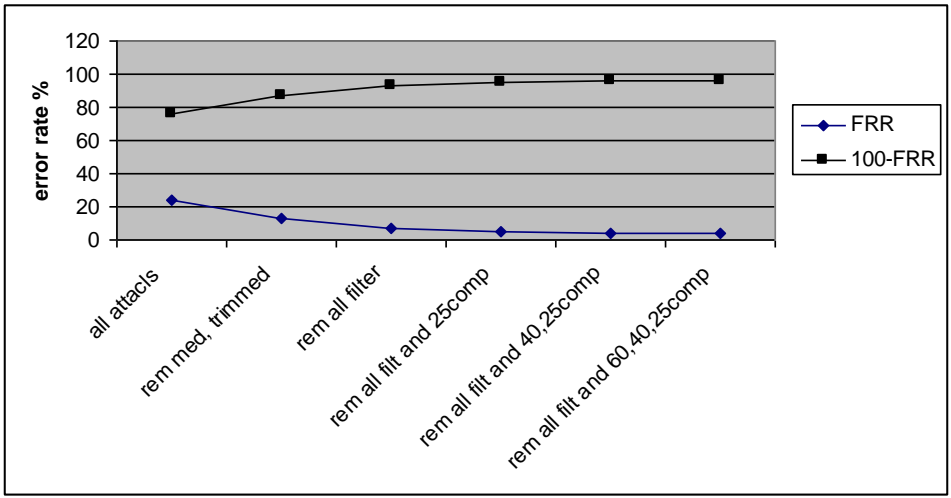

*Figure 5.24. FRR and (100-FRR) for fingerprint database after applying attacks on their watermarked versions.*

# Chapter 6
# Conclusions and Future Work

## 6.1.    Conclusions

During the development of this thesis, several issues were identified that will yield contributions to the field of AFIS. Among these issues, the following conclusions can be enumerated:

1. All steps of Fingerprint Identification System FIS and all its components are studied.
2. Thinning is a critical pre-processing step in fingerprint identification system.
3. A new thinning algorithm for fingerprints is proposed which is proved to be fast and more accurate.
4. Different matching algorithms are studied and a new minutiae-based matching algorithm is proposed.
5. The proposed matching algorithm is proved to be:
   - rotation and translation invariant,
   - more fast, and
   - less need for storage.
6. Recognition accuracy is 98% with enrollment time of 6 sec and matching time of 0.001 sec.
7. Many attacks can threat the FIS, so data must be secured all the way from the scanner to the application.
8. Watermarking is applied to secure FIS. It is proved to be invisible and robust against Gaussian noise addition, and JPEG compression with high and moderate quality factors.
9. After watermarking and without attacks, recognition accuracy does not change and still is 98%.
10. After applying above attacks, recognition accuracy decreases slightly to reach 96%.

## 6.2.    **Future work**

The following points are some suggestions for further study:

1. To search for a more accurate and fast thinning algorithm.
2. Trying to achieve higher values of recognition accuracy in proposed matching algorithm.
3. Searching for a more fast and effective enhancement method for fingerprints to lessen the overall enrollment time.
4. Trying to enhance the robustness requirement of watermarking against other classes of attacks.
5. Studying how to make the watermark embedding algorithm faster.
6. Implementing and realizing the proposed algorithms, thinning, matching, and watermarking, using FPGA port.

# Appendices

## Appendix A
## Core point detection Matlab source code

This appendix is the MATLAB source code of the core point detection in fingerprints.

```matlab
function [XofCenter,YofCenter] = get_core(fingerprint)
x=[-16:1:16];
y=[-16:1:16];
dimx=size(x,2);
dimy=size(y,2);
%gaussian variance, order of complex filter
variance=sqrt(55);
order=1;
filter_core=zeros(dimx,dimy);
filter_delta=zeros(dimx,dimy);
for ii=1:dimx
    for jj=1:dimy
        exponent=exp(-(x(ii)^2+y(jj)^2)/(2*variance^2));
        % filter core
        factor=x(ii)+i*y(jj);
        filter_core(ii,jj)=exponent*factor^order;
        % filter delta
        factor=x(ii)-i*y(jj);
        filter_delta(ii,jj)=exponent*factor^order;
    end
end
% Gaussian Low Pass Filter ----------
%------------------------------------
x=[-16:1:16];
y=[-16:1:16];
dimx=size(x,2);
dimy=size(y,2);
variance=sqrt(1.2);
filter=zeros(dimx,dimy);
for ii=1:dimx
    for jj=1:dimy
        exponent=exp(-(x(ii)^2+y(jj)^2)/(2*variance^2));
        filter(ii,jj)=exponent;
    end
end
%------------------------------------
img=fingerprint;
img=double(img);
```

```matlab
%------------------------------------------------------------
% complex field at 0 level
[gx,gy]=gradient(img);
num=(gx+i*gy).^2;
den=abs((gx+i*gy).^2);
pos=find(den);
num(pos)=num(pos)./den(pos);
z=zeros(size(img,1),size(img,2));
z(pos)=num(pos);
pos=find(den==0);
z(pos)=1;
%*********************************
% ------------------- parameters --------------------------
bxv=8;            % size of block of variance
byv=8;
threshold_var=20;  % threshold of variance
dimsclose=10;   % dimension of closing operation
dimseerode=44;   % dimension of erosion operation
[dimx,dimy]=size(fingerprint);
%------------------------------------------------------------
temp=z;
[temp,dimxt,dimyt]=mirror(temp);
z_f=conv2fft(temp,filter_core,'same');
z_f=recrop(z_f,dimxt,dimyt);
z_f=abs(z_f);
% resize--------------------
imgd=double(fingerprint);
dimxr=dimx-mod(dimx,bxv);
dimyr=dimy-mod(dimy,byv);
imgr=imgd(1:dimxr,1:dimyr);
%--------------------------
nbx=dimxr/bxv;
nby=dimyr/byv;
mat_var=zeros(dimxr,dimyr);
for ii=1:nbx
    for jj=1:nby
        block=imgr((ii-1)*bxv+1:ii*bxv,(jj-
1)*byv+1:jj*byv);
        media=sum(sum(block))/(bxv*byv);
        variance=1/(bxv*byv)*sum(sum(abs(media.^2-
block.^2)));
        mat_var((ii-1)*bxv+1:ii*bxv,(jj-
1)*byv+1:jj*byv)=sqrt(variance);
    end
end
mat_ok=zeros(dimxr,dimyr);
pos= mat_var>threshold_var;
mat_ok(pos)=1;
mat_ok(dimx,dimy)=0;
mat_ok=imclose(mat_ok,ones(dimsclose));
```

```
mat_ok=imerode(mat_ok,ones(dimseerode));
%----------------------------------------------------------
% now calculate core point of each block
matrix_final=z_f.*mat_ok;

[max_vector,position_vector]=max(matrix_final);
[~,position]=max(max_vector);
y_max=position;
x_max=position_vector(position);

XofCenter=y_max;
YofCenter=x_max;
```

## The following partial functions are used by the get_core function:

```
function [out,dimxt,dimyt]=mirror(in)
% mirroring
[dimxt,dimyt]=size(in);

in_memo=in;

extx=20;
exty=20;

out=zeros(dimxt+2*extx,dimyt+2*exty);
out(extx+1:extx+dimxt,exty+1:exty+dimyt)=in_memo;

for ii=1:extx
    out(extx-ii+1,:)=out(extx+ii,:);
    out(extx+dimxt+ii,:)=out(extx+dimxt-ii+1,:);
end

for ii=1:exty
    out(:,exty-ii+1)=out(:,exty+ii);
    out(:,exty+dimyt+ii)=out(:,exty+dimyt-ii+1);
end
```

```
function [out] = conv2fft(z1,z2,shape,shape2)
if ((nargin==3)&&(isa(shape,'char')))
    if strcmp(shape,'same')
        z1x=size(z1,1);
        z1y=size(z1,2);
        z2x=size(z2,1);
        z2y=size(z2,2);
```

```
        if any(any(imag(z1)))||any(any(imag(z2)))
            out=(ifft2(fft2(z1,z1x+z2x-1,z1y+z2y-
1).*fft2(z2,z1x+z2x-1,z1y+z2y-1)));
        else
            out=real(ifft2(fft2(z1,z1x+z2x-1,z1y+z2y-
1).*fft2(z2,z1x+z2x-1,z1y+z2y-1)));
        end

        px=((z2x-1)+mod((z2x-1),2))/2;
        py=((z2y-1)+mod((z2y-1),2))/2;

        out=out(px+1:px+z1x,py+1:py+z1y);
        return;
    end
…end

function [out]=recrop(in,dimxt,dimyt)
% recroppping

extx=20;
exty=20;

out=in(extx+1:extx+dimxt,exty+1:exty+dimyt);
```

# Appendix B
# Minutiae extraction and matching algorithm Matlab source codes

This appendix is the MATLAB source code of the minutiae extraction from fingerprints together with the proposed minutiae table construction which corresponds to that fingerprint. Also, the performance evaluation m-file is shown.

```
f={     '..\db\101_1.tif'
        '..\db\101_2.tif'
        '..\db\101_3.tif'

        '..\db\102_1.tif'
        '..\db\102_2.tif'
        '..\db\102_3.tif'

        '..\db\103_1.tif'
        '..\db\103_2.tif'
        '..\db\103_3.tif'

        '..\db\104_1.tif'
        '..\db\104_2.tif'
        '..\db\104_3.tif'

        '..\db\105_1.tif'
        '..\db\105_2.tif'
        '..\db\105_3.tif'

        '..\db\106_1.tif'
        '..\db\106_2.tif'
        '..\db\106_3.tif'

        '..\db\107_1.tif'
        '..\db\107_2.tif'
        '..\db\107_3.tif'

        '..\db\108_1.tif'
        '..\db\108_2.tif'
        '..\db\108_3.tif'

        '..\db\109_1.tif'
```

```
        '..\db\109_2.tif'
        '..\db\109_3.tif'

        '..\db\110_1.tif'
        '..\db\110_2.tif'
        '..\db\110_3.tif'

        '..\db\111_1.tif'
        '..\db\111_2.tif'
        '..\db\111_3.tif'

        '..\db\112_1.tif'
        '..\db\112_2.tif'
        '..\db\112_3.tif'

        '..\db\113_1.tif'
        '..\db\113_2.tif'
        '..\db\113_3.tif'

        '..\db\114_1.tif'
        '..\db\114_2.tif'
        '..\db\114_3.tif'

        '..\db\115_1.tif'
        '..\db\115_2.tif'
        '..\db\115_3.tif'

        '..\db\116_1.tif'
        '..\db\116_2.tif'
        '..\db\116_3.tif'

        '..\db\117_1.tif'
        '..\db\117_2.tif'
        '..\db\117_3.tif'

        '..\db\118_1.tif'
        '..\db\118_2.tif'
        '..\db\118_3.tif'

        '..\db\119_1.tif'
        '..\db\119_2.tif'
        '..\db\119_3.tif'

        '..\db\120_1.tif'
        '..\db\120_2.tif'
        '..\db\120_3.tif'
    };
M=[];
```

```matlab
my_tracks=zeros(length(f),1);
for fi=1:length(f)
    I=imread(f{fi});
    [~,~,~,~,enh_img]=fft_enhance_cubs(I);
    %figure,imshow(enh_img)

    [x_core,y_core]=get_core(enh_img);
    I=enh_img;
    I=adaptiveThres(double(I),32,0);%figure,imshow(I)
    I=im2double(bwmorph(I,'thin',Inf));%figure,imshow(I)
    I=im2double(bwmorph(I,'clean'));%figure,imshow(I)
    I=im2double(bwmorph(I,'hbreak'));%figure,imshow(I)
    I1=im2double(bwmorph(I,'spur'));
    I2=MinutiaPointExtraction(I1);
    I3=RemoveFalseTermination(I1,I2);
    I3=MinutiaPointExtraction(I3);
    I3=RemoveFalseBifurcation(I3);
    tic;Index=or((I3==1),(I3==3));
    [Y,X]=find(Index==1);
    Type = zeros(length(X),1);
    dist_from_core = zeros(length(X),1);
    FpPlot(I3,X,Y,x_core,y_core);
    for i=1:length(X)
        dx=X(i)-x_core;
        dy=Y(i)-y_core;
        dist_from_core(i)=sqrt((dx^2)+(dy^2));
        Type(i)=I3(Y(i),X(i));
    end
    max_dist=max(dist_from_core);
    track=0;step=10;
    no_tracks=ceil(max_dist/step);
    my_tracks(fi)=no_tracks;
    table=zeros(no_tracks,2);
    for i=1:step:step*no_tracks
        track=track+1;
        for j=1:length(X)
            if (dist_from_core(j)>=i) &&
(dist_from_core(j)<i+step)
                if Type(j)==1
                    table(track,1)=table(track,1)+1;
                else
                    table(track,2)=table(track,2)+1;
                end
            end
        end
    end
    M{fi}=table;
end
min(my_tracks)
save('Minutia.mat','M');
```

```
clear M
```

The following functions are used by the get_min function:

The "fft_enhance_cubs" function is available at [58]
The "get_core function" is available at appendix A.
Functions MinutiaPointExtraction,
RemoveFalseTermination, RemoveFalseBifurcation,
getLength (used in function RemoveFalseTermination), and
FpPlot are as shown below:

```
function Im=RemoveFalseTermination(Im,Imp)
[a,b]=size(Imp);
for i=1:a
    for j=1:b
        if (Imp(i,j)==1)
            [L,Im]=getLength(i,j,Im,1);
            if (L<22)
                [~,Im]=getLength(i,j,Im,0);
            end
        end
    end
end
```

```
function ret=MinutiaPointExtraction(I)
[a,b]=size(I);
hj=b/2;
hi=a/2;
ioff=[-1  -1  -1   0  1  1   1  0];
joff=[-1   0   1   1  1  0  -1 -1];
P=zeros(1,9);
for i=1:a
    for j=1:b
        if I(i,j)==1
            for k=1:8
                try
                    P(k)=I(i+ioff(k),j+joff(k));
                catch
                    P(k)=0;
                end
            end
            P(9)=P(1);C=0;
            for k=1: 8
                C=C+abs(P(k+1)-P(k));
            end
            ret(i,j)=C/2;
```

```
            if (ret(i,j)==1)% || (ret(i,j)==3)
                ok=0;
                if (j<=hj)
                    for jj=1:j-1
                        if I(i,jj)==1,ok=1;break;end
                    end
                else
                    for jj=j+1:b
                        if I(i,jj)==1,ok=1;break;end
                    end
                end
                if (ok==0)
                    ret(i,j)=0;
                else
                    ok=0;
                    if (i<=hi)
                        for ii=1:i-1
                            if I(ii,j)==1,ok=1;break;end
                        end
                    else
                        for ii=i+1:a
                            if I(ii,j)==1,ok=1;break;end
                        end
                    end
                end
                if (ok==0),ret(i,j)=0;end
            end
        end
    end
end

function [ret,x]=getLength(i,j,x,newval)
if x(i,j)==0
    ret=0;
    return;
end
x=double(x);
ret=0;
v=inf;
x(i,j)=2;
ioff=[-1   -1 -1   0  1  1   1  0  ];
joff=[-1   0    1   1  1  0  -1 -1  ];
while not(or(v==0,v==2))
    x(i,j)=2;
    kb=0;
    for k=1:8
        try
            P(k)=(x(i+ioff(k),j+joff(k))==1);
        catch
            P(k)=0;
```
178

```
        end
        if (P(k)==1)
            kb=k;
        end
    end
    v=sum(P);
    ret=ret+v;
    if (kb==0)
        break;
    else
        i=i+ioff(kb);
        j=j+joff(kb);
    end
end
try
    i=i-ioff(kb);
    j=j-joff(kb);
    x(i,j)=1;
end
x(x==2)=newval;
x=logical(x);
```

```
function Imp=RemoveFalseBifurcation(Imp)
[a,b]=size(Imp);
s=10;
for i=s:a-s
    for j=s:b-s
        if (Imp(i,j)==3)
            try
                x=Imp(i-s:i+s,j-s:j+s);
                c=sum(sum(x==3));
                if c>1
                    x(x==3)=0;
                    Imp(i-s:i+s,j-s:j+s)=x;
                end
            end
        end
    end
end
```

```
function FpPlot(mint,x,y,xc,yc)
hold off;
imshow(mint)
hold on;
plot(xc,yc,'o','MarkerEdgeColor','m','MarkerFaceColor','y',
'MarkerSize',10);
for i=1:length(x)
    if mint(y(i),x(i))==1
        plot(x(i),y(i),'wo','linewidth',1);
    else
```

```
        plot(x(i),y(i),'wd','linewidth',1);
    end
end
hold off
```

The following M-file is used to evaluate the performance of the proposed matching algorithm (FAR, FRR, EER, ZeroFAR, ZeroFRR, ROC):

```
close all
clc
clear all
load('MinutiaDB'); %tracks=13 salts=4 fingers=20
no_prints=3
% load('MinutiaDB1_Afeb11');  %tracks=10 fingers 10,92 and
all their copies
% are deleted so fingers are 98 and length =784 salts=3
% load('MinutiaDB1_Bfeb11');%tracks=14 finger 9 and all its
copies are deleted so
% fingers are 9 and length is 72 salts=4
% load('MinutiaDB2_Afeb11');%tracks=11 salts=3 fingers=100
no_prints=8
% load('MinutiaDB2_Bfeb11');%tracks=15 salts=5 fingers=10
no_prints=8
% load('MinutiaDB3_Afeb11');%tracks=14 salts=4 fingers=100
no_prints=8
% load('MinutiaDB3_Bfeb11');%tracks=14 salts=4 fingers=10
no_prints=8
% load('MinutiaDB4_Afeb11');%tracks=8 salts=2 fingers=100
no_prints=8
% load('MinutiaDB4_Bfeb11');%tracks=11 salts=3 fingers=10
no_prints=8

%defining the constant representing the fingerprints
database for
%MinutiaeDB1_Bfeb11
fingers=20;
no_prints=3;
length=fingers*no_prints;
%No of tracks to identify the print
mins=13;

%No of random points to introduce noise
salts=4;%for db1_A

%maximum minatues to change
maxsalts=1;
```

```matlab
%thresholds of matching
thmax=70;

%Now in 2D space
rand('state',0101487403)
% defaultStream.State = savedState;
far=zeros(thmax,thmax);
frr=zeros(thmax,thmax);
for i=1:length
    temp=M{i}(1:mins,1:2);
    for r=1:salts
        mintype=randi(2);
        mintrack=randi(mins);
        salt=randi(maxsalts+1)-1;
        temp(mintrack,mintype)=temp(mintrack,mintype)+salt;
        if temp(mintrack,mintype)<0
            temp(mintrack,mintype)=0;
        end
        mintype=3-mintype;
        temp(mintrack,mintype)=temp(mintrack,mintype)-salt;
        if temp(mintrack,mintype)<0
            temp(mintrack,mintype)=0;
        end
    end
    for k=0:fingers-1
        temp21=M{no_prints*k+1}(1:mins,1:2);
        temp22=M{no_prints*k+2}(1:mins,1:2);
        temp23=M{no_prints*k+3}(1:mins,1:2);
        %           temp24=M{no_prints*k+4}(1:mins,1:2);
        %           temp25=M{no_prints*k+5}(1:mins,1:2);
        %           temp26=M{no_prints*k+6}(1:mins,1:2);
        %           temp27=M{no_prints*k+7}(1:mins,1:2);
        %           temp28=M{no_prints*k+8}(1:mins,1:2);
        same=0;
        if floor((i-1)/no_prints)==k
            same=1;
        end
        match=0;
        d11 = sum(abs(temp-temp21));
        d12 = sum(abs(temp-temp22));
        d13 = sum(abs(temp-temp23));
        %           d14 = sum(abs(temp-temp24));
        %           d15 = sum(abs(temp-temp25));
        %           d16 = sum(abs(temp-temp26));
        %           d17 = sum(abs(temp-temp27));
        %           d18 = sum(abs(temp-temp28));

        d2=floor(nthroot(d11.*d12.*d13,no_prints));
```

```
        %
d2=floor(nthroot(d11.*d12.*d13.*d14.*d15.*d16.*d17.*d18,no_
prints));

        if d2(1,1)==0||d2(1,1)==1
            d2(1,1)=2;
        end
        if d2(1,2)==0||d2(1,2)==1
            d2(1,2)=2;
        end

        for thx=1:thmax,
            for thy=1:thmax
                if same==1
                    if d2(1,1)>thx || d2(1,2) > thy
                        frr(thx,thy)=frr(thx,thy)+1;
                    end
                else
                    if d2(1,1)<=thx && d2(1,2) <= thy
                        far(thx,thy)=far(thx,thy)+1;
                    end
                end
            end
        end
    end
end
far=far /(length*(fingers-1));
frr=frr /length;
% Finding the intersection of the 2 surfaces
x=1:thmax;
y=x;
z1=far;
z2=frr;

figure;
% Visualize the two surfaces
surface(x, y, z1, 'FaceColor', [1 0 0], 'EdgeColor',
'none');
surface(x, y, z2, 'FaceColor', [0 0 1], 'EdgeColor',
'none');
view(3); camlight; axis vis3d
xlabel('threshold2')
ylabel('threshold1')
zlabel('error rate')
hold on

% Take the difference between the two surface heights and
find the contour
% where that surface is zero.
```

```
zdiff = z1 - z2;
C = contours(x, y, zdiff, [0 0]);

% Extract the x- and y-locations from the contour matrix C.
xL = C(1, 2:end);
yL = C(2, 2:end);

% Interpolate on the first surface to find z-locations for
the intersection
% line.
zL = interp2(x, y, z1, xL, yL);

% Visualize the line.
line(xL, yL, zL, 'Color', 'k', 'LineWidth', 2);

[m,idx]=min(zL);
fprintf('EER=%f  at thx=%f and thy=%f and
sumth=%f\n',m,yL(idx),xL(idx),yL(idx)+xL(idx));

thxi(1)=floor(xL(idx));
thyi(1)=floor(yL(idx));
farv(1)=far(thyi(1),thxi(1));
frrv(1)=frr(thyi(1),thxi(1));
differ(1)=abs(farv(1)-frrv(1));

thxi(2)=floor(xL(idx));
thyi(2)=floor(yL(idx))+1;
farv(2)=far(thyi(2),thxi(2));
frrv(2)=frr(thyi(2),thxi(2));
differ(2)=abs(farv(2)-frrv(2));

thxi(3)=floor(xL(idx))+1;
thyi(3)=floor(yL(idx));
farv(3)=far(thyi(3),thxi(3));
frrv(3)=frr(thyi(3),thxi(3));
differ(3)=abs(farv(3)-frrv(3));

thxi(4)=floor(xL(idx))+1;
thyi(4)=floor(yL(idx))+1;
farv(4)=far(thyi(4),thxi(4));
frrv(4)=frr(thyi(4),thxi(4));
differ(4)=abs(farv(4)-frrv(4));
maxv=1000;
for i=1:4

    if differ(i) <maxv
        temp=i;
        maxv=differ(i);
```

```matlab
    end
end

fprintf('FAR=%f , FRR=%f and diff=%f at thx=%d and
thy=%d\n',farv(temp),frrv(temp),differ(temp),thyi(temp),thx
i(temp));

% surf(yL(idx),xL(idx),m)
hold off

%find the zeroFAR
zero_loc_far=find(far==0);
temp_frr=frr(zero_loc_far);
zero_far=min(temp_frr);
[zfar_th1 zfar_th2]=find(frr==zero_far)

%find the zeroFRR
zero_loc_frr=find(frr==0);
temp_far=far(zero_loc_frr);
zero_frr=min(temp_far);
[zfrr_th1 zfrr_th2]=find(far==zero_frr)

plot(log(100*far(:,35)),log(100*frr(:,35)))
xlabel('log 100*FAR')
ylabel('log 100*FRR')
axis([-1 3 0.5 4])
```

# Appendix C
# Thinning algorithm Matlab source code

This appendix is the MATLAB source code of the proposed thinning algorithm for fingerprints.

```matlab
close all
clear all
t1 = cputime;
CrnFgp=imread('70_1.tif');
% CrnFgp=imread('101_3.tif');
% imshow(CrnFgp)
th=245; %to suppress noise
% imshow(CrnFgp)
N=size(CrnFgp);
blksize=8;
N(1)=floor(N(1)/blksize)*blksize;
N(2)=floor(N(2)/blksize)*blksize;

% determine horizontal and diagonal ridges
newarray=zeros(N(1)/blksize,N(2)/blksize);%where each block
in the fingerprint corresponds to a cell in this array
% if it contains zero means no ridge at all,
% 1 means a horizontal ridge,
% 2 means a vertical ridge

%%%%%%%%%%%%%%%%%%%%%%%%%%%%%%%%%%%%%%%%%%%%%%%%%%%%%%%
% Determine the type of each block before
thinning(horizontal or vertical ridge)
for i=1:blksize:N(1)
    for j=1:blksize:N(2)
        test=CrnFgp(i:i+blksize-1,j:j+blksize-1);%obtain a
block
        if (find(test<=th))%to be sure not background
            sumblkcols=sum(test,1);%calculate the sum of
columns of block
            sumblkrows=sum(test,2);%calculate the sum of
rows of block
            [min1,mat1]=min(sumblkcols);
            [min2,mat2]=min(sumblkrows);
            if min1>min2%horz

newarray(ceil(i/blksize),ceil(j/blksize))=1;
            else        %vert

newarray(ceil(i/blksize),ceil(j/blksize))=2;
            end
```

```matlab
        end
    end
end

%correcting errors of ridge type results
for type_row=2:N(1)/blksize-1
    for type_col=2:N(2)/blksize-1
        type=newarray(type_row,type_col);
        left_type=newarray(type_row,type_col-1);
        right_type=newarray(type_row,type_col+1);
        top_type=newarray(type_row-1,type_col);
        bottom_type=newarray(type_row+1,type_col);
        if type==2%detect error of a wrongly detected vert
ridge between two horz ridges
            %             detect error of a wrongly
detected vert ridge whereas it is a horz edge
            if
(left_type==1&&right_type==1)||(top_type==1&&bottom_type==1
)||...

(left_type==0&&right_type==1)||(left_type==1&&right_type==0
)
                newarray(type_row,type_col)=1;%update the
value
                %remove a noisy block
            elseif
left_type==0&&right_type==0&&top_type==0&&bottom_type==0
                newarray(type_row,type_col)=0;
            end
        elseif type==1%detect error of a wrongly detected
horz ridge between two vert ridges
            %             detect error of a wrongly
detected horz ridge whereas it is a vert edge
            if
(top_type==2&&bottom_type==2)||(left_type==2&&right_type==2
)||...

(top_type==0&&bottom_type==2)||(left_type==2&&right_type==0
)
                newarray(type_row,type_col)=2;%update the
value
                %remove a noisy block
            elseif
left_type==0&&right_type==0&&top_type==0&&bottom_type==0
                newarray(type_row,type_col)=0;
            end
        end
    end
end
newarray_lab = bwlabel(newarray,4);
```

```matlab
newarray_close = bwmorph(newarray_lab,'close');
newarray_open = bwmorph(newarray_close,'open');
inBound = bwperim(newarray_open);
inArea=newarray_open-inBound;
[CrnFgp,newarray] =
SegmImg(CrnFgp,newarray,inBound,inArea,blksize,0);
N=size(CrnFgp);
N(1)=floor(N(1)/blksize)*blksize;
N(2)=floor(N(2)/blksize)*blksize;
ThinFgp=ones(N);%initialize Thin fingerprint image
%get the indices of blk_type1 alone
[k,l]=find(newarray==1);
i=1;count=1;
first=l(i);
x(count)=k(i);y(count)=l(i);
for i=2:length(k)
    count=count+1;
    if (l(i)==first)&&(k(i)==k(i-1)+1)
        x(count)=k(i);
        y(count)=l(i);
    else
        x(count)=0;y(count)=0;
        count=count+1;
        x(count)=k(i);y(count)=l(i);
        first=l(i);
    end
end
count=count+1;
x(count)=0;y(count)=0;
x_type1=x;y_type1=y;

%get the indices of blk_type2 alone
clear k l x y
[k,l]=find(newarray==2);
count=0;x=0;
for i=1:length(k)
    first=k(i);
    mat=find(x==first);%not to repeat the search of this
index
    if isempty(mat)
        count=count+1;
        x(count)=k(i);y(count)=l(i);
        loc=find(k==first);
        for w=2:length(loc)
            count=count+1;
            if l(loc(w))==l(loc(w-1))+1
                x(count)=k(loc(w));
                y(count)=l(loc(w));
            else
                x(count)=0;y(count)=0;
```

```
                    count=count+1;
                    x(count)=k(loc(w));y(count)=l(loc(w));
                end
            end
            count=count+1;
            x(count)=0;y(count)=0;
        end
    end
x_type2=x;y_type2=y;
%search for the local minimum in each group of consecutive
blocks of type1
w=1;
zeroloc=find(x_type1==0);
count0=length(zeroloc);
for h=1:count0
    for i=(x_type1(w)-1)*blksize+1:x_type1(zeroloc(h)-
1)*blksize
        if i+2<=N(1)
            for j=(y_type1(w)-
1)*blksize+1:y_type1(w)*blksize
                if
(CrnFgp(i,j)>CrnFgp(i+1,j))&&(CrnFgp(i+1,j)<CrnFgp(i+2,j))
                    ThinFgp(i+1,j)=0;
                end
            end
        end
    end
    w=zeroloc(h)+1;
end
%search for the local minimum in each group of consecutive
blocks of type2
w=1;where0=0;
zeroloc=find(x_type2==0);
count0=length(zeroloc);
for h=1:count0
    for i=(x_type2(w)-1)*blksize+1:x_type2(w)*blksize
        for j=(y_type2(w)-1)*blksize+1:y_type2(zeroloc(h)-
1)*blksize
            if j+2<=N(2)
                if
(CrnFgp(i,j)>CrnFgp(i,j+1))&&(CrnFgp(i,j+1)<CrnFgp(i,j+2))
                    ThinFgp(i,j+1)=0;
                end
            end
        end
    end
    w=zeroloc(h)+1;
end
% figure,imshow(ThinFgp)
ThinFgp=~ThinFgp;
```

```matlab
% figure,imshow(ThinFgp)
ThinFgp=bwmorph(ThinFgp,'clean');
% figure,imshow(ThinFgp)
%handling bifurcation problem
for i=4:N(1)-4
    for j=4:N(2)-4
        if ThinFgp(i,j)==1
            tempT1=(ThinFgp(i-1:i+1,j-1:j+1));
            if (sum(sum(tempT1))==2)
temp=double(CrnFgp(i-1:i+1,j-1:j+1));
                temp(2,2)=260;%to exclude it from
comparison
                [m,nr]=min(temp);
                [m,nc]=min(m);
                minr=nr(nc);minc=nc;
                if ThinFgp(i+minr-2,j+minc-2)==0
                    tmp=ThinFgp(i+minr-2-1:i+minr-
2+1,j+minc-2-1:j+minc-2+1);
                    if sum(sum(tmp))>=3
                        ThinFgp(i+minr-2,j+minc-2)=1;
                        if (ThinFgp(i+minr-2,j+minc-2-
                        1)==1)&&(ThinFgp(i+minr-2,j+minc-2-
                        2)==0)%wrong left duplicate

                        ThinFgp(i+minr-2,j+minc-2-1)=0;
                        elseif (ThinFgp(i+minr-2,j+minc-
                        2+1)==1)&&(ThinFgp(i+minr-2,j+minc-
                        2+2)==0)%wrong right duplicate

                        ThinFgp(i+minr-2,j+minc-2+1)=0;
                        end
                    end
                end
            end
        end
    end
end
figure,imshow(ThinFgp)
t2 = cputime;
sprintf('execution time is %2.2d sec',t2-t1)
```

## The following function is used by the thinning function

```matlab
function [roiImg,roiarray] =
SegmImg(in,newarray,inBound,inArea,blksize,noShow)
% construct a ROI rectangle for the input fingerprint
image and return the
%  segmented fingerprint
```

```
% With the assumption that only one ROI region for each
fingerprint image
[iw,ih]=size(in);
tmplate = zeros(iw,ih);
[w,h] = size(inArea);
tmp=zeros(iw,ih);

left = 1;
right = h;
upper = 1;
bottom = w;

le2ri = sum(inBound);
roiColumn = find(le2ri>0);
left = min(roiColumn);
right = max(roiColumn);

up2dw=sum(inBound,2);
roiRow = find(up2dw>0);
upper = min(roiRow);
bottom = max(roiRow);

roiImg  = in(blksize*upper-(blksize-
1):blksize*bottom,blksize*left-(blksize-1):blksize*right);
roiarray = newarray(upper:bottom,left:right);

% figure,imshow(roiImg)

if nargin == 4
    colormap(gray);
    imagesc(tmplate);
end;
```

# Appendix D
# Cryptographic methods

## Encryption and digital signatures

There are two types of key-based encryption algorithms: *symmetric* and *public* key. A key is typically a string of bits and the longer the length of the string, the harder it is to break the encrypted message, but the longer it takes to encrypt and decrypt a message. Symmetric cryptography uses a single secret key for both encrypting and decrypting a message. The advantage of this type of encryption is that a large number of fast and effective algorithms are available and the secrecy of the key guarantees a high level of security. The disadvantage of symmetric encryption lies in the difficulty of key distribution. Public-key cryptography uses a pair of keys for each entity that takes part in the communication. Here the key that is used to encrypt a message is different from the key used to decrypt the message (and hence the key pair is called "asymmetric"). Furthermore, the encryption and decryption keys cannot be derived from each other, even if a large amount of messages encrypted from a public or a private key is available. One of the keys of the pair is called the private key that is held by a specific person and kept secret. The other key is called the public key and is made known to everybody who wants to communicate with this specific person.

The advantage of public-key cryptography over symmetric cryptography is that key management becomes easier. The main disadvantage is that it is several orders of magnitude slower than symmetric cryptography [80]. Rivest-Shamir-Adleman (RSA) [81] is the most popular public-key algorithm. These symmetric and asymmetric encryption techniques can be used to secure a communication link for data integrity. However, they do not solve the non-repudiation problem. Digital signature is the technology that can ensure/certify that a message has been sent by a specific person who cannot deny having sent the message.

The popular RSA public-key algorithm itself is a straightforward way to digitally sign and encrypt a message.

**Timestamp and challenge-response**

Encryption solves the problem of data integrity and digital signatures solve the problem of non-repudiation. However, a digitally signed encrypted message from A to B can still be "sniffed" from a communication channel and replayed at a later time. The receiver, B, will not be able to know that this replayed message did not come from A or was not intended for B. A simple solution to this problem requires that the recipient store all the previous messages it has received and compare a new message against them to ascertain that the new message is not a copy of a previously received message. This approach requires the recipient to have a very large amount of storage space and a fast comparison algorithm. An optimization of this approach may involve storing hashes of the messages instead of the entire messages and comparing the hash of a new message with them. However, the storage and comparison time required in this solution are still prohibitive. Therefore the most popular solutions to the replay attacks involve building either a timestamp or a challenge-response mechanism in the communication link.

In a timestamp-based scheme, when A wants to communicate with B, she sends B an encrypted and signed request message that includes a timestamp taken from her internal clock. When B receives the request, he decrypts it and checks the timestamp and compares it with his internal clock. If the difference in time is within a threshold (typically computed based on the expected time of message transfer), B responds to A by sending her the requested data and his timestamp in an encrypted form. Upon receiving the response from B, A decrypts the message and checks the timestamp. This solution requires that the clocks of the sender and receiver be synchronized, which is difficult to achieve in practice.

The popular challenge-response schemes are based either on nonce or session keys. These schemes do not require the use of timestamps. In the nonce-based scheme, if A wants to communicate with B, she sends B an encrypted and signed request message that includes a random number (called the nonce). B responds to A by sending her the requested data and the nonce in an encrypted form. Upon receiving the response from B, A decrypts the message and the nonce and compares the received nonce with the one she had sent.

# Appendix E
# Watermark embedding and extraction algorithms Matlab source codes

This appendix is the MATLAB source code of the proposed watermark embedding and extraction algorithms for fingerprints.

```matlab
Function [org_wat]=emb_fgp(img_file,fgpno)
close all
load Minutia
%%%%%compose the binary watermark from minutiae table
% max no. of minutiae in all database was 12 so 4 binary
digits are enough
map=M{fgpno};
s=size(map);
no_of_bin=4;
bitmap=zeros(s(1),no_of_bin*s(2));
for i=1:s(1)
    for j=1:s(2)
        decnum=map(i,j);
        rem=zeros(1,no_of_bin);
        quo=10;
        t=no_of_bin+1;
        while quo>0
            t=t-1;
            decnum=double(decnum)/2;
            quo=floor(decnum);
            if(decnum-quo)==0.5
                rem(t)=1;
            else
                rem(t)=0;
            end
            decnum=quo;
        end
        binnum=rem;
        incol=(j-1)*no_of_bin;%to adjust the index in
bitmap matrix
        for inj=1:no_of_bin
            bitmap(i,incol+inj)=rem(inj);
        end
    end
end
org_wat=bitmap(:);
q=0.1;A=100;B=1000;
org_img=imread(img_file);
```

```matlab
s=size(org_img);
%augmenting 0 and 1 to the first of the watermark
org_wat=[0 1 org_wat'];
wat_len=length(org_wat);
rand('state',23021979);
perm_sig=org_wat(randperm(wat_len));%permute signature
%get the locations of embedding in the host image
c=2;raw=0.18;n=0;
rand('state',0101487403)
for i=1:s(1)
    for j=1:s(2)
        %    change the background to make it not white to
make watermarking invisible
        if(org_img(i,j)>=245)&&(org_img(i,j)<=255)
            org_img(i,j)=org_img(i,j)-20;
        end
        if rand<raw && (i>c && i<=s(1)-c) && (j>c &&
j<=s(2)-c)
            n=n+1;
            iz(n)=i;
            jz(n)=j;
        end
    end
end
% figure,imshow(org_img);
%n is no. of pixels that will be changed
n_rep=floor(n/wat_len);%no. of repititions
w_pix=0;%initializing the no. of watermarked pixels
ind_perm_sig=1;
org_img=double(org_img);
%%%%from paper hiding fingerprint minutiae in images
%compute the average and standard deviation at each pixel
avg_img=org_img;
std_img=org_img;c=2;
for i=c+1:s(1)-c
    for j=c+1:s(2)-c
        temp1=org_img(i-c:i+c,j-c:j+c);
        avg_img(i,j)=mean(mean(temp1));
        temp2=[org_img(i-c:i+c,j)',org_img(i,j-c:j-
1),org_img(i,j+1:j+c)];
        std_img(i,j)=std(temp2);
    end
end
%compute the gradient magnitude image
%from http://en.wikipedia.org/wiki/Sobel_operator
filt=fspecial('sobel');
Gy=filter2(filt,org_img);
Gx=filter2(filt',org_img);
grad_img=sqrt(Gx.^2+Gy.^2);
%initialize wat image
```

```
ind=1;
bit_rep=n_rep;%initialize bit_rep
wat_img=org_img;
while w_pix<(n_rep*wat_len)
    if bit_rep==n_rep
        bit_rep=0;
        sig_bit=perm_sig(ind_perm_sig);
        ind_perm_sig=ind_perm_sig+1;
    end

wat_img(iz(ind),jz(ind))=wat_img(iz(ind),jz(ind))+(2*sig_bi
t-1)*avg_img(iz(ind),jz(ind))*q*...

(1+(1/A)*std_img(iz(ind),jz(ind)))*(1+(1/B)*grad_img(iz(ind
),jz(ind)));
    w_pix=w_pix+1;
    bit_rep=bit_rep+1;
    ind=ind+1;
end
s=Similar(double(wat_img(:)),double(org_img(:)))

wat_img=uint8(wat_img);
% figure,imshow(wat_img);
imwrite(wat_img,'wat_fgp.tif','tif'); %writing the result
watermarked fingerprint image

function
dec_wat=ext_fgp_kut(wat_img,sec_key1,sec_key2,org_wat,raw,c
,fgpno)
s=size(wat_img);
sig_len=length(org_wat);
%get the locations of embedding in the host image
rand('state',sec_key1)
n=0;
for i=1:s(1)%c+1:s(1)-c
    for j=1:s(2)%c+1:s(2)-c
        if rand<raw && (i>c && i<=s(1)-c) && (j>c &&
j<=s(2)-c)
            n=n+1;
            iz(n)=i;
            jz(n)=j;
        end
    end
end

%n is no. of pixels that are changed
n_rep=floor(n/sig_len);%no. of repitions
%computing threshold value
rand('state',sec_key2);
```

```matlab
perm_ind=randperm(sig_len);
ind_0=find(perm_ind==1);%to access first bit (0) of
signature
ind_1=find(perm_ind==2);%to access second bit (1) of
signature
bit_0_place=n_rep*(ind_0-1)+1;
bit_1_place=n_rep*(ind_1-1)+1;
wat_img=double(wat_img);
%computing delta_k of bit 0
i=1;
for ind=bit_0_place:bit_0_place+(n_rep-1)
    cur_val=wat_img(iz(ind),jz(ind));
    pred_val=(1/(4*c))*(sum(wat_img(iz(ind)-
c:iz(ind)+c,jz(ind))))+...
        sum(wat_img(iz(ind),jz(ind)-c:jz(ind)+c))-
2*cur_val);
    delta_k(i)=cur_val-pred_val;
    i=i+1;
end
delta_0=(1/(raw*s(1)*s(2)))*sum(delta_k);
%computing delta_k of bit 1
i=1;
for ind=bit_1_place:bit_1_place+(n_rep-1)
    cur_val=wat_img(iz(ind),jz(ind));
    pred_val=(1/(4*c))*(sum(wat_img(iz(ind)-
c:iz(ind)+c,jz(ind))))+...
        sum(wat_img(iz(ind),jz(ind)-c:jz(ind)+c))-
2*cur_val);
    delta_k(i)=cur_val-pred_val;
    i=i+1;
end
delta_1=(1/(raw*s(1)*s(2)))*sum(delta_k);
%computing threshold
thr=(delta_0+delta_1)/2;

sig_ind=1;
for i=1:n_rep:n-n_rep
    bit_rep=0;
    if sig_ind<=sig_len
        for ind=i:i+(n_rep-1)
            bit_rep=bit_rep+1;
            cur_val=wat_img(iz(ind),jz(ind));
            pred_val=(1/(4*c))*(sum(wat_img(iz(ind)-
c:iz(ind)+c,jz(ind))))+...
                sum(wat_img(iz(ind),jz(ind)-c:jz(ind)+c))-
2*cur_val);
            delta_k(bit_rep)=cur_val-pred_val;
        end

delta_bar(sig_ind)=(1/(raw*s(1)*s(2)))*sum(delta_k);
```

```matlab
        if delta_bar(sig_ind)>thr
            perm_sig(sig_ind)=1;
        else
            perm_sig(sig_ind)=0;
        end
    end
    sig_ind=sig_ind+1;
end
rand('state',sec_key2);
real_ind=randperm(sig_len);
ext_sig(real_ind)=perm_sig;%inverse permute signature
s=Similar(org_wat,ext_sig(:))
%remove the first 0,1
ext_sig=ext_sig(3:end);
tracks=length(ext_sig)/8;
ext_sig=reshape(ext_sig,tracks,8);
% figure, imshow(ext_sig)
dec_wat=zeros(tracks,2);
for i=1:tracks
    no1=num2str(ext_sig(i,1:4));
    no2=num2str(ext_sig(i,5:8));
    dec_wat(i,1)=bin2dec(no1);
    dec_wat(i,2)=bin2dec(no2);
end
fid=fopen('amira1.txt','at')
fprintf(fid,'%.2f\n',s*100)
fclose(fid)

delta_bar(real_ind)=delta_bar;
figure,plot(delta_bar,'k*'),hold
on,plot(thr*ones(sig_len),'k')
xlabel('bit index'),ylabel('delta bar')
text(15,thr+0.01,'threshold value')

to reconstruct the fingerprint image
initialize the pred wat image
pred_wat_img=wat_img;
for i=c+1:size(wat_img,1)-c
    for j=c+1:size(wat_img,2)-c
        pred_wat_img(i,j)=(1/(4*c))*(sum(wat_img(i-
c:i+c,j))+...
            sum(wat_img(i,j-c:j+c))-2*wat_img(i,j));
    end
end

s=Similar(wat_img(:),pred_wat_img(:))
figure,imshow(uint8(pred_wat_img))
```

أخيرا، صحة البيانات التي يتم نقلها عبر قنوات الاتصال يجب أن تكون آمنة طوال المسافة من الجهاز الماسح لبصمة الإصبع وحتى التطبيق. يتحقق هذا الأمر بشكل خاص بطرق التشفير. لهذا تم اقتراح خوارزم لإدراج علامة مائية، حيث يتم فيه إدراج جدول الخصائص المقترح في بصمة الإصبع الخاصة به باستخدام مفتاحين سريين لزيادة الأمان. لا يحتاج خوارزم إدراج العلامة المائية إلى صورة البصمة الأصلية لاستخراج العلامة المائية. تم دراسة تطبيقين من تطبيقات الأمان للعلامة المائية التي يمكن استعمالها لضمان نقل آمن لصور بصمة الإصبع من وكالات الاستخبارات وحتى قاعدة البيانات المركزية للصور واستعمالها كذلك لتجنب العديد من أنواع الهجوم على أنظمة المقاييس الحيوية.

أظهرت نتائج التجارب العملية أن العلامة المائية غير مرئية وقادرة على مقاومة إضافة الضوضاء الجاوسيانية والضغط بمعاملات جودة عالية ومتوسطة. وجد كذلك أن بعد إدراج العلامة المائية وبدون تطبيق أي هجوم على الصور، لم تتغير دقة التعرف بل ظلت كسابقها 98%. أما بعد إضافة الضوضاء الجاوسيانية والضغط بمعاملات جودة عالية ومتوسطة على صور بصمة الإصبع بعد إدراج العلامة المائية فيها، انخفضت دقة التعرف انخفاضا بسيطا حيث وصلت إلى 96%.

## كلمات مفتاحية:

المقاييس الحيوية، بصمات الأصابع، التقطيع، معالجة الصورة، التعرف على بصمة الإصبع، جابور فلتر، تحسين بصمة الإصبع، جعل الصورة ثنائية اللون، خوارزمات الترقيق، الهيكل، تحليل مجال الاتجاه، استخراج السمات، المقارنة باستخدام التفاصيل، التحديد المتكيف للنقطة الفريدة، نقطة اللب، تمييز بصمة الإصبع، إثبات صحة بصمة الإصبع، إثبات صحة الهوية بطريق آمنة، نموذج التهديد، العلامة المائية، أمان المعلومات، إخفاء البيانات.

# مستخلص

**أميرة محمد عبد الموجود محمد صالح**

**خوارزم آمن ومحسن للتعرف على بصمة الإصبع**

**رسالة دكتوراة**

**جامعة عين شمس ـ كلية الهندسة ــ 2011**

أصبح التعرف على الأشخاص باستخدام خصائص المقاييس الحيوية ظاهرة ملحة في مجتمعنا. خلال السنوات الماضية، أدت الحاجة إلى الأمان في نطاق واسع من التطبيقات إلى الاهتمام المتزايد بهذا الأمر. بين العديد من هذه المقاييس الحيوية احتلت بصمة الإصبع المكانة الأكثر عملية. هذا الأمر كان بسبب أن التعرف على بصمة الإصبع يتطلب أدنى جهد من المستخدم، كذلك لا يأخذ معلومات أكثر مما هو ضروري لعملية التعرف على بصمة الإصبع، وأيضا تمدنا نسبيا بأداء جيد. السبب الآخر لانتشار بصمة الإصبع هو السعر الرخيص لحساسات بصمة الإصبع التى يسرت إدماجها في لوحات مفاتيح الحاسبات الشخصية والكروت الذكية والمعدات اللاسلكية.

تعتبر مرحلة ترقيق صورة بصمة الإصبع الداخلة لنظام التعرف على بصمة الإصبع خطوة حرجة حيث يعتمد أداء الخوارزم المسؤول عن استخراج تفاصيل بصمة الإصبع اعتمادا كبيرا على جودة الخوارزم المستعمل في الترقيق. لذلك تم اقتراح خوارزم سريع لترقيق صورة بصمة الإصبع. هذا الخوارزم يعمل مباشرة على الصورة ذات التدريج الرمادي لبصمة الإصبع بدلا من الصورة الأبيض والأسود لها لأن تحويل بصمة الإصبع للأبيض والأسود يسبب الكثير من التفاصيل الزائفة وأيضا يزيل الكثير من السمات المهمة. تم تقييم أداء خوارزم الترقيق وأوضحت نتائج التجارب العملية أن خوارزم الترقيق المقترح سريع ودقيق.

الخطوة التالية للترقيق هي استخراج خصائص صورة بصمة الإصبع. تستعمل هذه الخصائص مع النموذج الذي نحصل عليه من قاعدة البيانات في مرحلة مطابقة بصمة الإصبع. في هذه الدراسة، تم اقتراح طريقة جديدة لمطابقة بصمة الإصبع تعتمد على الخصائص المميزة لهذه البصمة. الفكرة الرئيسية هي أن كل بصمة إصبع تمثل بجدول مكون من عمودين تحديدا في قاعدة البيانات. في هذا الجدول، يتم تسجيل عدد الخصائص المميزة (النهايات والتفريعات) لهذه البصمة الموجودة في مسارات ذات عرض معين حول نقطة المنتصف لهذه البصمة. كل صف في هذا الجدول يمثل مسار محدد، في العمود الأول، يتم تسجيل عدد النهايات في كل مسار، في العمود الثاني، يتم تسجيل عدد التفريعات في كل مسار. بما أن الجدول الممثل لبصمة الإصبع في قاعدة البيانات لا يحتوي على أي تحديد لإحداثيات أو زاوية خصائص البصمة، الخوارزم لا يتأثر بدوران أو تحريك بصمة الإصبع وكذلك لا يحتاج مساحة كبيرة للتخزين. أوضحت نتائج التجارب العملية أن دقة التعرف للخوارزم المقترح هي 98% بمعدل خطأ متساو قيمته 2%.

بسم الله الرحمن الرحيم

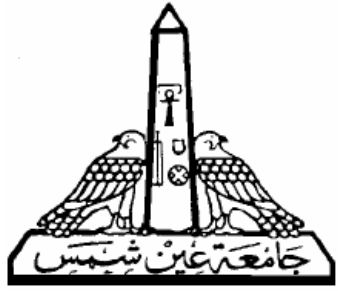

كلية الهندسة
قسم هندسة الحاسبات والنظم

# خوارزم آمن ومحسن للتعرف على بصمة الإصبع

رسالة
مقدمة للحصول على درجة الدكتوراة في الهندسة الكهربية
(هندسة الحاسبات والنظم)

مقدمة من
**المهندسة / أميرة محمد عبد الموجود محمد صالح**
بكالوريوس الهندسة الكهربية (هندسة الحاسبات والنظم)
كلية الهندسة جامعة عين شمس 2000
ماجستير في الهندسة الكهربية (هندسة الحاسبات والنظم)
كلية الهندسة جامعة عين شمس 2004

تحت إشراف
**الأستاذ الدكتور/ عبد المنعم عبد الظاهر وهدان**
**الأستاذ الدكتور/ أيمن محمد وهبة**
**الدكتور/ أيمن محمد بهاء الدين صادق**
قسم هندسة الحاسبات والنظم
القاهرة - 2011